\documentclass[letterpaper, 10 pt, conference]{ieeeconf}  

\IEEEoverridecommandlockouts                              

\usepackage{subcaption}
\usepackage{wrapfig}
\usepackage{amsmath}
\usepackage{graphicx}
\usepackage{amsfonts}
\usepackage{siunitx}
\usepackage{dsfont}
\usepackage{hyperref}
\usepackage{scalerel}
\usepackage{booktabs}

\newcommand{\sect}[1]{Sec.~\ref{#1}}
\newcommand{\fig}[1]{Fig.~\ref{#1}}
\newcommand{\eq}[1]{Eq.~\eqref{#1}}

\newcommand{\tab}[1]{Tab.~\ref{#1}}

\usepackage{float}

\usepackage{siunitx}
\usepackage{xcolor}

\newcommand{\mdptime}[0]{t}
\newcommand{\mdpstate}[0]{s}
\newcommand{\mdpaction}[0]{a}
\newcommand{\nextstate}[0]{\mdpstate_{\mdptime+1}}
\newcommand{\state}[0]{\mdpstate_\mdptime}
\newcommand{\action}[0]{a_t}
\newcommand{\hor}[0]{h}
\newcommand{\dynmod}[0]{f}
\newcommand{\traj}[0]{\mdpstate_{\mdptime+\hor}=\dynmod_\theta(\state,h,\theta_\pi)}	
\newcommand{\step}[0]{\nextstate=\dynmod_\theta(\state,\action)}	

\usepackage{tikz}

\newcommand{\cblock}[3]{
  \hspace{-1.5mm}
  \begin{tikzpicture}
    [
    node/.style={square, minimum size=10mm, thick, line width=0pt},
    ]
    \node[fill={rgb,255:red,#1;green,#2;blue,#3}] () [] {};
  \end{tikzpicture}%
}
\makeatletter
\newcommand{\bigplus}{%
  \DOTSB\mathop{\mathpalette\mattos@bigplus\relax}\slimits@
}
\newcommand\mattos@bigplus[2]{%
  \vcenter{\hbox{%
    \sbox\z@{$#1\sum$}%
    \resizebox{!}{0.9\dimexpr\ht\z@+\dp\z@}{\raisebox{\depth}{$\m@th#1+$}}%
  }}%
  \vphantom{\sum}%
}
\makeatother
\usepackage{amssymb}
\usepackage{fdsymbol}
\usepackage{wasysym}

\renewcommand{\vec}[1]{\boldsymbol{#1}}				

\newcommand{\degrees}[0]{\circ}

\newcommand{\D}[0]{\mathcal{D}} 				

\DeclareMathOperator*{\argmax}{arg\,max}

\newcommand{\titlelong}[0]{Learning Accurate Long-term Dynamics \\ for Model-based Reinforcement Learning}
\newcommand{\website}[0]{\url{https://sites.google.com/view/trajectory-prediction/}}

\usepackage{bm}
\newcommand{\citet}[1]{\cite{#1}}
\newcommand{\citep}[1]{\cite{#1}}

\begin{document}

\title{\LARGE \bf \titlelong{}}

\author{Nathan Lambert\textsuperscript{1}, Albert Wilcox\textsuperscript{1}, Howard Zhang\textsuperscript{1}, Kristofer S. J. Pister\textsuperscript{1}, and Roberto Calandra\textsuperscript{2}
\thanks{Corresponding author: Nathan O. Lambert, \tt \href{mailto:nol@berkeley.edu}{nol@berkeley.edu}}
\thanks{\textsuperscript{1}Department of Electrical Engineering and Computer Sciences, University of California, Berkeley, USA.}%
\thanks{\textsuperscript{2}Facebook AI Research, Menlo Park, CA, USA.}%
}

\maketitle
\thispagestyle{empty}
\pagestyle{empty}


\begin{abstract}
	Accurately predicting the dynamics of robotic systems is crucial for model-based control and reinforcement learning.
The most common way to estimate dynamics is by fitting a one-step ahead prediction model and using it to recursively propagate the predicted state distribution over long horizons.
Unfortunately, this approach is known to compound even small prediction errors, making long-term predictions inaccurate.
In this paper, we propose a new parametrization to supervised learning on state-action data to stably predict at longer horizons -- that we call a trajectory-based model.
This trajectory-based model takes an initial state, a future time index, and control parameters as inputs, and directly predicts the state at the future time index.
Experimental results in simulated and real-world robotic tasks show that trajectory-based models yield significantly more accurate long term predictions, improved sample efficiency, and the ability to predict task reward.
With these improved prediction properties, we conclude with a demonstration of methods for using the trajectory-based model for control.

\end{abstract}

\section{INTRODUCTION}
\label{sec:intro}

Model-based reinforcement learning (MBRL) has emerged as a compelling approach for control across multiple domains of robotics by using a learned forward dynamics model for planning~\citep{williams2017information, chua2018deep, schrittwieser2020mastering}. 
At the base of most current MBRL algorithms is the idea of learning \textit{one-step ahead forward dynamics models} -- that is, given a state and an applied action the model will predict the future state.
Once a one-step ahead model is learned, it can subsequently be used for planning or control by recursively applying the model to the state to unroll long-horizon trajectories.
While this approach has proved successful in many domains and is well-motivated from the related field of optimal control~\citep{garcia1989model, kirk2004optimal}, it is well understood that a major limit of current MBRL algorithms is the insufficient accuracy of the predictions over long-horizons.
In fact, one-step models critically suffer from compounding errors of the dynamics~\citep{chua2018deep, asadi2019combating, xiao2019learning}, and recent work has shown that the performance of one-step ahead prediction models is not necessarily correlated with downstream control performance~\citep{lambert2020objective}.

In this work, we propose and study a new class of feed-forward dynamics models focused on capturing not the behavior of single steps, but the long-term time dependant evolution of a trajectory as a whole.
The main intuitions behind this model are that: 
1) sequences of actions and states are usually strongly correlated with their neighbors across time (i.e., across a trajectory) and
2) in quantifying the quality of long-horizon predictions, it might be preferable to have higher uncertainty over the single steps, if the predication of the trajectory as a whole is more accurate. 
This is particularly crucial when the dynamics models are used for planning, where the relative ranking of the trajectories will directly impact the decision making (e.g., it might be relatively unimportant to know how we reach a location as long as we know that we can reach it accurately). 

\begin{figure}[t]
    \centering
    \begin{subfigure}{0.49\linewidth}
        \centering
        \includegraphics[width=\linewidth]{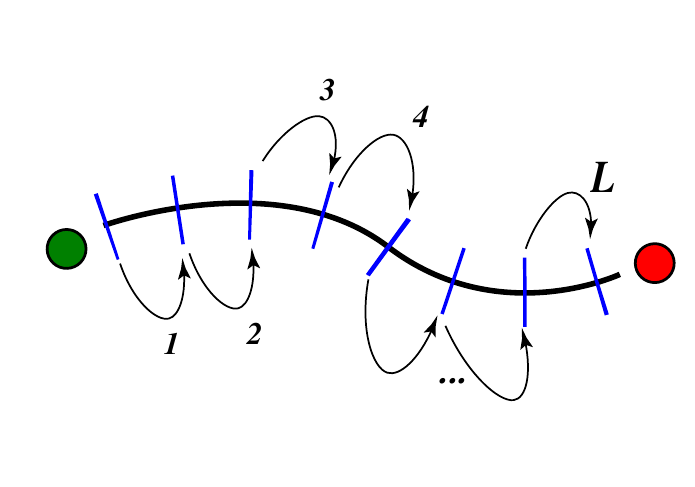}
        \caption{One-step models: \\\hspace{\textwidth} \centering $\step$
        } 
        \label{fig:sketch-step}
    \end{subfigure}
    \hfill
    \begin{subfigure}{0.49\linewidth}  
        \centering 
        \includegraphics[width=\linewidth]{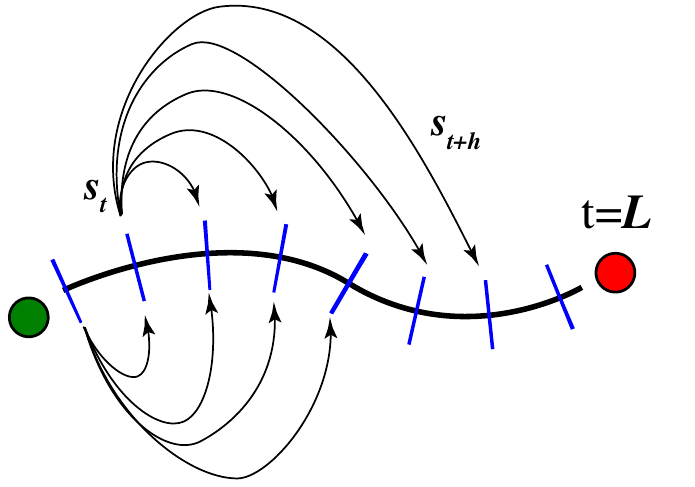}
        \caption{Trajectory-based models: \\\hspace{\textwidth} \centering  $\traj$
        }    
        \label{fig:sketch-traj}
    \end{subfigure}
    \caption{The model formulation for one-step models (\textit{a}) and our new \textbf{trajectory-based models} (\textit{b}). 
    }
    \label{fig:sketch1}
    \vspace{-10pt}
\end{figure}

Our contribution is to propose a new class of deep predictive models, \textbf{trajectory-based models}, that are focused on capturing the long-term, time-dependant nature of dynamics.
The new models re-frame the supervised learning problem to predict future states from the parameters of a controller parameters and a time index, rather than actions. 
The trajectory-based prediction steps is compared visually to individual steps in~\fig{fig:sketch1}.

This provides several benefits compared to traditional one-steps ahead models: improved data-efficiency, better computational complexity which can be further parallelized, a more principled treatment of uncertainty over trajectories, the capability of modeling continuous time steps, and finally more accurate long-term predictions.
We evaluate our trajectory-based approach on simulated and real-world data, and demonstrate the advantages of this approach for tasks with a long-horizon to: 
a) accurately predict long-term behaviors of stable robot dynamics, and clearly outperforms the one-step models after about 50 time-steps;
b) improve the sample efficiency;
c) accurately predict downstream task reward.
With these benefits, we show how the model can be used in an iterative learning task and planning via model predictive control.
We believe that this approach is a small but significant step towards understanding and overcoming the limitations of the dynamics models in current model-based reinforcement learning algorithms.


\section{RELATED WORKS}
\label{sec:related}

\subsection{Model Learning for Control}
The foundation of trajectory-based models is the context of using predictions for control~\citep{kirk2004optimal} and learning dynamics for unknown systems~\citep{nguyen2011model}. 
In robotics, combining these two directions is the method of data-driven model predictive control (MPC), which directly leverages a dynamics model to choose an action~\citep{shim2003decentralized, wieber2006trajectory, klanvcar2007tracking}. 
In recent years, data-driven approaches for model learning and control have become more prevalent.
Model-based reinforcement learning (MBRL) is the iterative framework of an agent acting to collect data, modeling the transition function with said data, and leveraging the model for control.
MBRL has shown impressive performance and sample efficiency on many robotics tasks~\citep{Deisenroth2011PILCO, williams2017information, chua2018deep, janner2019trust}.
Model-based planning approaches have been successfully using one-step predictions and model predictive control~(MPC) to select best actions~\citep{williams2017information, nagabandi2018learning, lambert2019low}.
This paper proposes a new technique of predicting trajectories for data-driven modeling of unknown dynamics in hopes to improve robotic learning tasks in MBRL, especially with MPC.
System identification~(SI) has a large history for learning models for control, but generally is not formulated as a iterative learning problem~\citep{ljung1999system}.

\subsection{Predicting with Single Steps Ahead}
Single-step dynamics models are effective across many domains, but they suffer from compounding errors.
\citet{chua2018deep} showed excellent downstream control performance with probabilistic neural network~(NN) ensembles, and similar approaches have been applied to many robotic tasks~\citep{williams2017information, lambert2019low}.
Methods mitigating the compounding error of multi-step predictions include re-framing the prediction in the lens of imitation learning~\citep{venkatraman2015improving}, using a multi-step value prediction model~\citep{asadi2019combating}, and leveraging a model with flexible prediction horizon~\citep{xiao2019learning}.
Additionally, single-step Gaussian Processes (GPs) have been applied to multiple lower-dimensional robotic learning tasks~\citep{Deisenroth2011PILCO}.
GP models have been studied in many contexts, such as the long term predictions of pedestrians~\citep{heravi2011long}. 

\subsection{Predicting Trajectories}
Predicting long-term trajectories has a long-history, dating back to autoregressive-moving-average models (ARMAX)~\cite{nguyen2011model}, and has been a growing area of research as robotic complexity outgrew the performance of historical approaches.
A common approach is to modify one-step prediction training to account for correlation across a trajectory with NNs or GPs. 
Specifically, \citet{nar2020learning} proposes a time-weighted loss function to learn unstable state-space systems, but does not apply it to high-dimensional, nonlinear systems.
\citet{lambert2020objective} showed increased correlation between model-accuracy and downstream reward when training with batches sampled from a single-trajectory instead of random transitions, but predictions still suffer from compounding errors.

Other related methods have attempted to create direct, structural links between models and trajectory training data.
With high dimensional images (rather than states) \citet{ke2019learning} uses an auto-regressive, recurrent network to predict observations in a latent state-space.
\citet{doerr2017optimizing} proposes a multi-step Gaussian Process for learning robotic control and the approach is studied further using the correlation between prediction steps in~\citet{hewing2020simulation}. 
\citet{wilson2014using} suggests using a new kernel in GPs to correlate across trajectory data and using simulated model rollouts to predict reward as a prior GP mean function.
These methods all leverage one-step models to create long-term dynamics, which we differentiate from by including time-dependence. 
Other exciting avenues of long-term prediction of dynamical systems are with neural ordinary differential equations~\citep{chen2018neural} and long short-term memory blocks (LSTMs)~\citep{ha2018world}, but both are yet to be successfully applied to prioperceptive (\textit{i.e.,} not vision based) robotics control tasks.
We baseline our method against LSTMs which have implicit time dependance, where our method includes it explicitly in the input.


\section{BACKGROUND}
\label{sec:background}

\textbf{Markov Decision Process} 
In this work, we formulate the prediction problem in the context of a Markov Decision Process (MDP) \citep{bellman1957markovian}.
An MDP is defined by the state $\state \in \mathbb{R}^{d_s}$, the action  $\action \in \mathbb{R}^{d_a}$, the reward function $r(\state,\action)\in \mathbb{R}$, and the transition function $f : \mathbb{R}^{d_s \times d_a} \mapsto \mathbb{R}^{d_s}$.
In this work, we look to learn a parametrized dynamics model $f_\theta(\state, \action)$ to approximate the true transition function.
We consider agents utilizing a control law $\pi(\cdot)$ to act in the environment for repeated trajectories~$\tau_k$. 
Aggregating the tuples of experience, the agent receives a dataset~$\D=\big\{(s_i, a_i, s_{i+1}) \big\}_{i=1}^N$.

\textbf{One-step Dynamics Models}
\label{sec:step}
Given the state~$\state$, and the action~$\action$, one-step dynamics models predicts the next state~$\nextstate$ (or often a delta compared to the current state) as
\begin{equation}
\nextstate=\dynmod_\theta(\state,\action)\,.
\label{eq:onestep}
\end{equation}
For the prediction of trajectories the dynamic model is then recursively applied such that
\begin{equation}
\mdpstate_{\mdptime+\hor}=\dynmod_\theta(\ldots \dynmod_\theta(\dynmod_\theta(\mdpstate_\mdptime,\mdpaction_\mdptime),\mdpaction_{\mdptime+1}) \ldots , \mdpaction_{\mdptime+\hor})\,.
\label{eq:onestepprop}
\end{equation}
For any model used to represent $\dynmod$ (including deep neural networks), there will inevitably be a prediction error~$\epsilon_\mdptime = \|\hat{s}_\mdptime - s_\mdptime\|$. 
Unfortunately, this error is not well studied can grow in a multiplicative manner with each recursion of the function evaluation as 
\begin{equation}
\mdpstate_{\mdptime+\hor}=\dynmod_\theta(\ldots \dynmod_\theta(\dynmod_\theta(\mdpstate_\mdptime,\mdpaction_\mdptime)+\epsilon_t,\mdpaction_{\mdptime+1})+\epsilon_{\mdptime+1} \ldots , \mdpaction_{\mdptime+\hor})+\epsilon_{\mdptime+\hor}\,.
\label{eq:onesteperrorprop}
\end{equation}
Hence, even small numerical errors will ultimately make predictions over long trajectories inaccurate.
This error growth with one-step models is often referred to as the compounding error problem in MBRL.

\textbf{Recurrent Dynamics Models (RNN)}
\label{sec:lstm}
RNN models predict the next state, $\nextstate$, as a function of the current hidden state of the network, $\nextstate=f_\theta(\zeta_t)$. 
The hidden state is outputted from the recurrent nodes of the network: given the current hidden state, $\zeta_t$, state, $\state$, and action, $\action$, the network passes information, as $\zeta_{t+1}=g_\theta(\state,\zeta_t,\action)$.
In this paper, we study the ability of LSTMs, which are often used to mitigate the vanishing gradient problem~\citep{ha2018world}, but they are also difficult to train in practice.


\section{TRAJECTORY-BASED DYNAMICS MODELS}
\label{sec:traj}

\subsection{Prediction Formulation}
We now describe our new trajectory-based dynamics models (which we refer in the rest of the manuscript as $T$) which focus on modeling trajectories over time rather than individual steps. 
The are two main intuitions behind the adoption of this type of model: 1) for control purposes it is often more valuable to have an accurate overall trajectory prediction compared to accurately predict single steps (which might compound error over long-term). 
This is even more important when planning, since for planning the relative ranking of the trajectories is what determines the eventual actions applied by control scheme such as MPC.
 2) trajectories are often strongly correlated in space and time; however, single-step models do not have efficient mechanism to enforce that.

To address the error compounding, an idea would be to replace the recursive call of \eq{eq:onestepprop} with a $n^\text{th}$~step prediction
\begin{equation}
\mdpstate_{\mdptime+\hor}=\dynmod_\theta(\mdpstate_\mdptime,\mdpaction_\mdptime,\mdpaction_{\mdptime+1} \ldots , \mdpaction_{\mdptime+\hor})\,,
\label{eq:nstepprop}
\end{equation}
which does not requires recursion, and is thus more likely to produce stable long-term predictions.
However, here we can observe how the dimensionality of the model to be learned depends on the length of the prediction into the future~$\hor$, and that in addition, this model is generally only capable of predicting the resulting $n^\text{th}$~step ahead prediction, but not its intermediate steps (this is not true for RNNs, but their formulation is more similar to \eq{eq:onestepprop}).
A first variant of this $n^\text{th}$~step ahead formulation would be to observe that the sequence of action $\mdpaction_\mdptime, \mdpaction_{\mdptime+1}, \ldots \mdpaction_{\mdptime+\hor}$ is typically generated by a generic, but single controller $\pi(\cdot)$ determined by parameters $\theta_\pi$, and thus we can rewrite as
\begin{equation}
\mdpstate_{\mdptime+\hor}=\dynmod_\theta(\mdpstate_\mdptime, \theta_\pi)\,.
\end{equation}
As long as the dimensionality of $\theta_\pi$ is smaller than the dimensionality of $\mdpaction_\mdptime, \mdpaction_{\mdptime+1}, \ldots \mdpaction_{\mdptime+\hor}$, this would result in an effective reduction of the dimensionality of our dynamics model and thus improved data-efficiency.
However, once again this model only allows us to predict the final state but not the trajectory that led us there.
Adding the notion of dynamic time-prediction is the final conceptual change to attempt to accurately predict long-term system dynamics in a data-efficient manner, by which we index the starting state at time $t$ and directly predict to the future, variable horizon of $h$ steps with one forward pass. 
The trajectory-based models predict the evolution from a starting state~$\state$, subject to the control parameters~$\theta_\pi$, to a future-time index~$t+h$, as
\begin{equation}
    s_{t+\hor}=\dynmod_\theta(\state,h,\theta_\pi)\,.
    \label{eqn:traj}
\end{equation}
Compared to the traditional recursive one-step ahead formulation of \eq{eq:onestep}, this formulation provides several benefits that we now detail.

\subsection{Benefits of Trajectory-based Models}
\textbf{Data-efficiency}
One advantage of this formulation is that we can perform a re-labeling trick over the dataset of collected trajectories, to significantly augment the dataset used to train the trajectory-based model.
We assume a dataset~$\D$ of $n$ collected trajectories~$\D = \{\tau^n\}$, each of fixed length~$L$.
For each collected trajectory $\tau^j= [s_0, \ldots, s_L]$ we can now extract $L-1$ subtrajectories $\tau_{i}^j=[s_i, \ldots, s_{L}]$ for $i=0 \ldots L-1$, and use them all for training the trajectory-based model.
By training on all sub-trajectories, the model gains two strengths: 
1) it can predict into the future from any state, not just those given as initial states from a environment, and 
2) the number of training points grows proportional to the square of trajectory length, as 
\begin{equation}
    N_{\text{train}}= n \sum_{t=1}^L t = n \frac{(L)(L-1)}{2} \approx n L^2\,.
    \label{eq:pts}
\end{equation}
This results in models that better exploit the temporal structure of the systems, while using less data.

\textbf{Computationally Efficient Planning} 
The trajectory-based models have a useful property of directly predicting entire trajectories instead of imagined roll-outs composed of repeated model evaluations.
For prediction propagation, by only passing in a vector of time horizons $\vec{h}$ from a current state, a planner can evaluate the future with one forward pass, alleviating the computational burden (as well as the compounding, multiplicative error) associated with evaluating sequentially many steps of one-step models.
In our model the predictions in a trajectory do not depend on the prediction at the previous step, which can dramatically increase the control frequency when planning online.

\begin{figure}[t]
    \vspace{6pt}
    \begin{center}
    \small{\cblock{0}{0}{0} True Trajectory (\textcolor[rgb]{.01,.01,.01}{\large{-}}) \quad
    \cblock{200}{190}{0} Probabilistic Traj. Model
    (\textcolor[rgb]{.78,.78,.01}{\large{$\circ$}}) \\
    \cblock{10}{200}{50} One-step Probabilistic Model
    (\textcolor[rgb]{.1,.8,.2}{$\diamondsuit$})
    } 
    \end{center}
    \centering
    \begin{subfigure}[t]{0.48\linewidth}
        \centering
        \includegraphics[width=\linewidth]{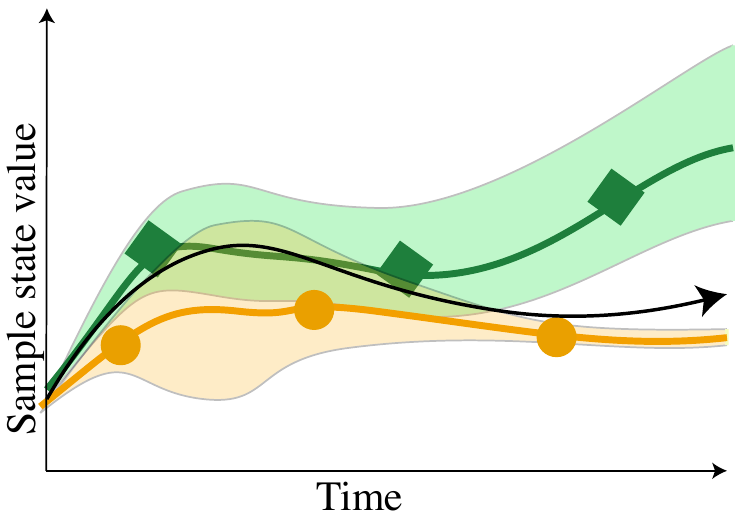}
        \caption{Uncertainty sketch.}    
        \label{fig:sketch-uncert}
    \end{subfigure}
    ~
    \begin{subfigure}[t]{0.48\columnwidth}  
        \centering 
        \includegraphics[width=\linewidth]{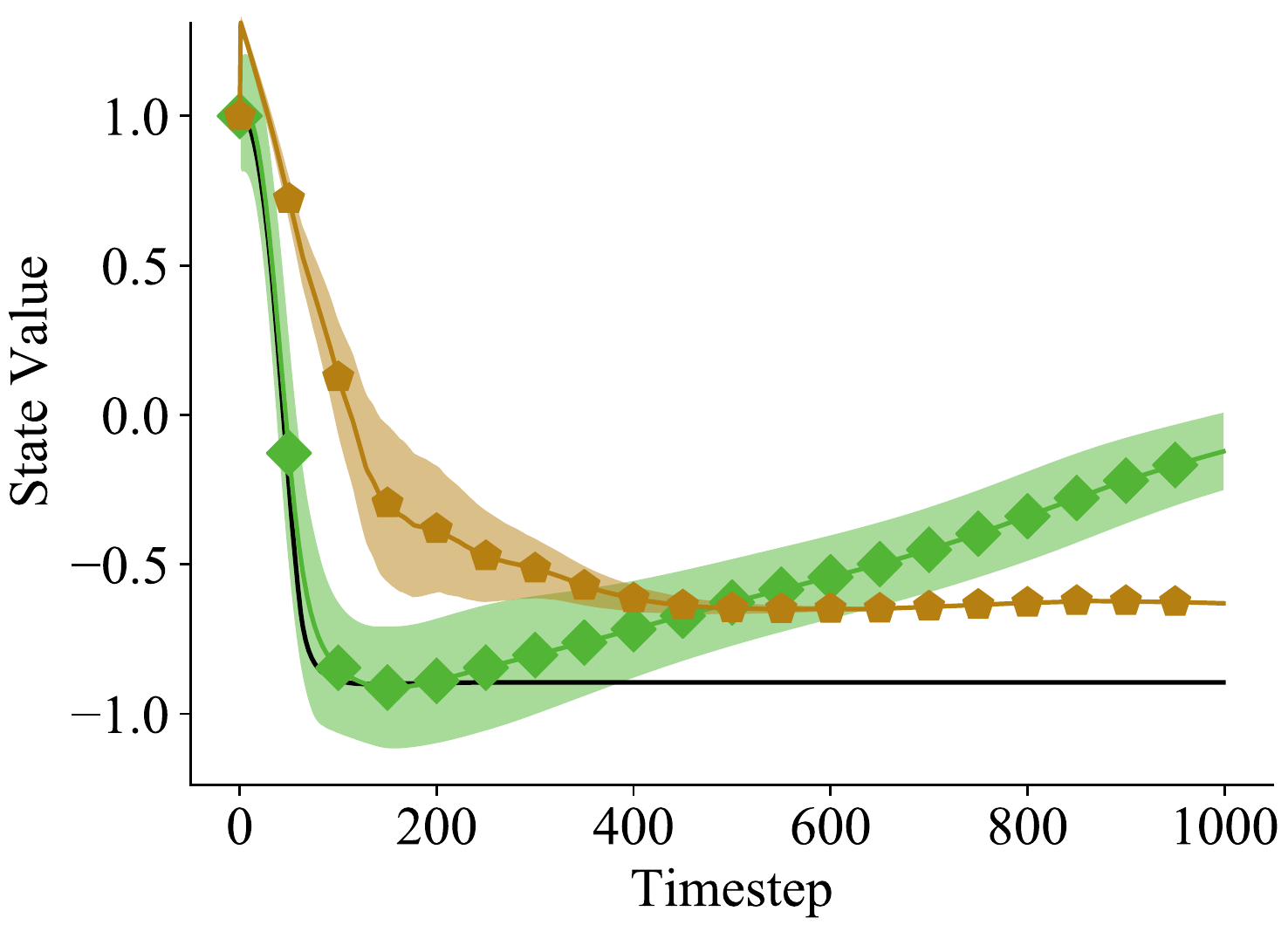}
        \caption{Experimental uncertainty.}    
        \label{fig:exp-uncert}
    \end{subfigure}
    \caption{The trajectory-based models have prediction uncertainty proportional to the epistemic uncertainty in the training dataset.
    (\textit{Left}) a sketch of the uncertainty mechanism. 
    Trajectory-based models have uncertainty that can shrink when more confident in the dynamics, which is opposed to one-step models that have predicted uncertainties that diverge at long horizons.
    (\textit{Right}) an example trial of a robotic prediction (from the reacher task in \sect{sec:exp}) highlights this uncertainty propagation with a probabilistic trajectory-based model and an one-step probabilistic model (\textit{P}).
    }
    \label{fig:sketch2}
\end{figure}

\textbf{Capturing Empirical Distribution over Trajectories}
One-step models commonly suffer from the issue of uncertainty explosion, where the predicted uncertainty over a trajectory typically keep increasing, and does not match the empirical uncertainty from the data.
By propagating time directly, our probabilistic trajectory-based model can instead capture the uncertainty of variation in dynamics in the training set (\textit{i.e.}, the model is more uncertain in areas of rapid movement and can become confident when motion converges), and the empirical uncertainty over trajectories.
The uncertainty propagation is drawn in \fig{fig:sketch-uncert} and an example experiment is shown in \fig{fig:exp-uncert}; both are compared to one-step models that have diverging uncertainty as the predicted states leave the training distribution.
Stable uncertainty estimates convey promise when planning on robotic hardware, where action choices are balanced against model uncertainty due to high cost-per-test.

\textbf{Continuous Time}
Traditional one-step ahead models assume a discrete quantization of time such that the sampling frequency is constant.
Instead, our model is agnostic to the use of discrete or continuous time, since the model can make use of data collected at arbitrary $h$ and explicitly interpolate between them.
While this property is not employed in the following experiments, this is a very desirable property that we aim to explore in future work.


\section{EXPERIMENTAL RESULTS}
\label{sec:results}

We now evaluate the proposed trajectory-based models.
In particular, we investigate the long term prediction accuracy, the ability for the trajectory-based model to predict unstable or periodic data, the sample efficiency benefit of the new parameterization, and using the new model for predicting experimental reward.
Code and an expanded manuscript are available on the website: \website{}.

\subsection{Experimental Setting}
\label{sec:exp}

\begin{figure}[t]
    \vspace{6pt}
    \centering
    \begin{subfigure}[t]{0.32\linewidth}
        \centering
        \includegraphics[width=\linewidth]{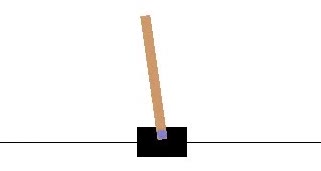}
        \caption{Cartpole.}    
        \label{fig:cp}
    \end{subfigure}
    \hfill
    \begin{subfigure}[t]{0.32\linewidth}  
        \centering 
        \includegraphics[width=\linewidth]{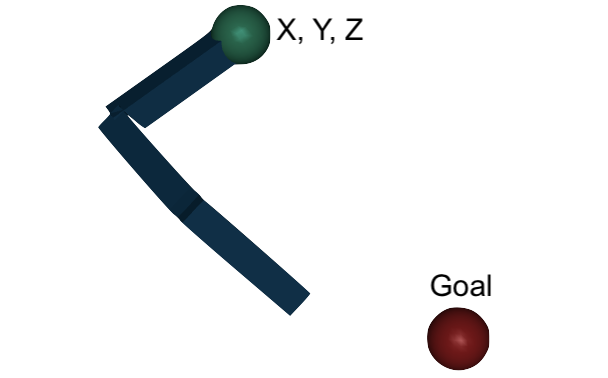}
        \caption{Reacher.}    
        \label{fig:rch}
    \end{subfigure}
    \hfill
    \begin{subfigure}[t]{0.32\linewidth}  
        \centering 
        \includegraphics[width=\linewidth]{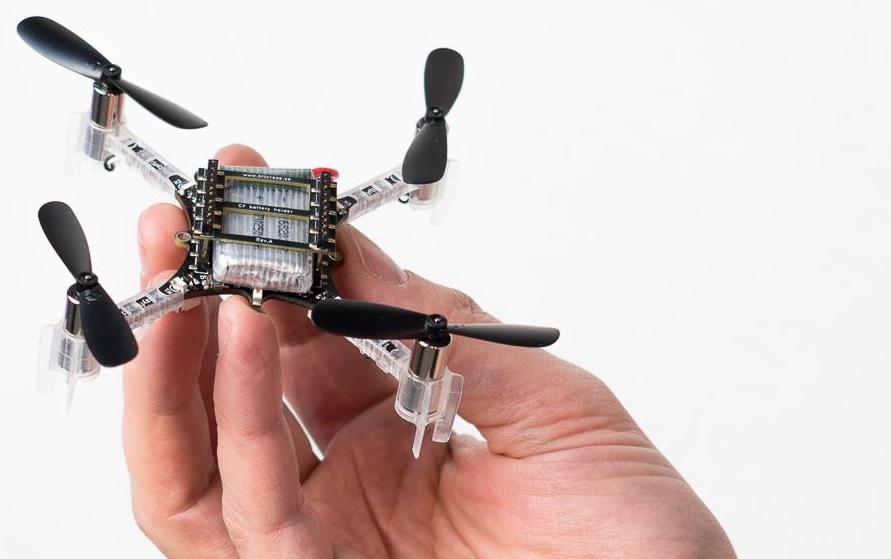}
        \caption{Quadrotor.}    
        \label{fig:quad}
    \end{subfigure}
    \caption{Experimental platforms.}
    \label{fig:robots}
    \vspace{-5pt}
\end{figure}

\textbf{Model Training}
Using the same notation and model training formulations in~\cite{chua2018deep}, we use four model types: $D,P,DE,PE$.
The deterministic model, $D$, and deterministic ensemble, $DE$, minimize the mean squared error (MSE) of predictions.
The probabilistic model, $P$, and probabilistic ensemble, $PE$, minimize the negative log likelihood (NLL) of a Gaussian distribution of state transitions.

All models normalize the input states and actions to a standard normal distribution $\mathcal{N}(0,1)$ in each dimension, and bounded control parameters are mapped to $[-1,1]$.
The feedforward models have two hidden layers of width 250, are optimized with Adam~\citep{adam}, with batch sizes of 32 for \textit{D,P} and 64 for \textit{T}, and learning rates of \num{2.5e-5} for \textit{P}  models, \num{5e-5} for \textit{D} and \num{8e-4} for \textit{T}.
Due to the rapid accruing of labeled data for the trajectory-based models, we cap the training set size at \num{1e5} by random downsampling.
The LSTMs are trained with Adam with a learning rate of \num{0.1}, with batches of sequences matching the trajectory length, $L$, and with normalization following~\citep{sutskever2014sequence}.

\begin{figure}[t]
    \vspace{6pt}
    \centering
    \begin{center}
    \small{\cblock{20}{128}{20} Re-compute $a_t$ from $\hat{s}_t$ (\textcolor[rgb]{.078,.5,.078}{$\mathbb{X}$})
    \cblock{200}{0}{0} Oracle Provides $a_t$ (\textcolor[rgb]{.78,.0,.0}{$\bigplus$})}\end{center} 
    \vspace{-10pt}
    \begin{center}
        \includegraphics[width=0.85\linewidth]{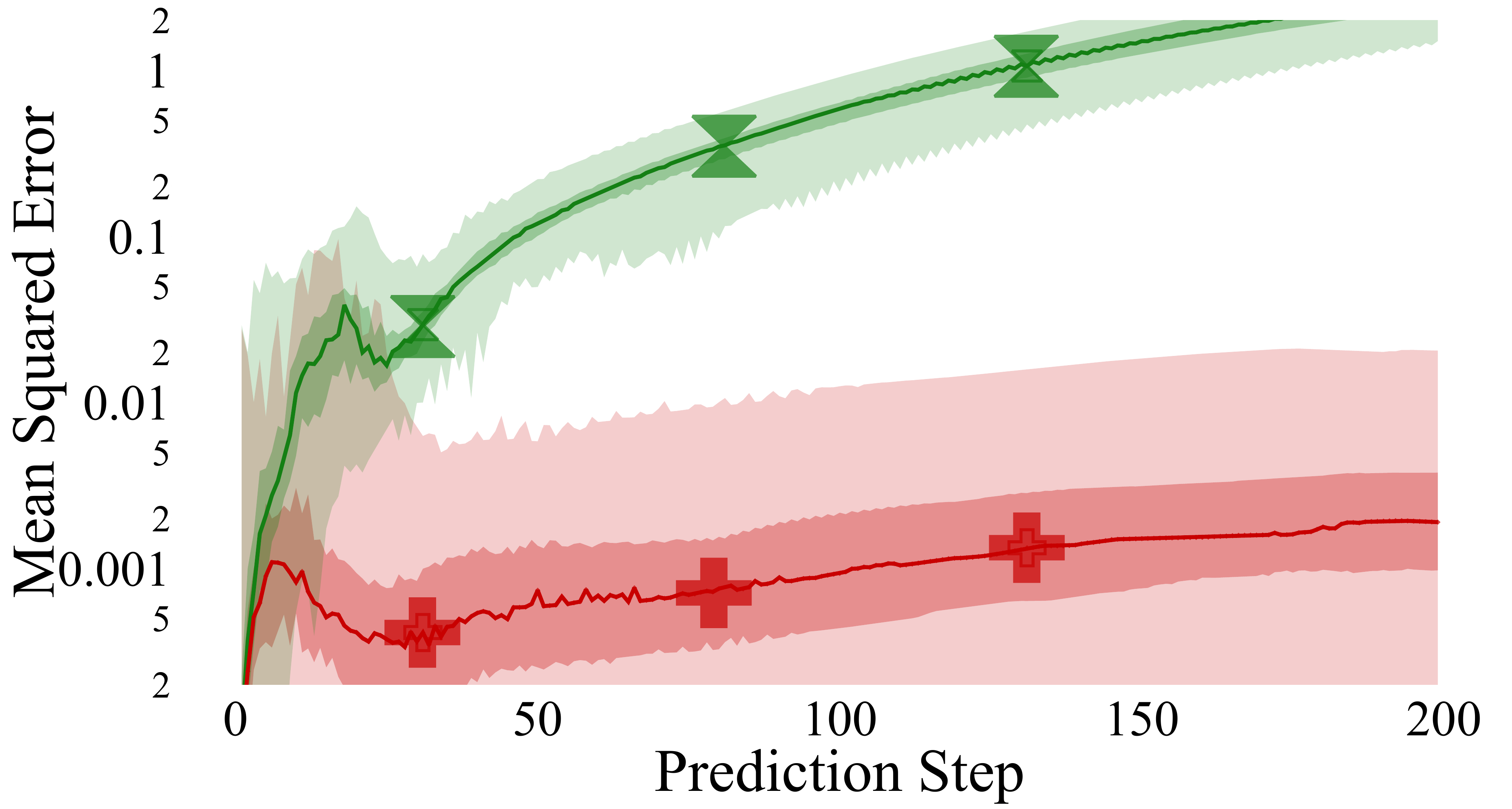}
        \label{fig:step}
    \end{center}
    \vspace{-10pt}

    ~ 
    \caption{ 
    The prediction accuracy of a $PE$ on Cartpole with and without re-computing the action at each step ($N_{\text{traj}} = 100$) shows the increased error from re-computing actions. 
    }
    \label{fig:step}
    \vspace{-5pt}
\end{figure}

\textbf{Cartpole (Simulated)}
We evaluate our models thoroughly on the cartpole task where the goal is to balance a mass over a sliding cart ($d_s=4$, $d_a=1$), shown in \fig{fig:cp}.
To have a continuous reward for prediction, we introduce a new reward function $r(s_t,a_t) = -(x^2 + \theta^2)$ and the remaining details are outlined in~\citet{chua2018deep}.
We evaluate predictions of state and reward of cartpole agents conditioned on a Linear Quadratic Regulator (LQR) control policy.
LQR solves the optimization $\min_u J(u) = \int_{0}^\infty \vec{s}^\top Q \vec{s} + \vec{a}^\top R \vec{a}\  \mathrm{d}t$ for the dynamics $\dot{\vec{s}} = \widetilde A\vec{s}+ \widetilde B\vec{u}$.
LQR control minimizes the expected cost based on a linearized dynamical system, $\tilde{A}, \tilde{B}$.
The control policy, $\pi(\cdot)$, of LQR takes the form of state feedback, $a=-\vec{K}\vec{s}$, where $\vec{K} \in \mathbb{R}^{d_a \times d_s}$.
For all experiments we use the following cost matrices to generate a optimal controller, $\vec{K}^*$: $Q=\operatorname{diag}(.5,\ .05,\ 1,\ .05), \quad R=[1]$, and then sample a random vector $\vec{m} \in \mathbb{R}^{4 \times 1}$ uniformly from the interval $[0.5,1.5]$ to create a variety of controllers $\vec{K}_i=\vec{m}_i\cdot \vec{K}^* \forall i$.

\textbf{Reacher (Simulated)}
For a higher dimensional task, we examine the 5 joint, three-dimensional, reacher manipulation task ($d_s=15$, $d_a=5$) in the Mujoco, OpenAI Gym environment~\citep{todorov2012mujoco, brockman2016openai}, shown in \fig{fig:rch}.
The task associated with the environment is to maneuver the end-effector of the arm from an initial position state to an end position state.
To create a diverse set of data for prediction, our experiments control the agent using a Proportional-Integral-Derivative (PID) controller with randomly generated parameter vectors $\vec{K}\in \mathbb{R}^{15}$.
The parameters of a PID control are defined by a vector of joint angle targets, $\vec{z}_d\in\mathbb{R}^5$, proportional constants, $\vec{K}_p\in\mathbb{R}^5$, integrative constants, $\vec{K}_I\in\mathbb{R}^5$, and derivative constants, $\vec{K}_D\in\mathbb{R}^5$, for each rotatory joint.
We set $\vec{K}_I=0$ for all experiments.
Given the joint angle~$z_i$ and the current error~$e_i=z_i-z_d$, the control command at the $i^{\text{th}}$ joint is $u_i = K_P\cdot e_i + K_D\cdot \dot{e}_i$.

\begin{figure*}[t]
    \begin{center}
    \small{\cblock{0}{0}{200} Deterministic, one-step: \textit{D} (\textcolor[rgb]{.0,.0,.78}{\large{$\circ$}})\quad
    \cblock{200}{0}{0} Trajectory-based: \textit{T}
    (\textcolor[rgb]{.78,.01,.01}{$\bigplus$})
    }
    \end{center}
    \begin{center} 
    \small{
    \cblock{0}{200}{00} Probabilistic, Ensemble one-step: \textit{PE} (\textcolor[rgb]{.0,.78,.0}{\large{$\upbowtie$}})\quad
    \cblock{40}{40}{40} Long Short-term Memory : \textit{LSTM}
    (\textcolor[rgb]{.2,.2,.2}{$\Box$})
    } 
    \end{center}
    \centering
    \begin{subfigure}{0.42\linewidth}
        \centering
        \includegraphics[width=\linewidth]{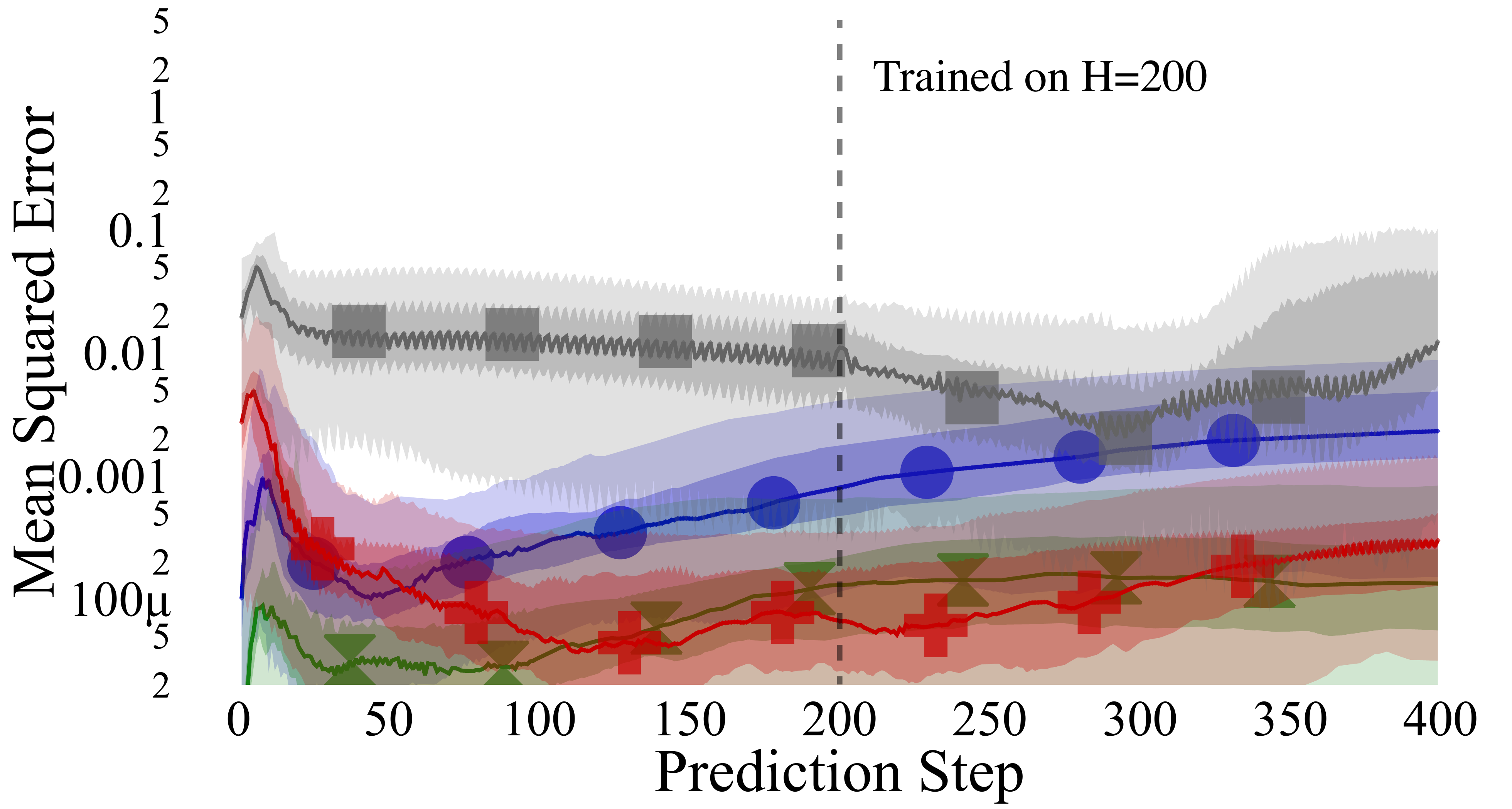}
        \caption{Cartpole (Simulated).}    
        \label{fig:pred-cp}
    \end{subfigure}
    ~ 
    \begin{subfigure}{0.42\linewidth}  
        \centering 
        \includegraphics[width=\linewidth]{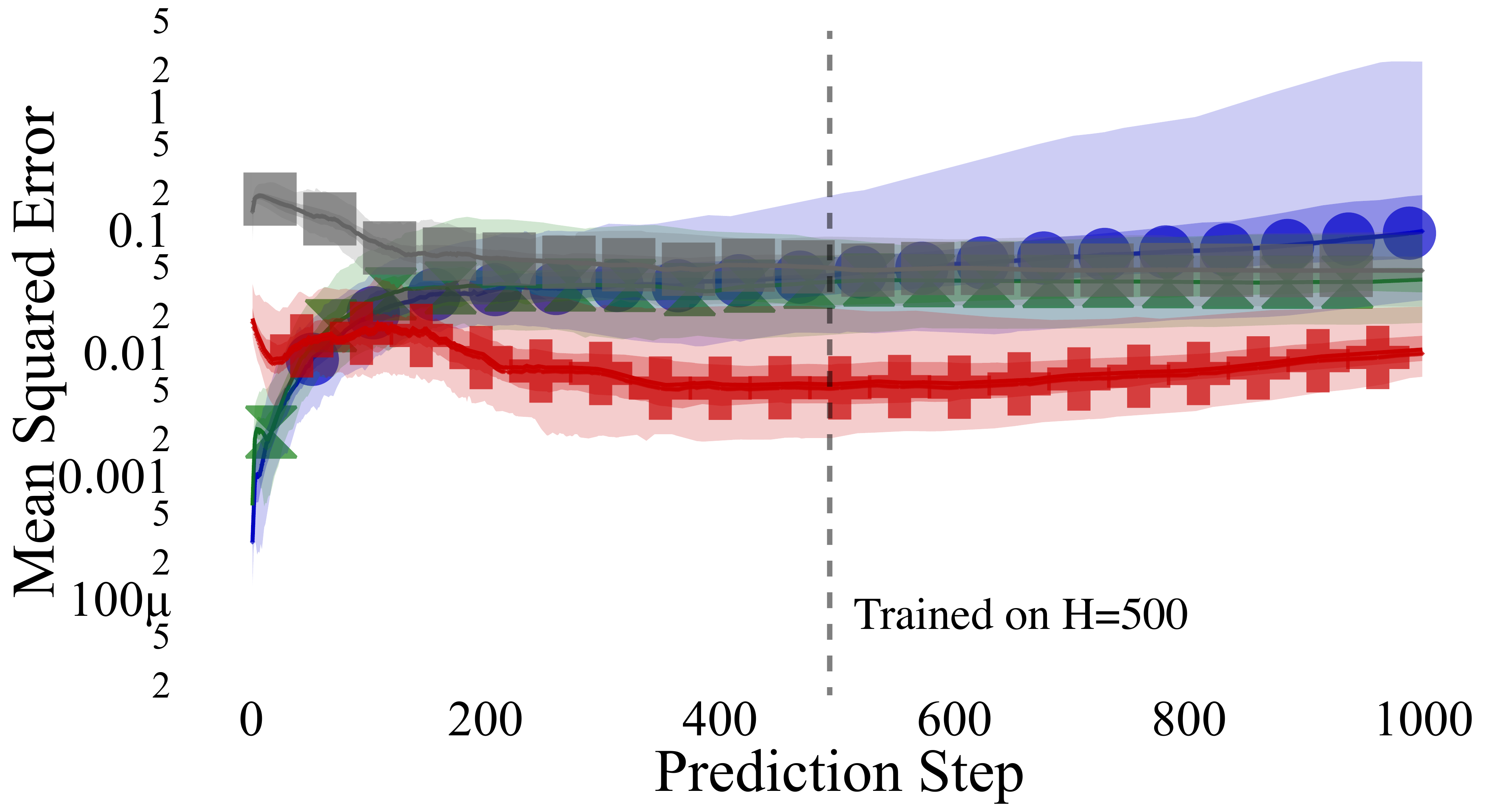}
        \caption{Reacher (Simulated).}    
        \label{fig:pred-rch}
    \end{subfigure}
    \\
    \begin{subfigure}{0.42\linewidth}
        \centering
        \includegraphics[width=\linewidth]{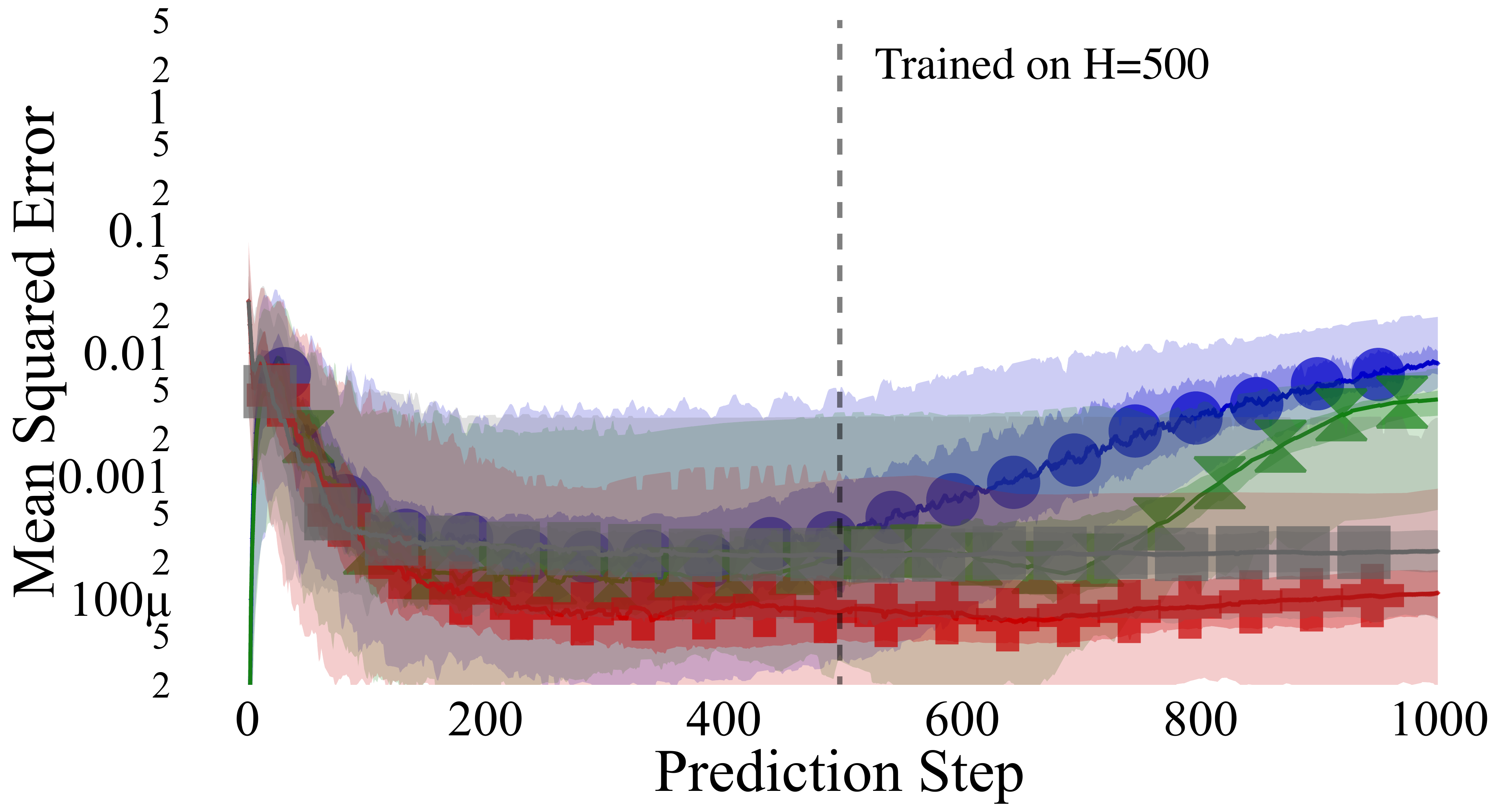}
         \caption{Quadrotor (Simulated)}    
        \label{fig:pred-sim-cf}
    \end{subfigure}
    ~ 
    \begin{subfigure}{0.42\linewidth}  
        \centering 
        \includegraphics[width=\linewidth]{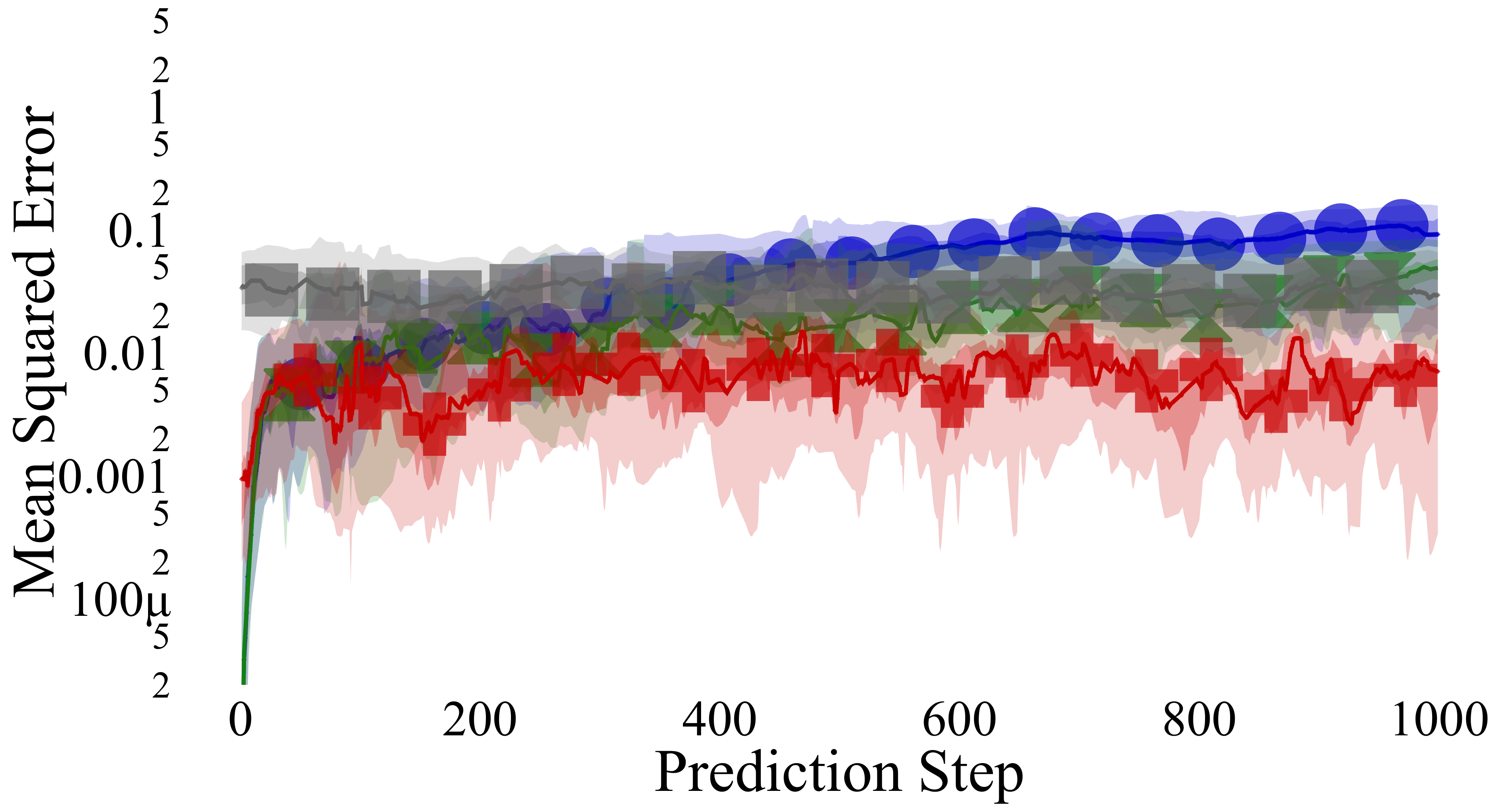}
        \caption{Quadrotor (Real Hardware)}    
        \label{fig:pred-exp-cf}
    \end{subfigure}
    \caption{The log-scale, mean-squared prediction error for the tested environments. 
    The simulated environments are tested on 250 validation trajectories and the experimental data is tested on 16 training trajectories due to experimental restrictions.
    Highlighted is the median error with the $65^{\text{th}}$ and $95^{\text{th}}$ percentile of errors.
    These figures highlight two properties:
    1) $5\times$ to $10\times$ gain in long term prediction accuracy via trajectory-based models for $\hor>50$ and 
    2) the uncertainty in one-step prediction continues to growth with the prediction step while trajectory-based error remains stable.
    The vertical line indicates the length of trajectories in the training distribution.
    }
    \label{fig:predict}
\end{figure*}

\textbf{Quadrotor (Simulated \& Real Hardware)}
We validated the prediction accuracy of trajectory-based models on a simulated and experimental low-level attitude control of a quadrotor ($d_s=5$, $d_a=4$).
The quadrotor model is based off the Crazyflie ~\citep{giernacki2017crazyflie}, shown in \fig{fig:quad}, an \SI{27}{\gram}, open-source micro-aerial vehicle.
The 12 state Euler-step simulation follows \cite{mahony2012multirotor} and has uniform Gaussian noise on all state variables sampled from $\sigma \sim \mathcal{N}(0,0.01)$. 
The simulated controller is a linear, pitch and roll PD controller with randomly sampled parameters.
For experimental data, we collected \SI{180}{s} of aggressive flight data with default PID rate-controllers. 
This data was broken down into trajectories of length 1000 randomly, which we used to validate our prediction mechanism.


\subsection{Long-term Prediction Accuracy}
\label{sec:pred}
We now demonstrate the ability of the trajectory-based models (\textit{T}) to more accurately predict long-horizon robotic dynamics by measuring the mean-squared error of the predicted trajectory versus the measured state, $\sum_{t=1}^H\|\hat{s_t}-s_t\|^2$.
We evaluate the ability to predict horizons of over 100 steps and trajectories longer than the original training distributions. 
An advantage of trajectory-based models over one-step models is that \textit{T} models lack a need to be given a time-series of actions from an oracle or to compute a new action from the current state.
When predicting to long horizons with one-step models, the error can compound and diverge rapidly if the predicted state is used to re-compute the action, shown in \fig{fig:step}. 
For a more competitive baseline in the remainder of our experiments, the one-step models predict the next state given the original action sequences,  $\{a_t\}^L_{t=0}$, and the trajectory-based models given only $\theta_\pi$.

The prediction accuracy for \textit{D}, \textit{PE}, \textit{LSTM}, and \textit{T} models with error $65^{\text{th}}$, $95^{\text{th}}$ percentiles (tested on 100 trajectories) is shown for the cartpole trained on 100 trials of 200 time-steps, \fig{fig:pred-cp}, 100 reacher trials of 500 time-steps, \fig{fig:pred-rch}, 100 simulated quadrotor trials of 500 time-steps, \fig{fig:pred-sim-cf}, and 16 experimental quadrotor trajectories of 1000 time-steps, \fig{fig:pred-exp-cf}. 
The experimental quadrotor trajectories all have the same control parameters, which the model could use to better generalize across trajectories, showing that adding time-dependence alone can improve long-term prediction accuracy.
All states are normalized to a range of $[0,1]$ before computing the MSE to match error across different states (\textit{i.e.}, the scale of a velocity is matched to the scale of an angle).
The trajectory-based models are less accurate for short horizons ($\hor<25)$, but converge to an improvement of up to $10\times$ reduction in MSE for long horizons both in simulation and experiment.
In practice, it is expected that some testing trajectories will extend beyond the expected length, which we evaluate by testing on trajectories of greater length than the training set.
Even with out of distribution time indices, the trajectory-based models maintain their improvement in accuracy over one-step models, removing any need for \textit{T} models to be trained on equal-length trajectories.

\begin{table*}[t]
    \vspace{5pt}
\begin{center}
\begin{tabular}{ l c c c }
\toprule
Prediction Mechanism & Prediction Mapping & Cartpole MSE ($r-\hat{r}$)$\pm \sigma$ & Reacher MSE ($r-\hat{r}$)$\pm \sigma$   \\ 
\midrule 
Direct mapping (GP)     &  $(\vec{K}, s_0) \mapsto \hat{\vec{r}}$  & 0.884 $\pm$ 1.873 & 0.127 $\pm$ 0.192\\
Direct mapping (NN)     &  $(\vec{K}, s_0) \mapsto \hat{\vec{r}}$  & 0.446 $\pm$ 1.189 &  0.074 $\pm$ 0.112\\
One-step; oracle (D)    & $\big(f_\theta(\cdot), \vec{a}_{t=0:L}, s_0 \big) \mapsto \hat{\vec{r}}$ & 0.057 $\pm$ 0.100 & 0.485 $\pm$ 2.048 \\
One-step; pred. actions (D) & $\big(f_\theta(\cdot), \vec{K}, s_0 \big) \mapsto \hat{\vec{r}}$ & 0.979 $\pm$ 1.725 & 0.072 $\pm$ 0.0932 \\
Trajectory-based (T)            & $\big(f_\theta(\cdot), \vec{K}, s_0 \big) \mapsto \hat{\vec{r}}$ & \textbf{0.010 $\pm$ 0.033} & \textbf{0.005 $\pm$ 0.006} \\
\bottomrule
\end{tabular}
\caption{The mean-squared predicted reward error across 100 simulated trajectories show the strength of learning long term dynamics for predicting episode reward.
The 100 trajectories have different initial states~$s_0$, and control parameters~$\vec{K}$. 
The rewards are normalized per the number of trial step -- 200 steps for the cartpole task and 500 for the reacher task.
}
\label{tab:pred_rew}
\end{center}
\end{table*}

\subsection{Accelerated Data Efficiency}
\label{sec:data}
\begin{figure}[t]
    \centering 
    \includegraphics[width=\linewidth]{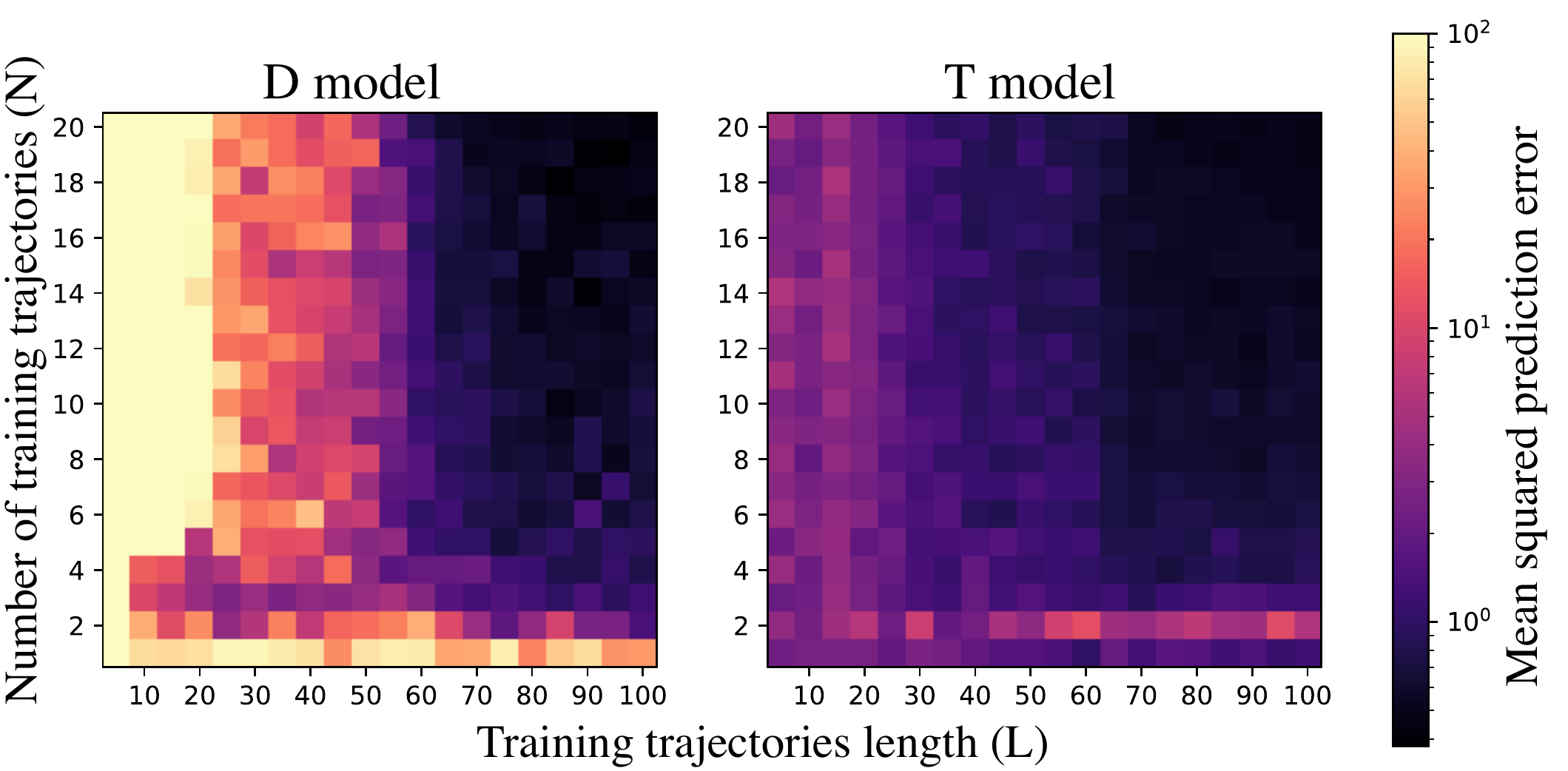}
    \caption{The median, cumulative prediction error of 5 models sweeping over the number $N$ and length $L$ of reacher training trajectories given on a constant validation set of trajectories of length 100.
    The trajectory-based model achieves substantially better prediction accuracy with both shorter and fewer trajectories in the training set.
    In the context of MBRL, the trajectory based model has better \textit{sample efficiency} by having lower cumulative prediction error when trained on fewer trajectories~$N$ (a slice at a specific length~$L$).
    }
    \label{fig:sample}

\end{figure}
Having sufficient labeled data is frequently a limiting factor in deep-learning tasks.
Recall from \sect{sec:traj} that the labelled data for trajectory-based models grows at a rate of the trajectory length squared, $L^2$.
Leveraging this accelerated accruing of data, we evaluate the ability of trajectory-based models to predict accurately in the low-data regime. 
In the reacher environment, we trained models on a grid of trajectory lengths, $L \in [5,100]$, and number of trajectories, $N \in [1,20]$, to predict a validation set of 10 trajectories of length 100. 
Given an initial state, $s_t$, we can predict a set horizon, $\hor$, into the future to obtain a simulated trajectory of states, $\hat{\tau} = \{\hat{s}_i\}_{i=t}^{t+h}$, and measure the mean-squared error (MSE) across all normalized state dimensions.
In \fig{fig:sample}, the trajectory-based models show improved performance for both a) shorter trajectories (L) (another link to predicting beyond the initial training length in \sect{sec:pred}) and b) fewer samples (N).
The regime of low number of training trajectories (N) represents an area of high value to robotic tasks via its potential for reduced  evaluations on a real robot.

\subsection{Predictive Episode Reward}
\label{sec:reward}
Predicting reward is tied to planning actions for robotic systems because if one can accurately predict rewards for an action set,  then one can correctly rank actions.
The trajectory-based models predict rewards in a simulated task by coupling reward prediction to stable long-term predictions.
In this section we compare the predicted reward, $\hat{\vec{r}}$, of an initial state, $\vec{s}_0$, and control parametrization $\vec{K}$ to the true cumulative reward,  $\vec{r} = \sum_{t=0}^L r(s_t)$.

We consider five methods for predicting the reward of an episode from an initial state and control parameters: 
1) the trajectory-based models, 
2) the one-step models given the action sequence from true-states (oracle), 
3) the one-step models with actions computed from predicted states (predicted-actions),  
4) Gaussian Processes (GPs), and
5) neural networks (NN) predicting directly from the initial state and control parameters to sum of reward.
Each candidate method is a different mapping in the following function-space: $h: (\vec{s}_0, \vec{K}, \theta) \mapsto \vec{r}$, where $\theta$ carries different model formulations. 
The dynamics models are used to predict future states via $\hat{s}_{t+1} = f_\theta(\cdot)$, with the one-step models taking in the previous predicted state and the trajectory-based models updating the time index.
The predicted trajectory reward uses the environment-defined reward function $r(s_t,a_t)$ summed over time (the reward functions are action independent). 
A Gaussian process (GP), defined by a mean vector, $\mu_\theta(\vec{x})$, and a covariance matrix, $k(\vec{x}_1,\vec{x}_2)$, or a neural network (NN) can predict the reward with no structured dynamics model. 
for the GP and the NN, the target rewards are normalized uniformly to $[-1,1]$ before prediction to aid in model training.

For both the cartpole and the reacher tasks, the trajectory-based model outperforms other methods in predicted reward accuracy.
The mean-squared predicted reward error across 100 trajectories is shown in \tab{tab:pred_rew}.
We hypothesize that the \textit{structured} learning of a dynamics model accurate across a trajectory improves reward prediction with knowledge of each step over predicting directly to the cumulative reward.

\subsection{Iterative Learning of Control Parameters}
\label{sec:iterative}
\begin{figure}[t]
    \vspace{6pt}
     \begin{center}
    \small{\cblock{0}{0}{200} Bayesian Optimization (\textcolor[rgb]{.0,.0,.78}{$\bigplus$})\ 
    \cblock{200}{0}{0} Trajectory Optimization 
    (\textcolor[rgb]{.8,.05,.05}{\large{$\circ$}})
    }
    \end{center}
    \centering
    \includegraphics[width=0.9\linewidth]{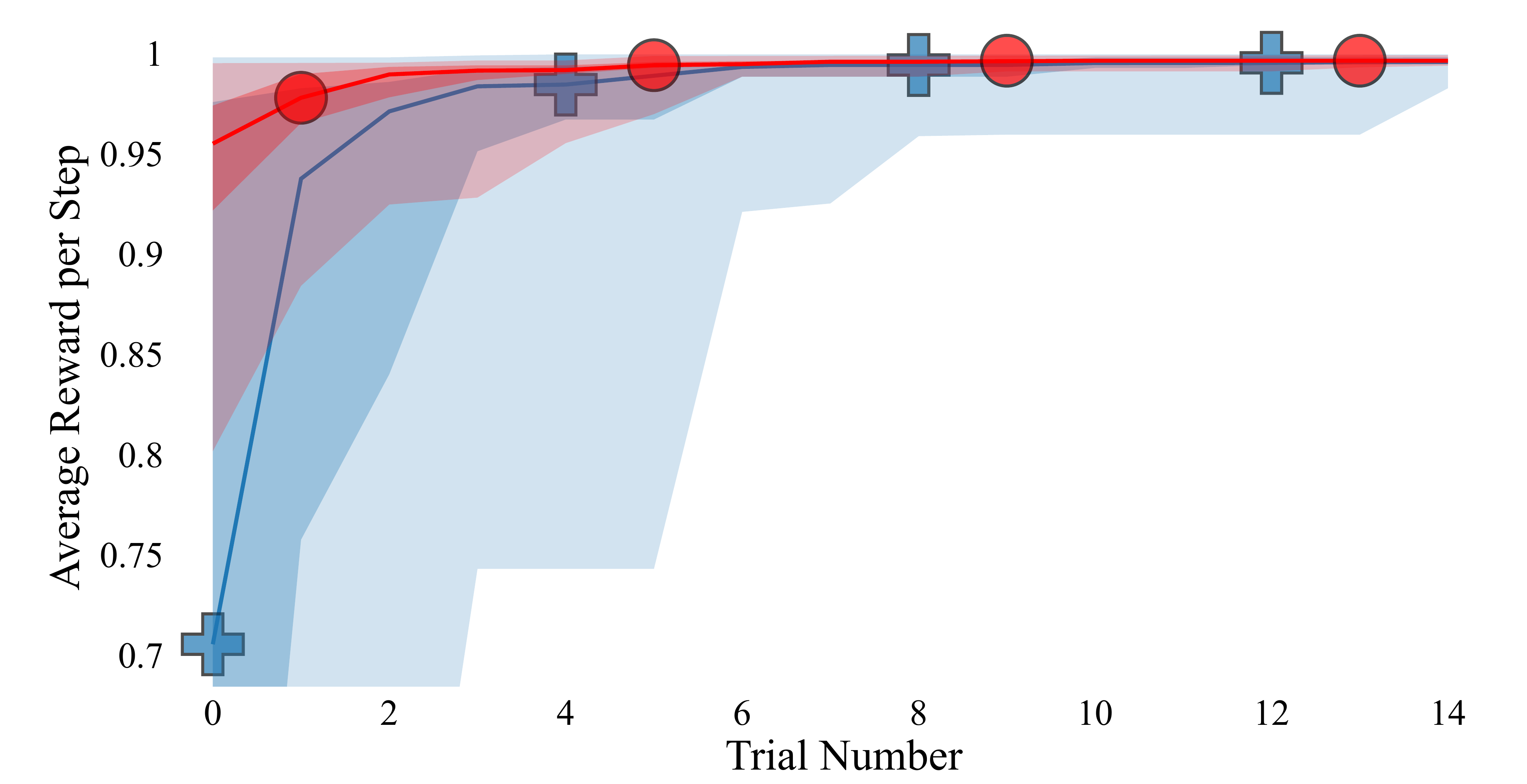}
    \caption{The cumulative maximum reward for iterative learning on cartpole. 
    The Trajectory Optimization with CMA-ES consistently solving the task in 2-4 trials (1 is the maximum normalized reward per step) -- faster than the baseline of Bayesian Optimization.
    The $66^\text{th}$ and $95^\text{th}$ percentiles over 25 trials are shaded.
    }
    \label{fig:iterative-param}
\end{figure}

We now compare the trajectory-based model against black-box optimization algorithms for the iterative learning of control parameters.
At each iteration, we retrain the trajectory-based model and then use it to simulate rollouts of different control parameters. 
Finally, we execute the best control parameters found on the real system, and a new iteration starts.
Ideally, a more accurate dynamics model over the trajectory will result in faster convergence and better performance.
In our experiments, we indicate this approach as \textit{Trajectory Optimization} and use covariance matrix adaptation evolution strategy (CMA-ES) to optimize the simulated rollouts.
However, CMA-ES can be replaced with any optimizer that does not require gradients on the reward function.
Trajectory Optimization generates new control parameters within the population of CMA-ES and simulates a trajectory of the trial length 200 for cartpole.
The simulated reward $\hat{r}$ is the sum over the predicted states, as in \sect{sec:reward}.
We compare this approach to the data-efficient black-box optimization algorithm Bayesian Optimization (BO) which iteratively optimize the control parameters without knowledge of the dynamics~\citep{marco2016automatic,calandra2016bayesian}.
The results of the experiments in \fig{fig:iterative-param} show that on the toy cartpole benchmark task Trajectory Optimization converge to excellent performance faster than Bayesian Optimization. 
This demonstrates that we can learn in an iterative manner trajectory dynamics model, and that they can successfully be applied for control. 

\subsection{Model Predictive Control with Trajectory-based Models}
\label{sec:mpc}
We now demonstrate how to use the trajectory-based model in a common control architecture -- model predictive control (MPC)~\citep{shim2003decentralized, wieber2006trajectory}.
MPC is a common tool for model-based reinforcement learning~\cite{chua2018deep,williams2017information,nagabandi2018learning}, and originated in the study of optimal control leveraging predictions to make decisions \cite{garcia1989model, kirk2004optimal}. 
MPC with learned one-step dynamics models is formulated as
\begin{align}
    a_t^* & = \argmax_{u_{t:t+\tau}} \sum_{i=0}^{\tau} r(\hat{s}_{t+i}, a_{t+i}),  \ \   \hat{s}_{t+1}   = f_\theta(\hat{s}_t,a_t).
    \label{eq:mpc}
\end{align}
\begin{figure}[t]
    \vspace{6pt}
    \centering 
    \includegraphics[width=0.8\linewidth]{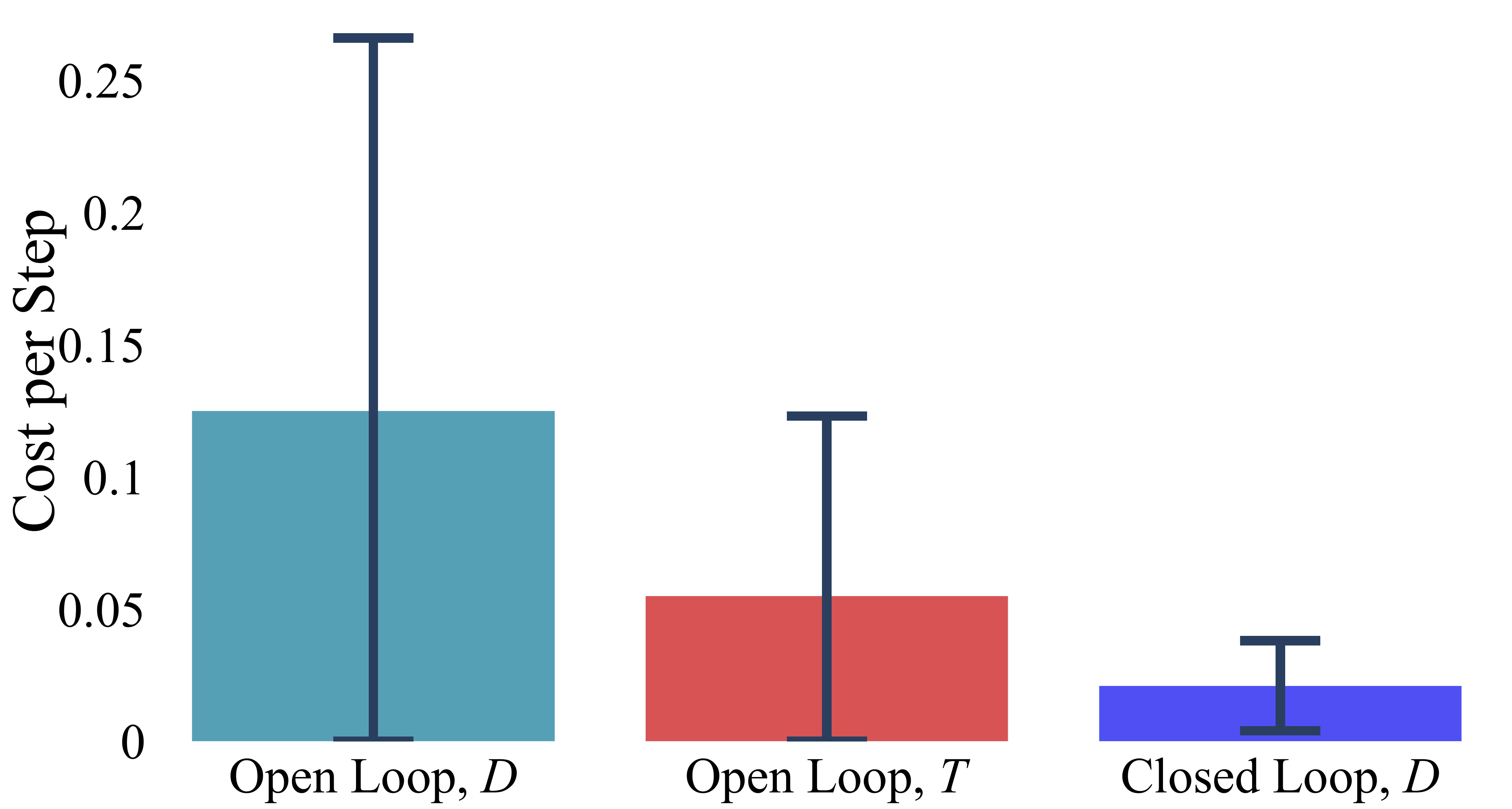}
    \caption{Comparison between trajectory-based and one-step ahead models planning from the initial state (i.e., open-loop) and the one-step model re-planning after each step (i.e., closed loop).
The trajectory-based planning improves performance compared to the one-step ahead in open-loop. 
However, the one-step model used in closed-loop performs the best, at the expense of increased computational complexity for re-planning.
The mean and standard deviation of the reward per step in the cartpole task are reported for 25 trials.}        
    \label{fig:planning}
\end{figure}
With the trajectory-based model, the control formulation needs a modification in how the candidate solutions are selected.
Sampling over control policies and computing the action from the current state, the new MPC formulation is
\begin{align}
    \quad \theta_{\pi,t}^* & = \argmax_{\theta_{\pi,{t:t+\tau}}} \sum_{i=0}^{\tau} r(\hat{s}_{t+i}, a_{t+i})\,, \\
     & \quad   \hat{s}_{t+\tau}   = f_\theta(s_t,\theta_{\pi,t},t+\tau)\,,  
    \quad  a_t^* = \theta_\pi^* (t)\,.
    \label{eq:mpc_traj}
\end{align}
MPC is known to be computationally intensive -- where some robots lack the computing infrastructure to run recent MBRL methods online -- so we compare leveraging the trajectory-based optimization only from the first state, and running that policy through the remaining of the trial.
This is a starting point, and re-planning online with varying update frequencies (holding the chosen policy for $T$ steps) would allow flexibility in control.

As a comparison to iterative learning of one set of control parameters, as in \sect{sec:iterative}, we compare the performance of MPC to that if the optimization is only run on the first time-step.
In this case we maintain random sampling to mirror common applications of MPC in MBRL, while the Trajectory Optimization in \sect{sec:iterative} leveraged the CMA-ES optimizer.
The preliminary results for the planning methods on the cartpole task are shown in \fig{fig:planning}. 
The trajectory-based model is limited by aggregating labelled data with MPC because it requires the sub-trajectories that it is labelled on to have constant control parameters. 
This can be partially overcome by re-planning at a lower frequency, but is an important direction for future work to better integrate the new models into existing MBRL literature.



\section{Discussion and Future Work}
\label{sec:disc}
There is \textit{no free lunch} with respect to numerical dynamics modelling.
The trajectory-based model's accurate long-term prediction and rapid data accruing is valuable to data-driven methods with parametrized controllers. 
One-step models will likely remain useful for other algorithms less focused on the long-term future, such as MBPO~\cite{janner2019trust}, or in situations where data is non-episodic, so applying the time-dependant structure could be forced. 
Another limitation of the trajectory-based model relative to the entire space of MRBL is that this model is designed for scenarios where the generalization across similar trajectories is useful, likely limiting the potential generalization of the model to one type of task instead of an entire state space.
By setting training structures and input-output pairs, any dynamics modeling algorithm is prioritizing its model capacity on a certain task.
%
Finally, our approach was demonstrated mainly on modeling low-dimensional parametrized controllers.
Future work will focus on evaluating and extending this approach with high-dimensional neural network policies, as well as applying it in online model-based reinforcement learning scenarios, including real-world hardware.

\section{Conclusion}
\label{sec:conclusion}
While one-step models have been successful across numerous robotic applications, this paper re-evaluates the paradigm of one-step models to better match the prediction mechanism with that of predicting and controlling trajectories.
In this paper, we introduce a new framework for the learning of trajectory-based dynamics for iterative tasks.
Trajectory-based models have some limitations, such as the need to access to policy parameters instead of just actions that can be gathered with multiple controllers.
However, our trajectory-based models also present several theoretical advantages, such as data-efficiency, computational cost, parallelization, continuous time predictions, and better handling of uncertainty.
The experimental results demonstrate strong performance in long-horizon prediction for simulated and real-world systems.
We believe that better understanding of trajectory-based models is an important step to overcome the current limitations of MBRL.








\bibliographystyle{IEEEtran}
\bibliography{main}

\begin{thebibliography}{10}
\providecommand{\url}[1]{#1}
\csname url@samestyle\endcsname
\providecommand{\newblock}{\relax}
\providecommand{\bibinfo}[2]{#2}
\providecommand{\BIBentrySTDinterwordspacing}{\spaceskip=0pt\relax}
\providecommand{\BIBentryALTinterwordstretchfactor}{4}
\providecommand{\BIBentryALTinterwordspacing}{\spaceskip=\fontdimen2\font plus
\BIBentryALTinterwordstretchfactor\fontdimen3\font minus
  \fontdimen4\font\relax}
\providecommand{\BIBforeignlanguage}[2]{{%
\expandafter\ifx\csname l@#1\endcsname\relax
\typeout{** WARNING: IEEEtran.bst: No hyphenation pattern has been}%
\typeout{** loaded for the language `#1'. Using the pattern for}%
\typeout{** the default language instead.}%
\else
\language=\csname l@#1\endcsname
\fi
#2}}
\providecommand{\BIBdecl}{\relax}
\BIBdecl

\bibitem{williams2017information}
G.~Williams, N.~Wagener, B.~Goldfain, P.~Drews, J.~M. Rehg, B.~Boots, and E.~A.
  Theodorou, ``Information theoretic mpc for model-based reinforcement
  learning,'' in \emph{International Conference on Robotics and Automation},
  2017, pp. 1714--1721.

\bibitem{chua2018deep}
K.~Chua, R.~Calandra, R.~McAllister, and S.~Levine, ``Deep reinforcement
  learning in a handful of trials using probabilistic dynamics models,'' in
  \emph{Neural Information Processing Systems}, 2018.

\bibitem{schrittwieser2020mastering}
J.~Schrittwieser, I.~Antonoglou, T.~Hubert, K.~Simonyan, L.~Sifre, S.~Schmitt,
  A.~Guez, E.~Lockhart, D.~Hassabis, T.~Graepel \emph{et~al.}, ``Mastering
  atari, go, chess and shogi by planning with a learned model,'' \emph{Nature},
  vol. 588, no. 7839, pp. 604--609, 2020.

\bibitem{garcia1989model}
C.~E. Garcia, D.~M. Prett, and M.~Morari, ``Model predictive control: theory
  and practice—a survey,'' \emph{Automatica}, vol.~25, no.~3, 1989.

\bibitem{kirk2004optimal}
D.~E. Kirk, \emph{Optimal control theory: an introduction}.\hskip 1em plus
  0.5em minus 0.4em\relax Courier Corporation, 2004.

\bibitem{asadi2019combating}
K.~Asadi, D.~Misra, S.~Kim, and M.~L. Littman, ``Combating the
  compounding-error problem with a multi-step model,'' \emph{arXiv preprint
  arXiv:1905.13320}, 2019.

\bibitem{xiao2019learning}
C.~Xiao, Y.~Wu, C.~Ma, D.~Schuurmans, and M.~M{\"u}ller, ``Learning to combat
  compounding-error in model-based reinforcement learning,'' \emph{arXiv
  preprint arXiv:1912.11206}, 2019.

\bibitem{lambert2020objective}
N.~Lambert, B.~Amos, O.~Yadan, and R.~Calandra, ``Objective mismatch in
  model-based reinforcement learning,'' \emph{Learning for Dynamics and Control
  (L4DC)}, pp. 761--770, 2020.

\bibitem{nguyen2011model}
D.~Nguyen-Tuong and J.~Peters, ``Model learning for robot control: a survey,''
  \emph{Cognitive processing}, vol.~12, no.~4, pp. 319--340, 2011.

\bibitem{shim2003decentralized}
D.~H. Shim, H.~J. Kim, and S.~Sastry, ``Decentralized nonlinear model
  predictive control of multiple flying robots,'' in \emph{IEEE International
  Conference on Decision and Control}, vol.~4, 2003, pp. 3621--3626.

\bibitem{wieber2006trajectory}
P.-B. Wieber, ``Trajectory free linear model predictive control for stable
  walking in the presence of strong perturbations,'' in \emph{IEEE-RAS
  International Conference on Humanoid Robots}, 2006, pp. 137--142.

\bibitem{klanvcar2007tracking}
G.~Klan{\v{c}}ar and I.~{\v{S}}krjanc, ``Tracking-error model-based predictive
  control for mobile robots in real time,'' \emph{Robotics and autonomous
  systems}, vol.~55, no.~6, pp. 460--469, 2007.

\bibitem{Deisenroth2011PILCO}
M.~P. Deisenroth and C.~E. Rasmussen, ``{PILCO: A Model-Based and
  Data-Efficient Approach to Policy Search},'' in \emph{International
  Conference on Machine Learning}, 2011, pp. 465--472.

\bibitem{janner2019trust}
M.~Janner, J.~Fu, M.~Zhang, and S.~Levine, ``When to trust your model:
  Model-based policy optimization,'' in \emph{Neural Information Processing
  Systems}, 2019, pp. 12\,498--12\,509.

\bibitem{nagabandi2018learning}
A.~Nagabandi, G.~Yang, T.~Asmar, R.~Pandya, G.~Kahn, S.~Levine, and R.~S.
  Fearing, ``Learning image-conditioned dynamics models for control of
  underactuated legged millirobots,'' in \emph{IEEE International Conference on
  Intelligent Robots and Systems}, 2018, pp. 4606--4613.

\bibitem{lambert2019low}
N.~O. Lambert, D.~S. Drew, J.~Yaconelli, S.~Levine, R.~Calandra, and K.~S.
  Pister, ``Low-level control of a quadrotor with deep model-based
  reinforcement learning,'' \emph{IEEE Robotics and Automation Letters},
  vol.~4, no.~4, pp. 4224--4230, 2019.

\bibitem{ljung1999system}
L.~Ljung, ``System identification,'' \emph{Wiley encyclopedia of electrical and
  electronics engineering}, pp. 1--19, 1999.

\bibitem{venkatraman2015improving}
A.~Venkatraman, M.~Hebert, and J.~A. Bagnell, ``Improving multi-step prediction
  of learned time series models,'' in \emph{AAAI Conference on Artificial
  Intelligence}, 2015.

\bibitem{heravi2011long}
E.~J. Heravi and S.~Khanmohammadi, ``Long term trajectory prediction of moving
  objects using gaussian process,'' in \emph{IEEE International Conference on
  Robot, Vision and Signal Processing}, 2011.

\bibitem{nar2020learning}
K.~Nar, Y.~Xue, and A.~M. Dai, ``Learning unstable dynamical systems with
  time-weighted logarithmic loss,'' \emph{arXiv preprint arXiv:2007.05189},
  2020.

\bibitem{ke2019learning}
N.~R. Ke, A.~Singh, A.~Touati, A.~Goyal, Y.~Bengio, D.~Parikh, and D.~Batra,
  ``Learning dynamics model in reinforcement learning by incorporating the long
  term future,'' \emph{arXiv preprint arXiv:1903.01599}, 2019.

\bibitem{doerr2017optimizing}
A.~Doerr, C.~Daniel, D.~Nguyen-Tuong, A.~Marco, S.~Schaal, T.~Marc, and
  S.~Trimpe, ``Optimizing long-term predictions for model-based policy
  search,'' in \emph{Conference on Robot Learning}, 2017, pp. 227--238.

\bibitem{hewing2020simulation}
L.~Hewing, E.~Arcari, L.~P. Fr{\"o}hlich, and M.~N. Zeilinger, ``On simulation
  and trajectory prediction with gaussian process dynamics,'' in \emph{Learning
  for Dynamics and Control}, 2020, pp. 424--434.

\bibitem{wilson2014using}
A.~Wilson, A.~Fern, and P.~Tadepalli, ``Using trajectory data to improve
  bayesian optimization for reinforcement learning,'' \emph{The Journal of
  Machine Learning Research}, vol.~15, no.~1, pp. 253--282, 2014.

\bibitem{chen2018neural}
R.~T. Chen, Y.~Rubanova, J.~Bettencourt, and D.~K. Duvenaud, ``Neural ordinary
  differential equations,'' in \emph{Neural Information Processing Systems},
  2018, pp. 6571--6583.

\bibitem{ha2018world}
D.~Ha and J.~Schmidhuber, ``World models,'' \emph{arXiv preprint
  arXiv:1803.10122}, 2018.

\bibitem{bellman1957markovian}
R.~Bellman, ``A {Markovian} decision process,'' \emph{Journal of mathematics
  and mechanics}, pp. 679--684, 1957.

\bibitem{adam}
D.~Kingma and J.~Ba, ``Adam: A method for stochastic optimization,''
  \emph{International Conference on Learning Representations}, 12 2014.

\bibitem{sutskever2014sequence}
I.~Sutskever, O.~Vinyals, and Q.~V. Le, ``Sequence to sequence learning with
  neural networks,'' in \emph{Neural Information Processing Systems}, 2014, pp.
  3104--3112.

\bibitem{todorov2012mujoco}
E.~Todorov, T.~Erez, and Y.~Tassa, ``Mujoco: A physics engine for model-based
  control,'' in \emph{IEEE International Conference on Intelligent Robots and
  Systems}, 2012, pp. 5026--5033.

\bibitem{brockman2016openai}
G.~Brockman, V.~Cheung, L.~Pettersson, J.~Schneider, J.~Schulman, J.~Tang, and
  W.~Zaremba, ``Openai gym,'' \emph{arXiv preprint arXiv:1606.01540}, 2016.

\bibitem{giernacki2017crazyflie}
W.~Giernacki, M.~Skwierczy{\'n}ski, W.~Witwicki, P.~Wro{\'n}ski, and
  P.~Kozierski, ``Crazyflie 2.0 quadrotor as a platform for research and
  education in robotics and control engineering,'' in \emph{IEEE International
  Conference on Methods and Models in Automation and Robotics (MMAR)}, 2017,
  pp. 37--42.

\bibitem{mahony2012multirotor}
R.~Mahony, V.~Kumar, and P.~Corke, ``Multirotor aerial vehicles: Modeling,
  estimation, and control of quadrotor,'' \emph{IEEE Robotics and Automation
  magazine}, vol.~19, no.~3, pp. 20--32, 2012.

\bibitem{marco2016automatic}
A.~Marco, P.~Hennig, J.~Bohg, S.~Schaal, and S.~Trimpe, ``Automatic {LQR}
  tuning based on gaussian process global optimization,'' in
  \emph{International Conference on Robotics and Automation}, 2016, pp.
  270--277.

\bibitem{calandra2016bayesian}
R.~Calandra, A.~Seyfarth, J.~Peters, and M.~P. Deisenroth, ``Bayesian
  optimization for learning gaits under uncertainty,'' \emph{Annals of
  Mathematics and Artificial Intelligence}, vol.~76, no. 1-2, pp. 5--23, 2016.

\end{thebibliography}

\clearpage

\clearpage
\newpage

\section{Appendix}





\subsection{Predicting With All Models}
\begin{figure}[h]
    \vspace{-10pt}
        \begin{center}
        \small{\cblock{0}{0}{200} D (\textcolor[rgb]{.0,.0,.78}{\large{$\circ$}})\ 
        \cblock{20}{128}{20} PE (\textcolor[rgb]{.078,.50,.078}{$\diamondsuit$}) \
        \cblock{200}{0}{0} T
        (\cblock{200}{0}{0})
        \cblock{0}{0}{0} True Trajectory (\textcolor[rgb]{.01,.01,.01}{\large{-}})}
        \end{center}
    \begin{center}
        \includegraphics[width=\linewidth]{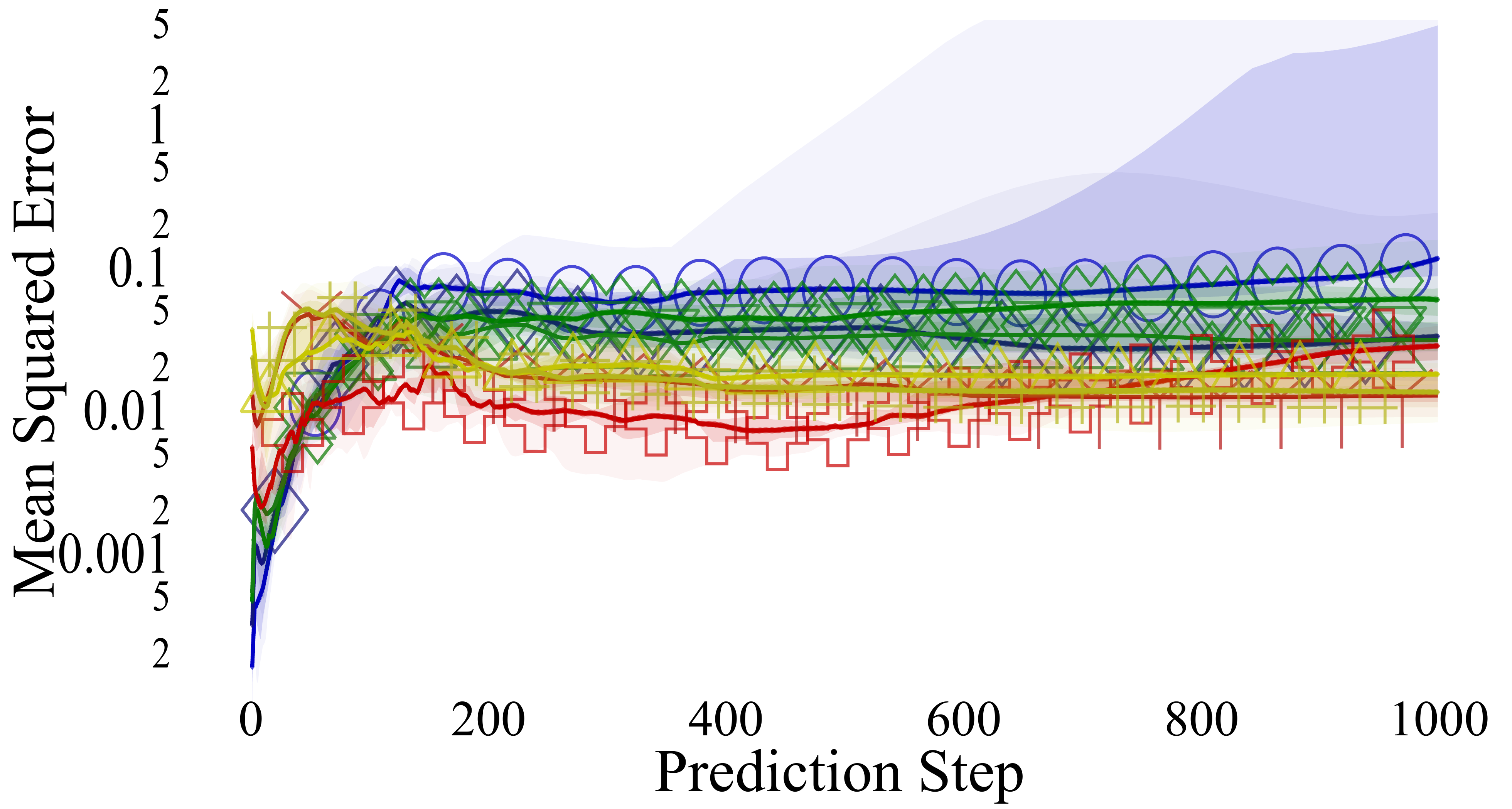}
        \caption{Median prediction error for the 8 model types shown in \tab{tab:models} on Reacher task.
        The trajectory models (red, yellow) perform better than one step-models (blue, purple) for long term predictions.
        }
        \label{fig:all_model}
    \end{center}
    \vspace{-10pt}
\end{figure}

A summary of the explored model-configurations is shown in \tab{tab:models}. 
In the cartpole and the reacher environments with one-step models there are trends of probabilistic models being more accurate then deterministic and ensembles being more accurate than single models (PE $>$ P $>$ DE $>$ D).
For the trajectory-based models, the advantage of ensembling remains, but the probabilistic models do not have a significant improvement over deterministic models (PE $>$ P, DE $>$ D, P $\approx$ D), potentially due to the better uncertainty management in the trajectory-based formulation.
An example with the median error for all 8 model types is shown in \fig{fig:all_model}. 
\begin{table}[h]
\begin{center}
\begin{tabular}{ |r|p{1cm}|p{1cm}|p{1.2cm}|p{1.2cm}|  }
\hline
& \multicolumn{2}{|c|}{Current Models} & \multicolumn{2}{|c|}{New Models} \\
 \hline
Single & D   & P    &T&   TP\\
 \hline
Ensembles & DE &   PE  & TE   &TPE\\
 \hline
\end{tabular}
\caption{\textit{left}: models used in various robotic learning tasks to capture different types of predictive uncertainty~\cite{chua2018deep}; \textit{right}: the new Trajectory-based models with different training variants we propose in \sect{sec:traj}.
}
\label{tab:models}
\end{center}
\end{table}

\subsection{Model Accuracy Versus Epoch}
\begin{figure}[h]
    \vspace{-10pt}
\begin{center}
        \includegraphics[width=\linewidth]{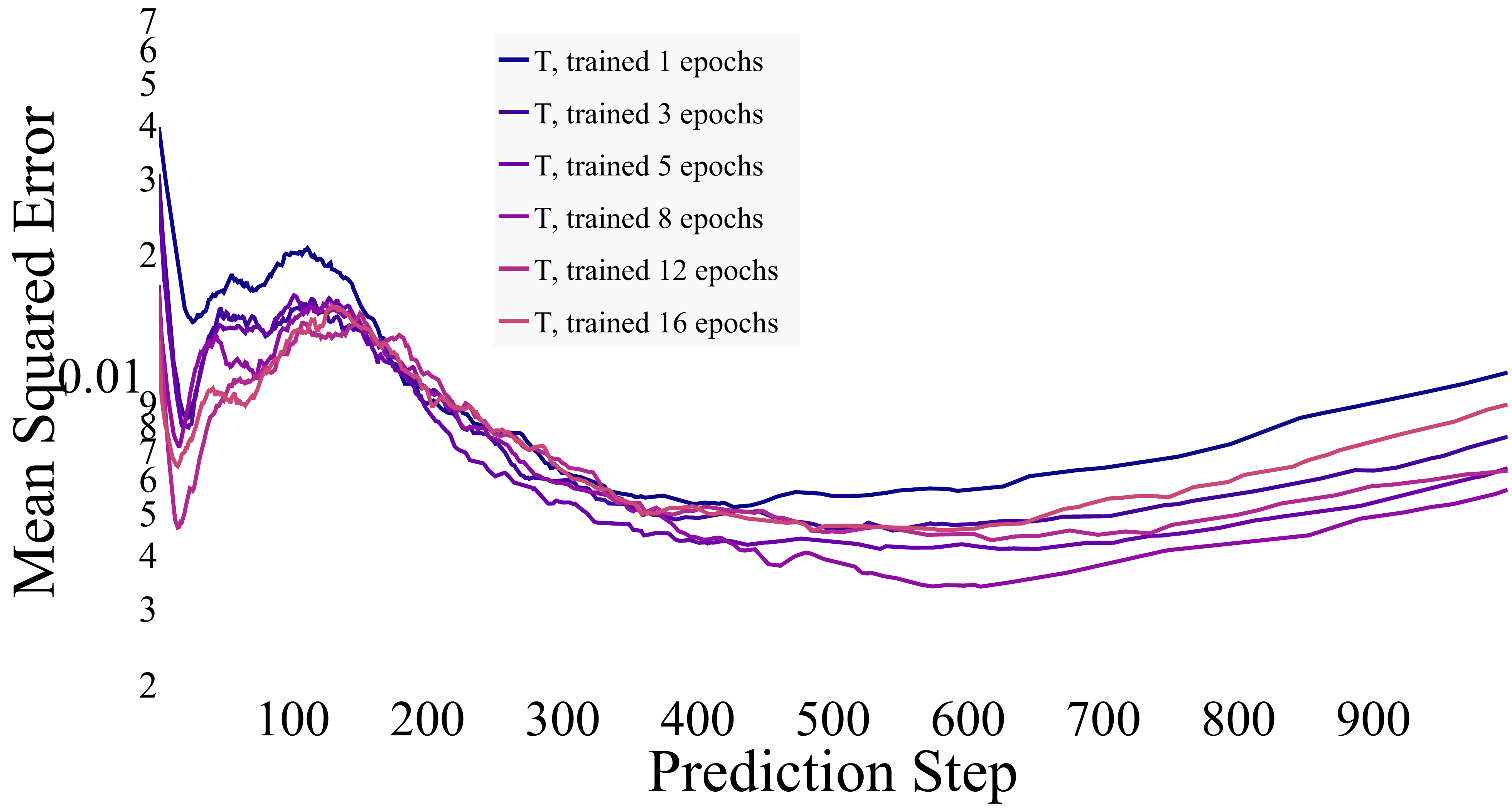}
        \caption{The prediction accuracy of the trajectory-based model when trained for different numbers of epochs, and evaluated on the test set used in \fig{fig:predict}.
        Note, the y-axis scale is changed to show the slight change in performance when tuning these models.
        The mean error per step is shown in \tab{tab:epoch_error}
        }
        \label{fig:per_epoch}
    \end{center}
    \vspace{-10pt}
\end{figure}

To test our training methods, we show the prediction error of our trajectory based models versus different number of training epochs.
The mean error per step is shown in \tab{tab:epoch_error} and visualized in \fig{fig:per_epoch}.

\begin{table}[h]
\begin{center}
 \begin{tabular}{ r c} 
 Epochs Trained & Avg. Prediction Error per Step (Test Set)  \\ [1.0ex]
 \hline 
 1 &  \num{1.25E-2}  \\
 \hline
 3 &   \num{1.15E-2}  \\
 \hline
  5 & \num{1.12E-2}  \\
 \hline
 \textbf{8} & \num{1.02E-2} $*$  \\ 
 \hline
 12 & \num{1.07E-2}  \\ 
  \hline
16 & \num{1.09E-2}
\end{tabular}
\caption{Prediction Error on Test Set versus Training Epochs. 
($*$ indicates minumum cumulative error on the test set of 100 trajectories).}
\label{tab:epoch_error}
\end{center}
\end{table}

\subsection{Additional Cartpole Details}
\paragraph{Cartpole Environment Modifications}
In order to create more diverse data, we made the stop-conditions in the environment less strict and increased the randomness of initial states.
We increased the stop condition on the pole angle by 2 to \SI{24}{\degrees} and the $x$-position by 2 to 9.6.
Additionally, we increased the randomness of the initial values of $x,\theta$ by a factor of 20.

\paragraph{Cartpole Linearization Equations}
This section includes the details needed to implement the LQR controller used in \sect{sec:results}.
Modeling the cartpole as a nonlinear dynamical system in 4 states and 1 input takes the form
$ \dot{x} = f(\vec{x},u)$. 
By taking the Jacobian of the dynamics function near the unstable equilibrium of $\theta=0$ yields a linear system of the form.
\begin{align}
\dot{\vec{x}} &= \widetilde A  \vec{x} + \widetilde B u \\
\widetilde A  &= 
\begin{bmatrix}
0 & 1 & 0 & 0 \\
0 & \frac{gm_{p}}{m_c} & 0 & 0 \\
0 & 0 & 0 & 1 \\
0 & 0 & \frac{g(m_c+m_p)}{lm_c} & 0 \\
\end{bmatrix} \\
\widetilde B &= 
\begin{bmatrix}
0 & \frac{1}{m_c} & 0 & \frac{-1}{lm_c}
\end{bmatrix}
\label{eq:cp-linearized}
\end{align}

\subsection{State Predictions}
\begin{figure}[H]
    \begin{center}
        \small{\cblock{0}{0}{0} Ground Truth
        (\textcolor[rgb]{.0,.0,.0}{\large{$-$}})\quad
        \cblock{200}{0}{0} Trajectory-based Prediction
        (\textcolor[rgb]{.78,.01,.01}{$\Box$}) 
        } 
    \end{center}
    \centering

    \begin{subfigure}{0.3\columnwidth}  
        \centering 
        \includegraphics[width=\linewidth]{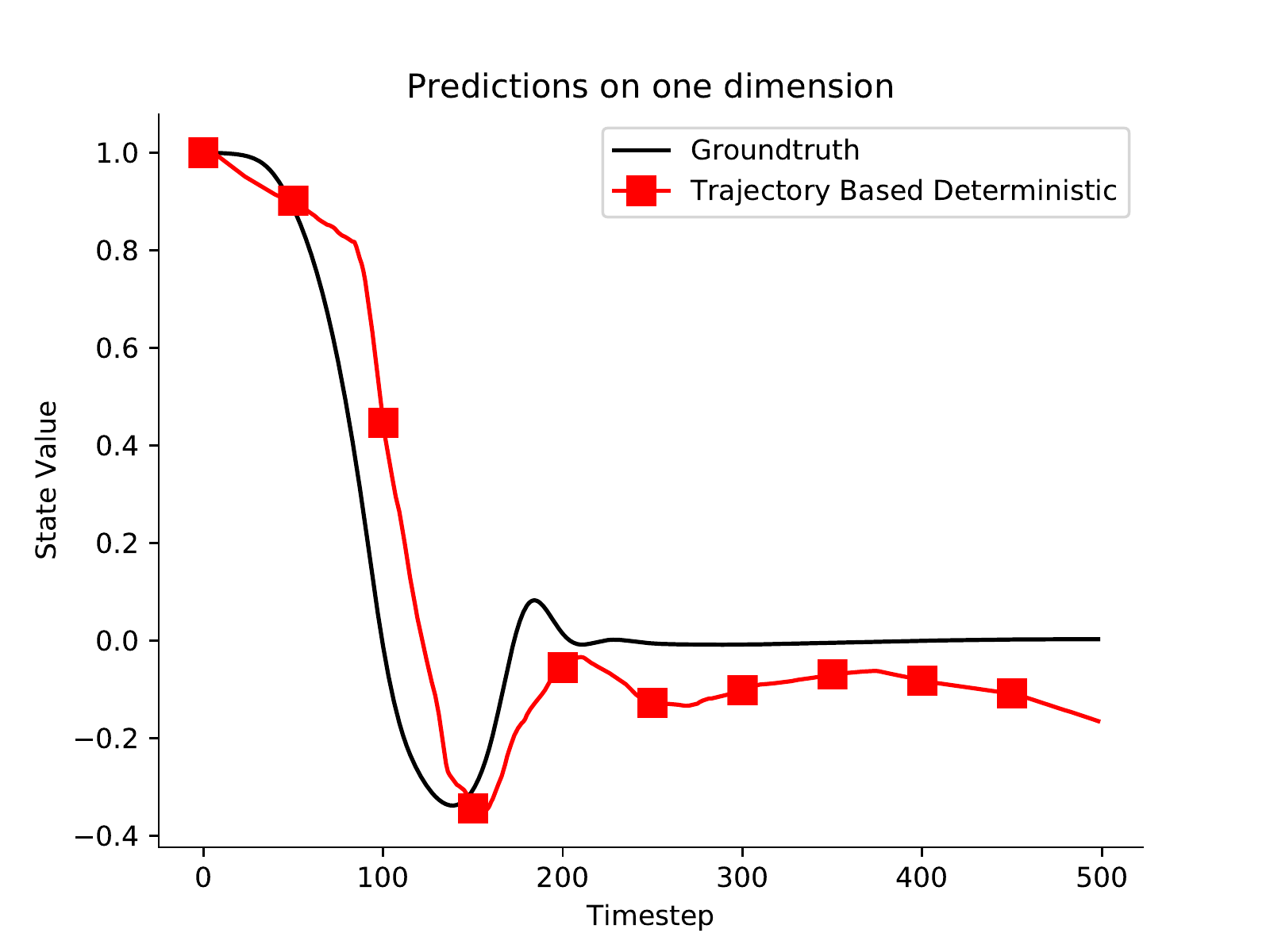}
        \caption{State 0.}    
        \label{fig:state-0}
    \end{subfigure}
    ~
    \begin{subfigure}{0.3\columnwidth}  
        \centering 
        \includegraphics[width=\linewidth]{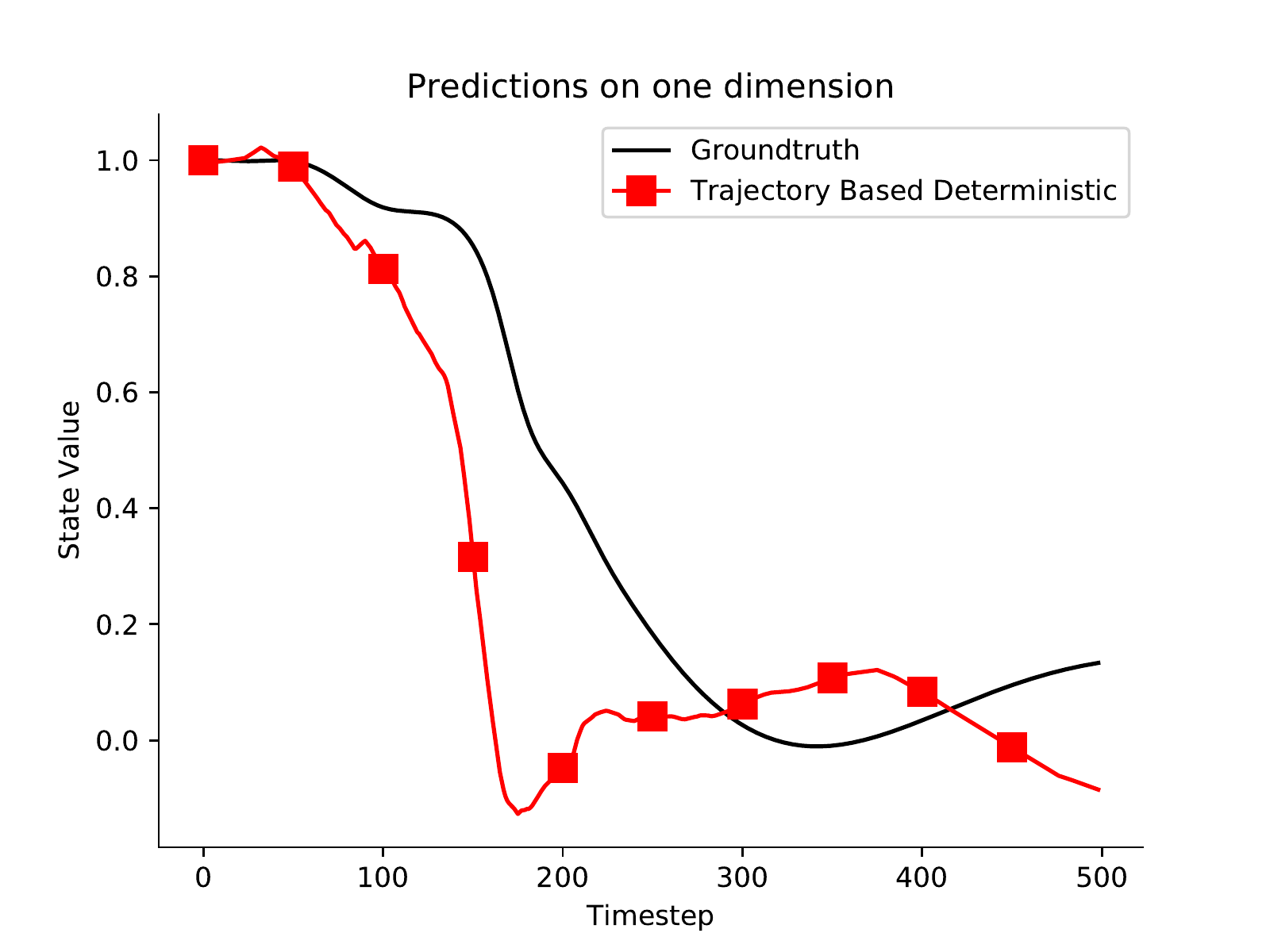}
        \caption{State 1.}    
        \label{fig:state-1}
    \end{subfigure}
    ~
    \begin{subfigure}{0.3\columnwidth}  
        \centering 
        \includegraphics[width=\linewidth]{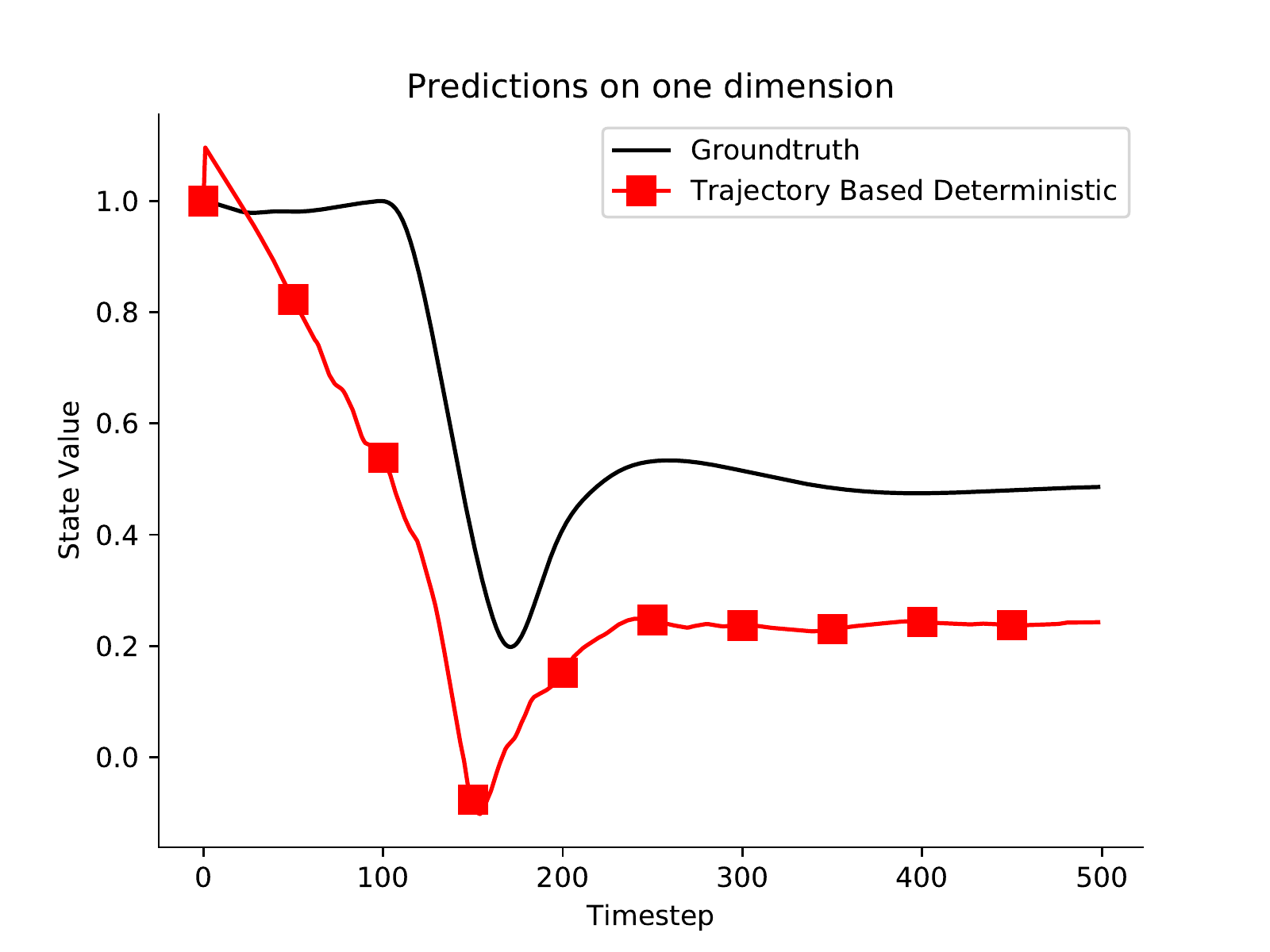}
        \caption{State 2.}    
        \label{fig:state-2}
    \end{subfigure}
    ~
    \begin{subfigure}{0.3\columnwidth}  
        \centering 
        \includegraphics[width=\linewidth]{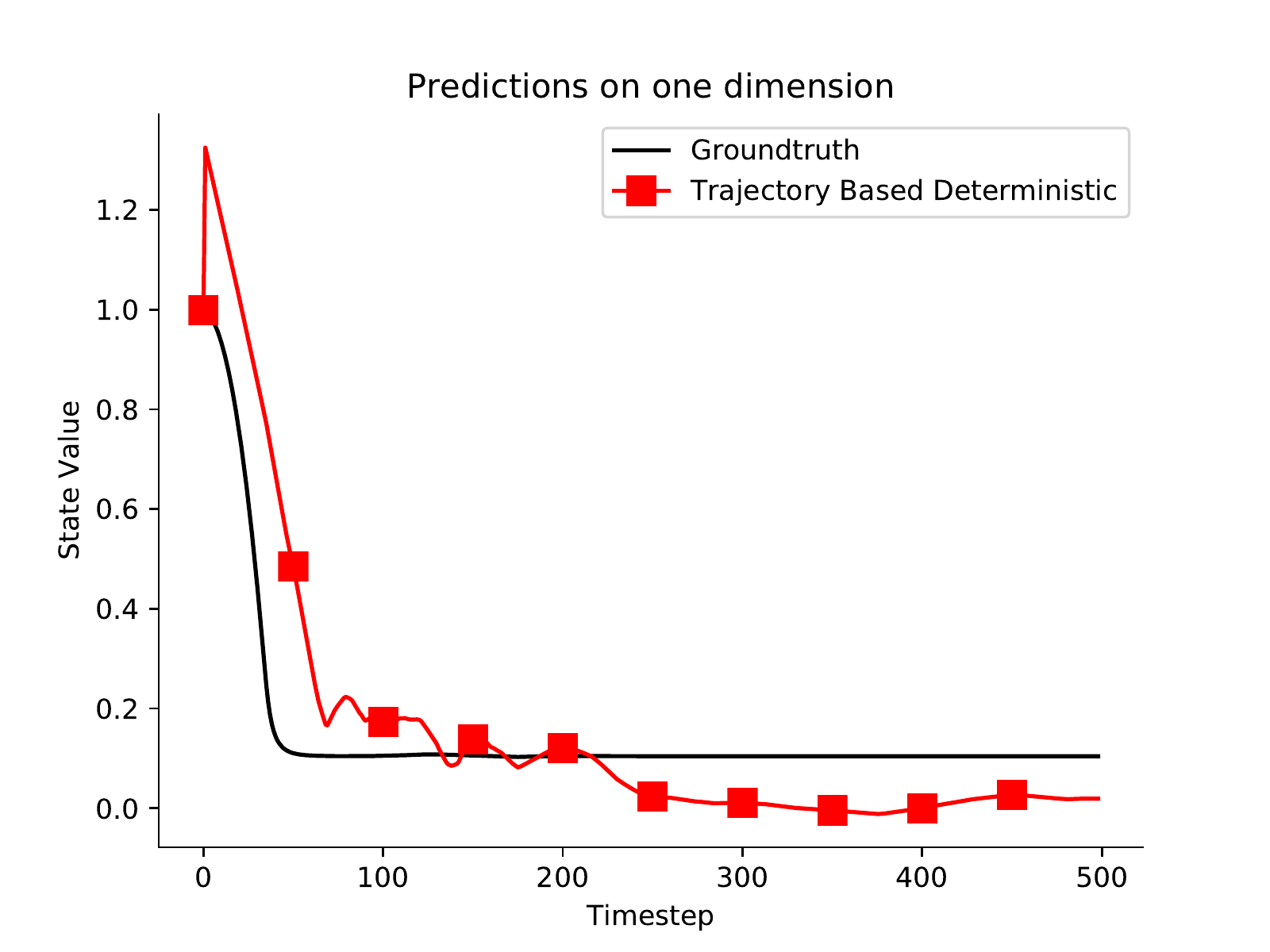}
        \caption{State 3.}    
        \label{fig:state-3}
    \end{subfigure}
    ~
    \begin{subfigure}{0.3\columnwidth}  
        \centering 
        \includegraphics[width=\linewidth]{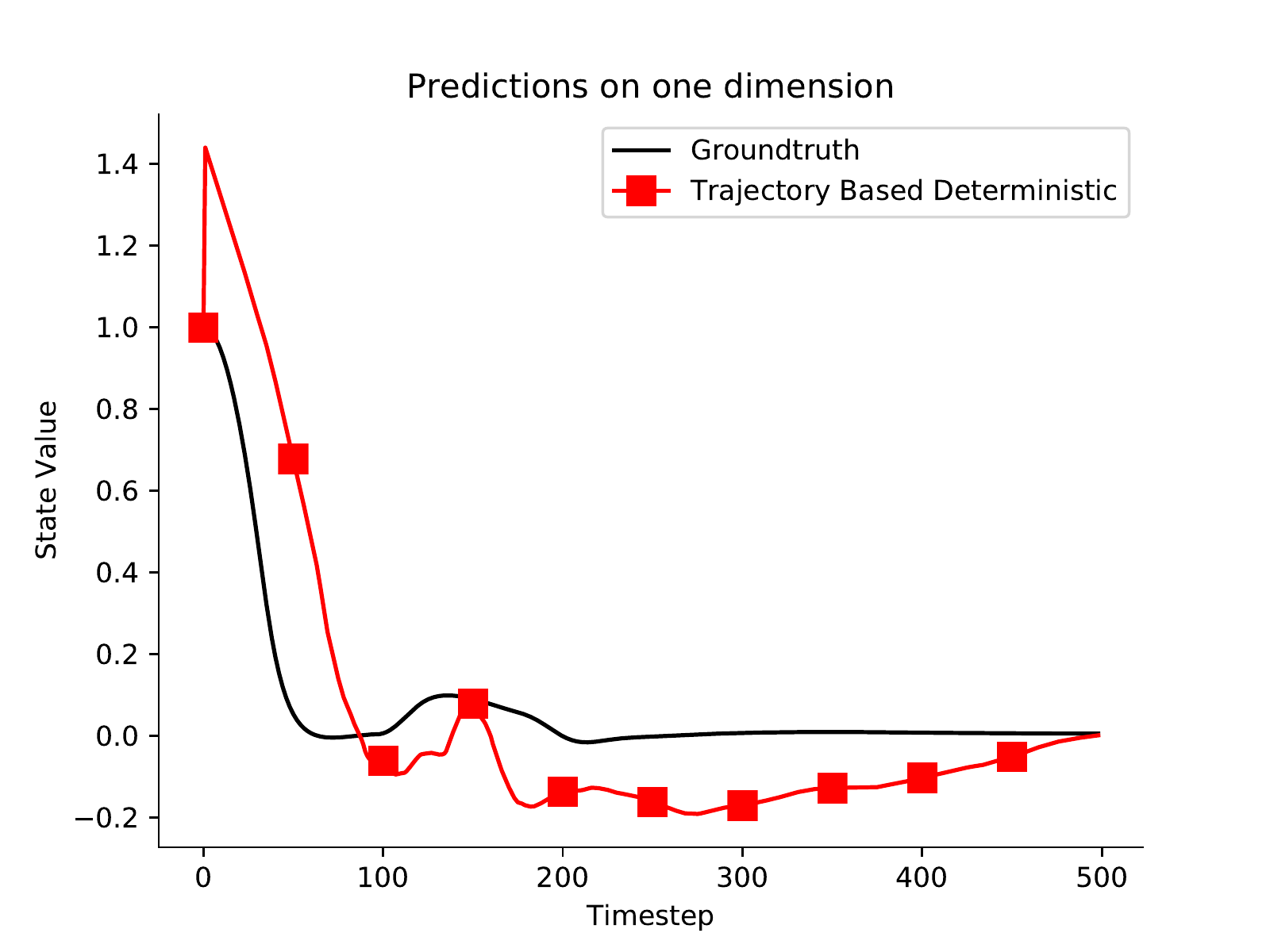}
        \caption{State 4.}    
        \label{fig:state-4}
    \end{subfigure}
    ~
    \begin{subfigure}{0.3\columnwidth}  
        \centering 
        \includegraphics[width=\linewidth]{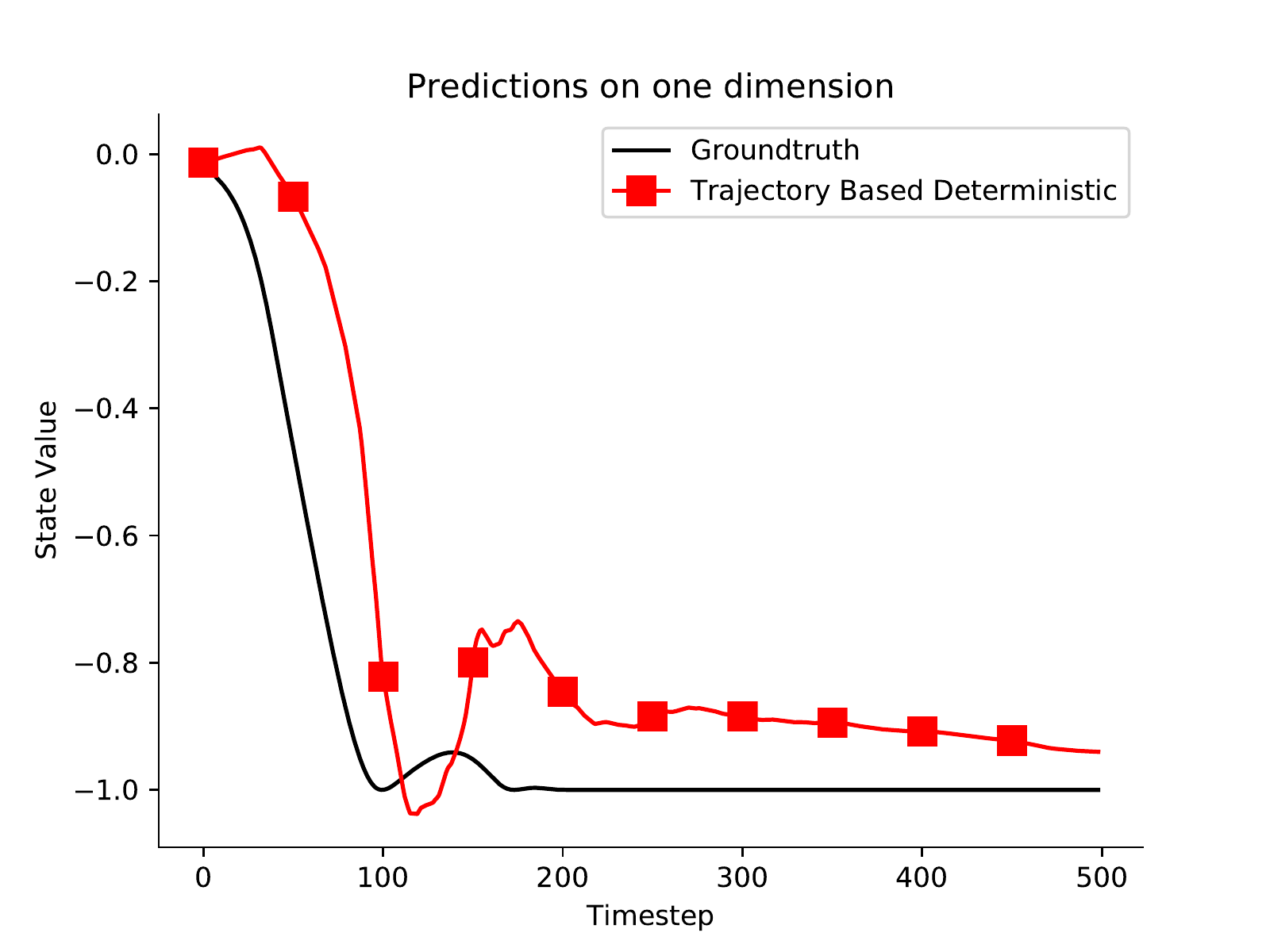}
        \caption{State 5.}    
        \label{fig:state-5}
    \end{subfigure}
    ~
    \begin{subfigure}{0.3\columnwidth}  
        \centering 
        \includegraphics[width=\linewidth]{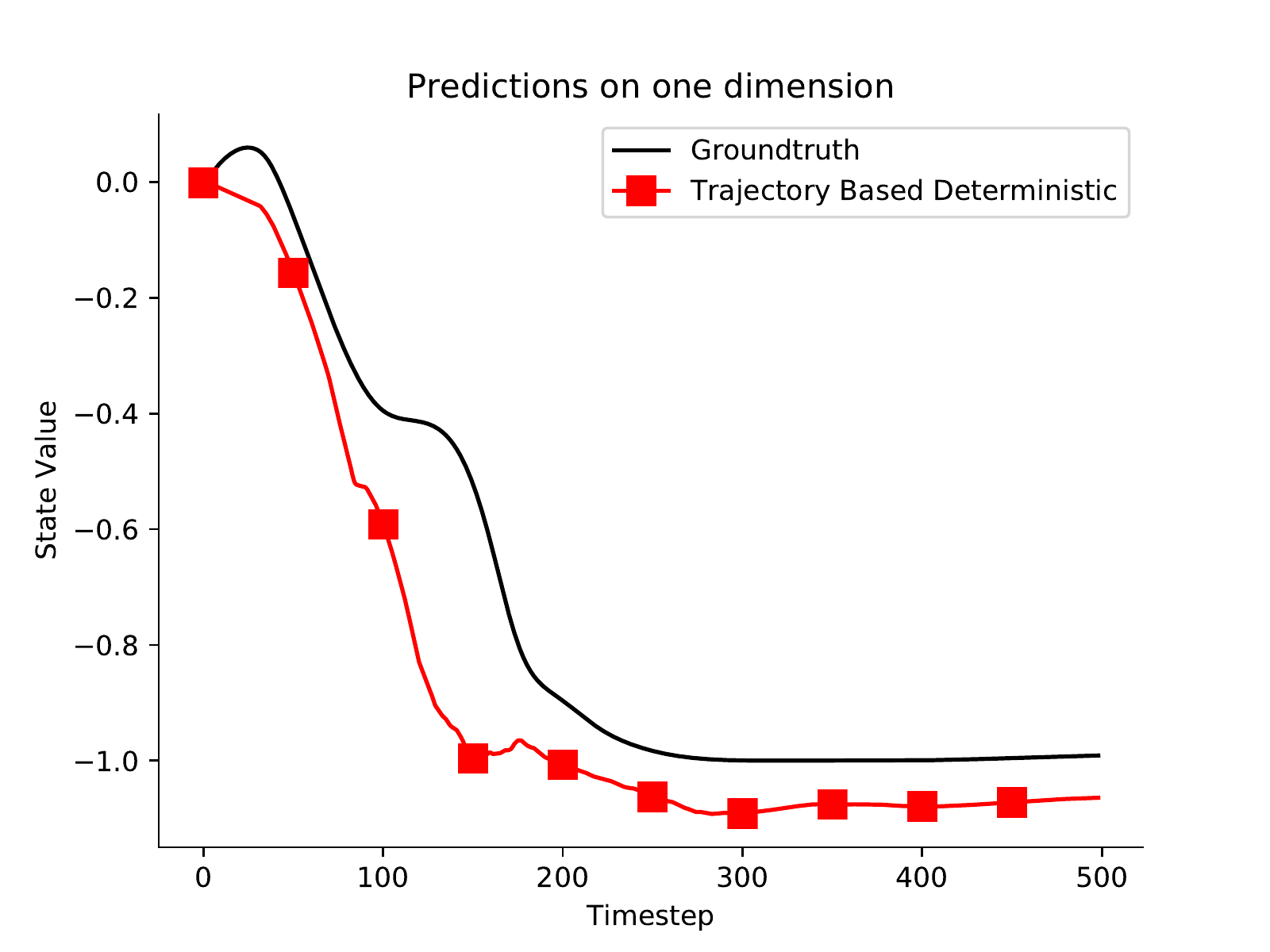}
        \caption{State 6.}    
        \label{fig:state-6}
    \end{subfigure}
    ~
    \begin{subfigure}{0.3\columnwidth}  
        \centering 
        \includegraphics[width=\linewidth]{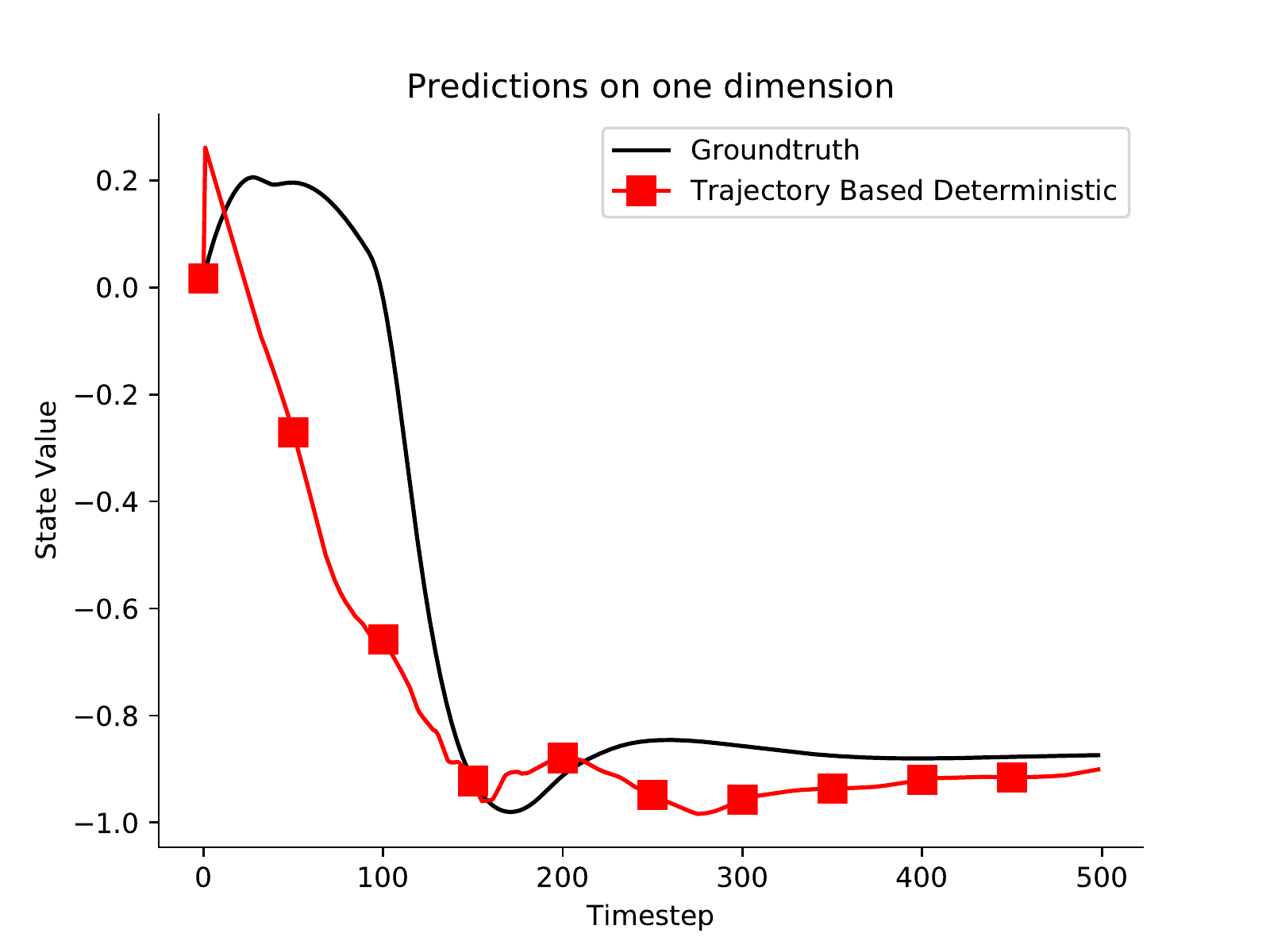}
        \caption{State 7.}    
        \label{fig:state-7}
    \end{subfigure}
    ~
    \begin{subfigure}{0.3\columnwidth}  
        \centering 
        \includegraphics[width=\linewidth]{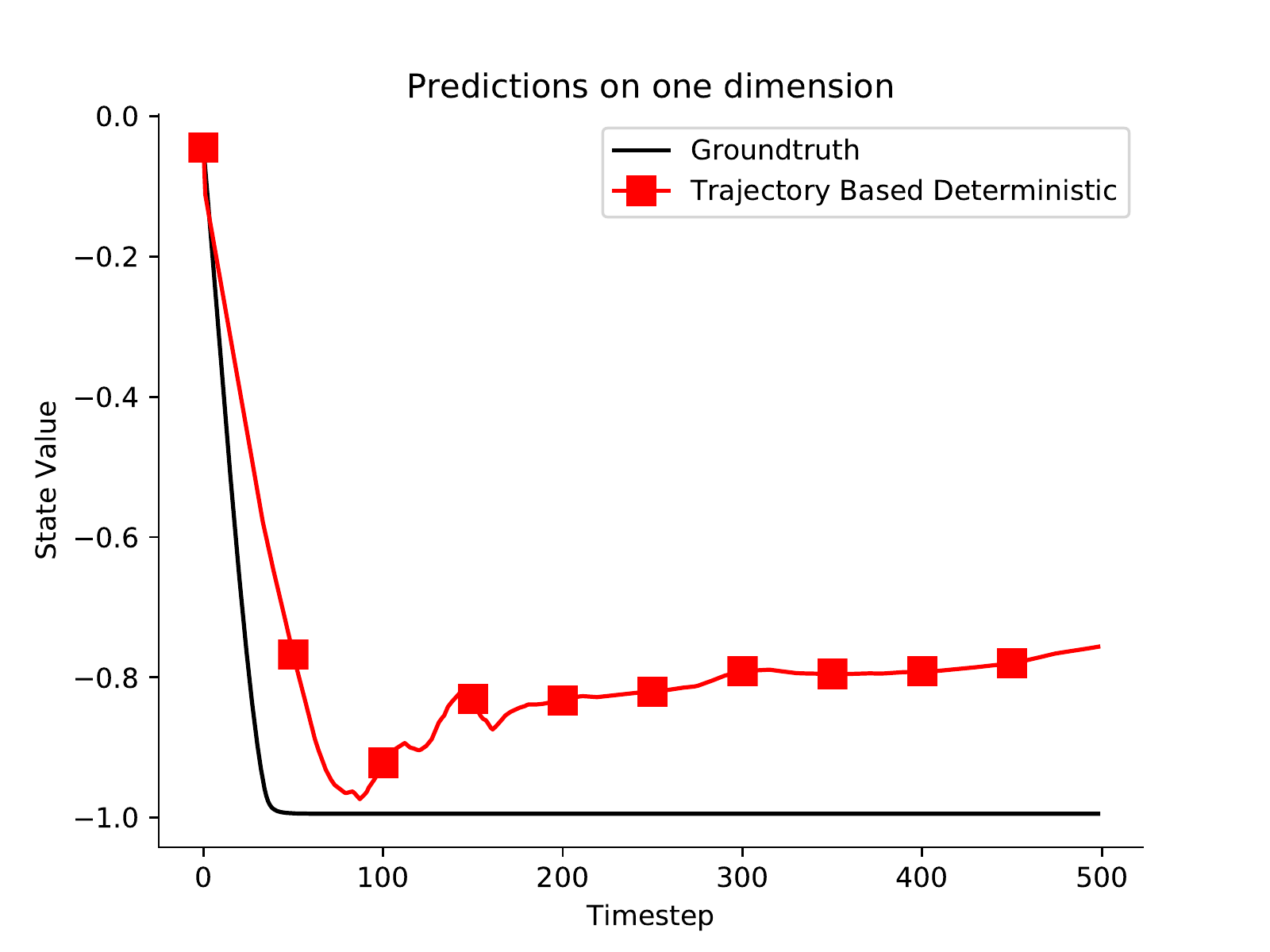}
        \caption{State 8.}    
        \label{fig:state-8}
    \end{subfigure}
    ~
    \begin{subfigure}{0.3\columnwidth}  
        \centering 
        \includegraphics[width=\linewidth]{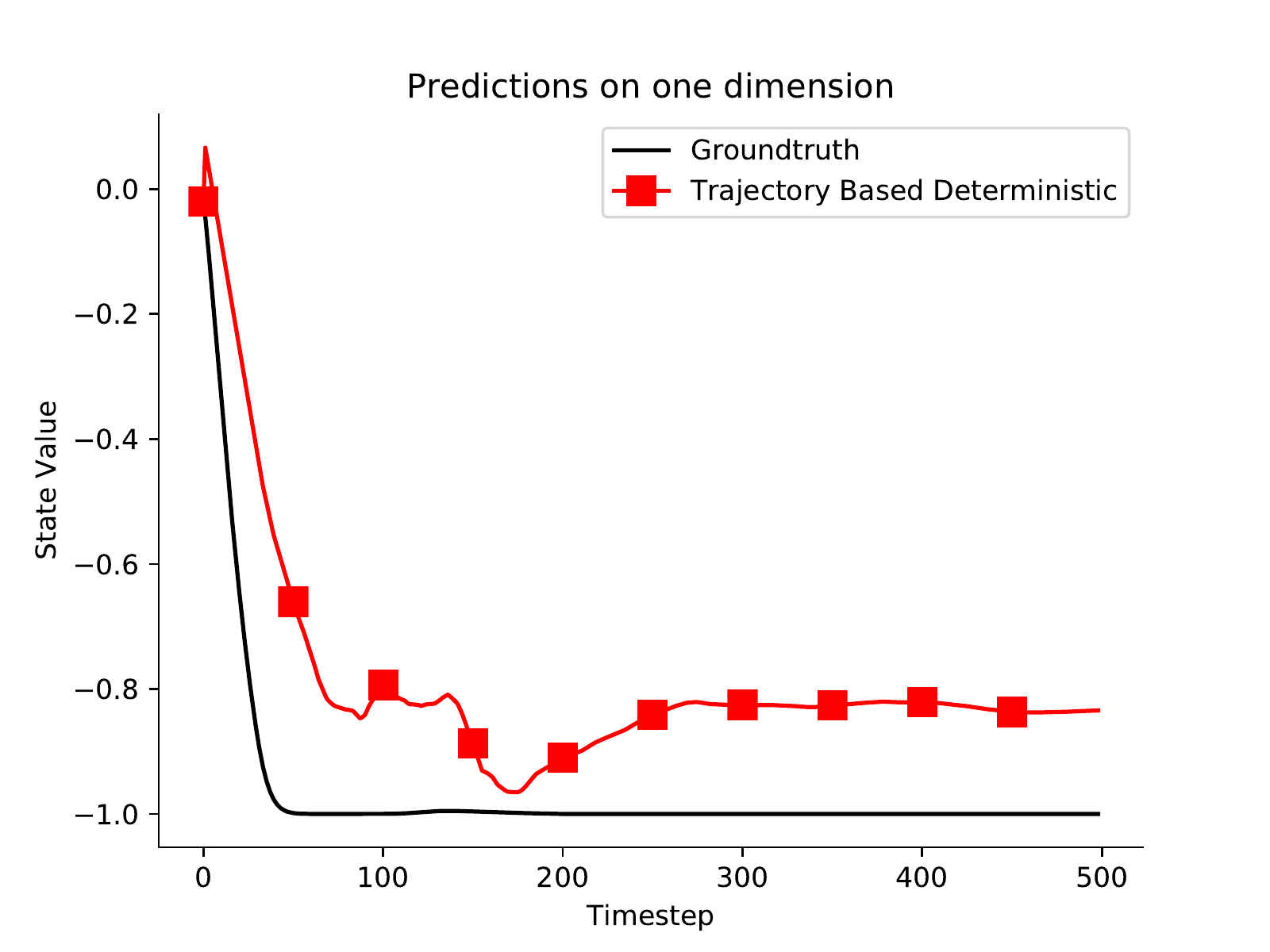}
        \caption{State 9.}    
        \label{fig:state-9}
    \end{subfigure}
    ~
    \begin{subfigure}{0.3\columnwidth}  
        \centering 
        \includegraphics[width=\linewidth]{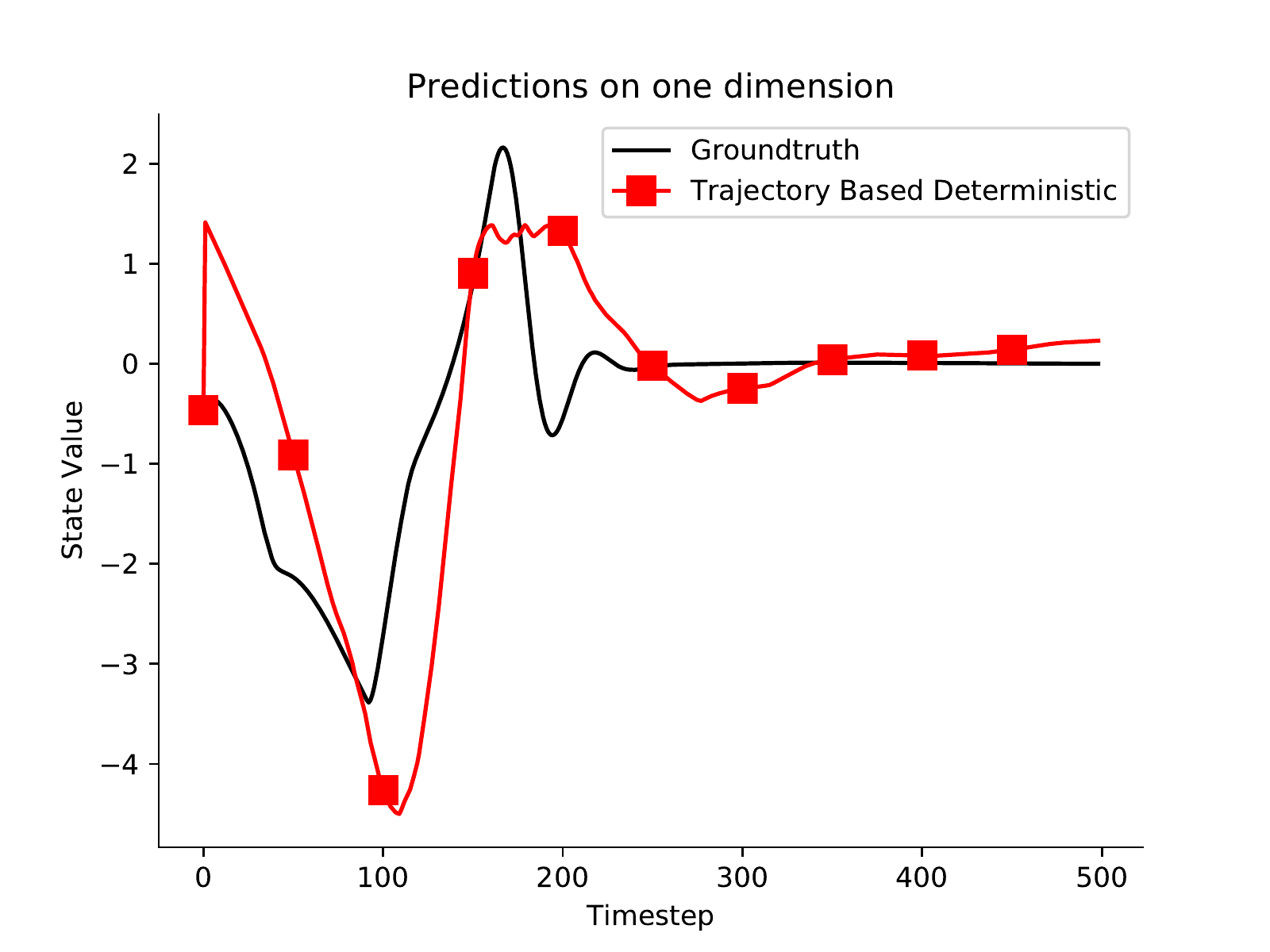}
        \caption{State 10.}    
        \label{fig:state-10}
    \end{subfigure}
    ~
    \begin{subfigure}{0.3\columnwidth}  
        \centering 
        \includegraphics[width=\linewidth]{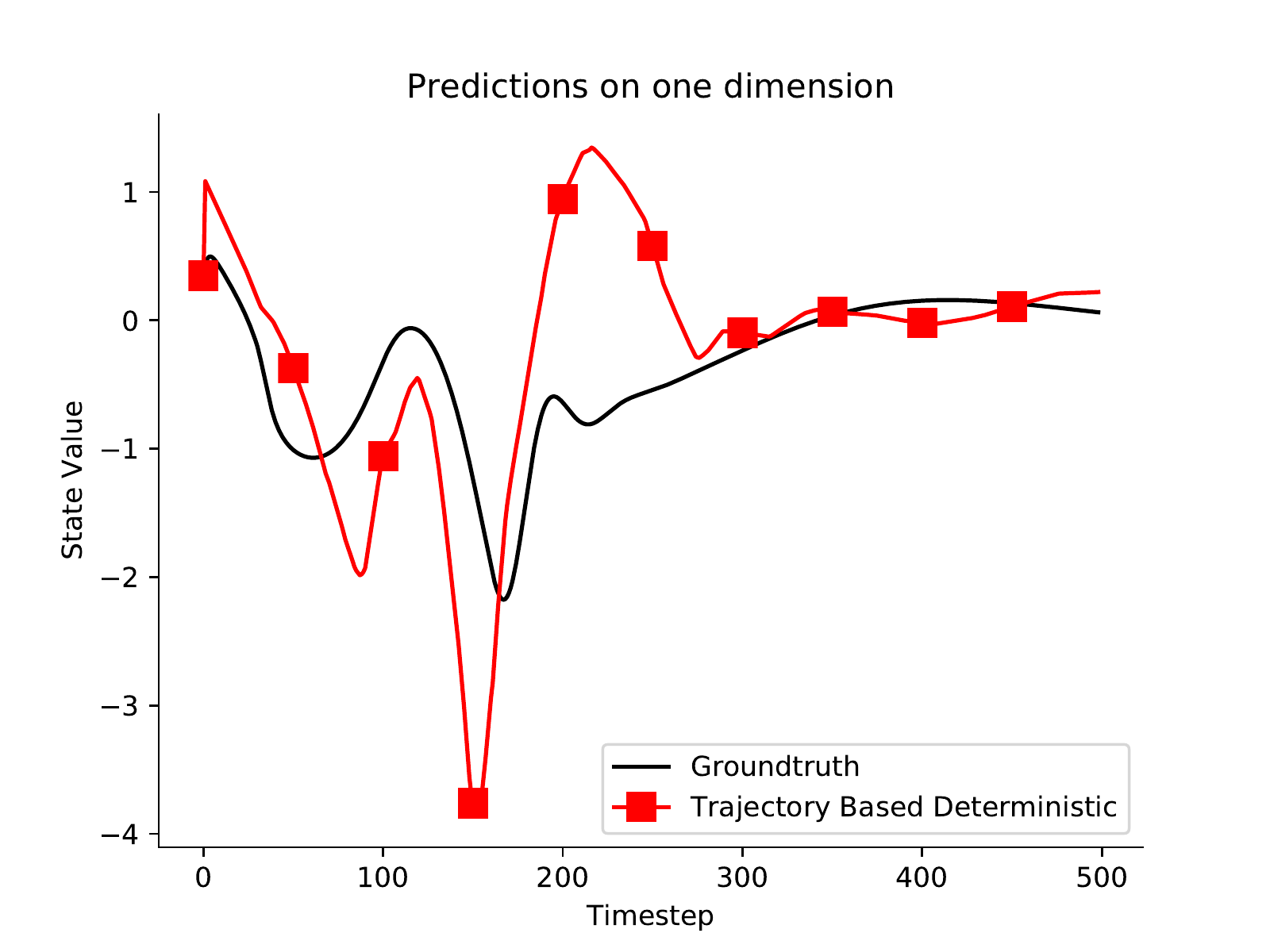}
        \caption{State 11.}    
        \label{fig:state-11}
    \end{subfigure}
    ~
    \begin{subfigure}{0.3\columnwidth}  
        \centering 
        \includegraphics[width=\linewidth]{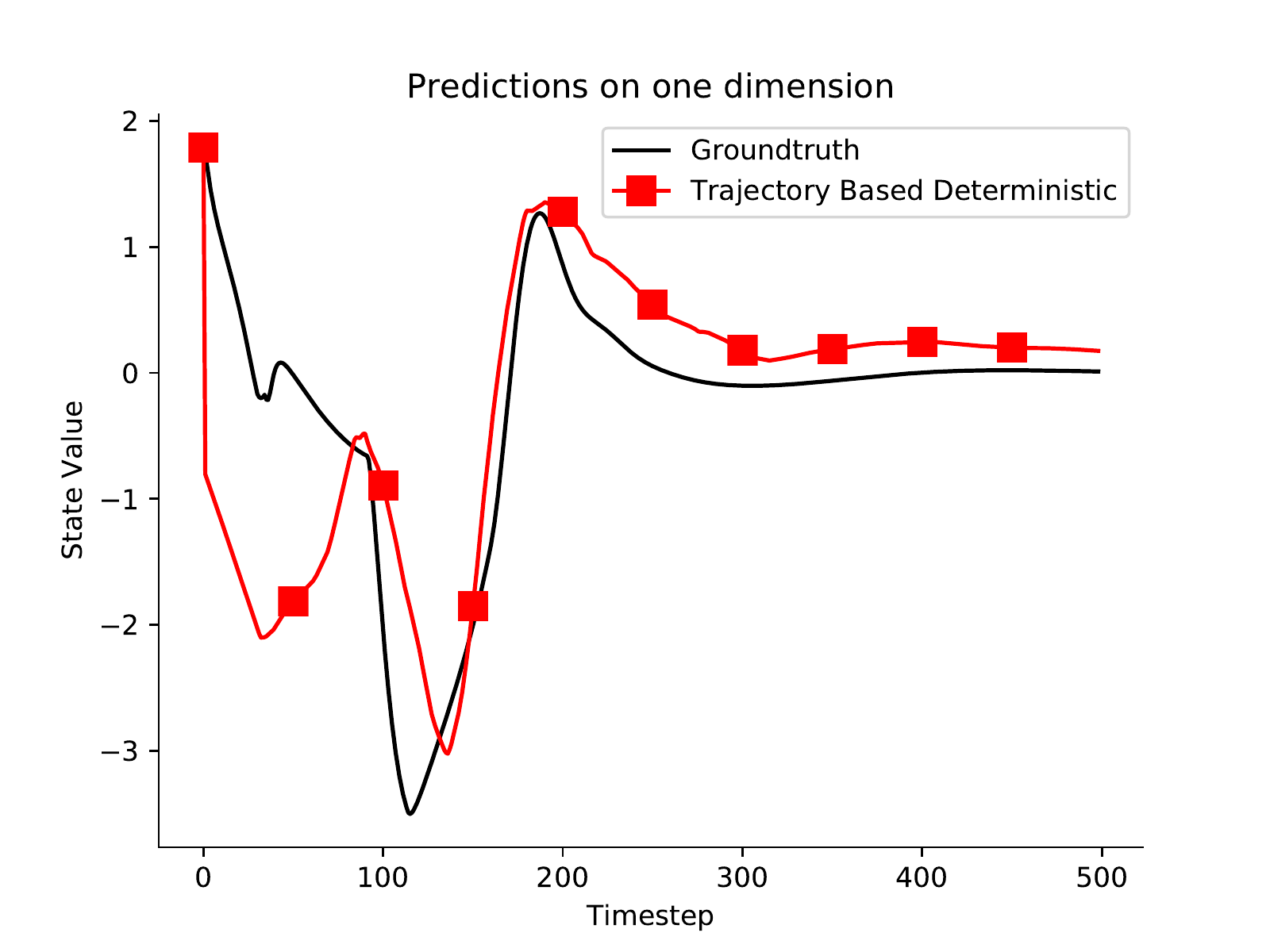}
        \caption{State 12.}    
        \label{fig:state-12}
    \end{subfigure}
    ~
    \begin{subfigure}{0.3\columnwidth}  
        \centering 
        \includegraphics[width=\linewidth]{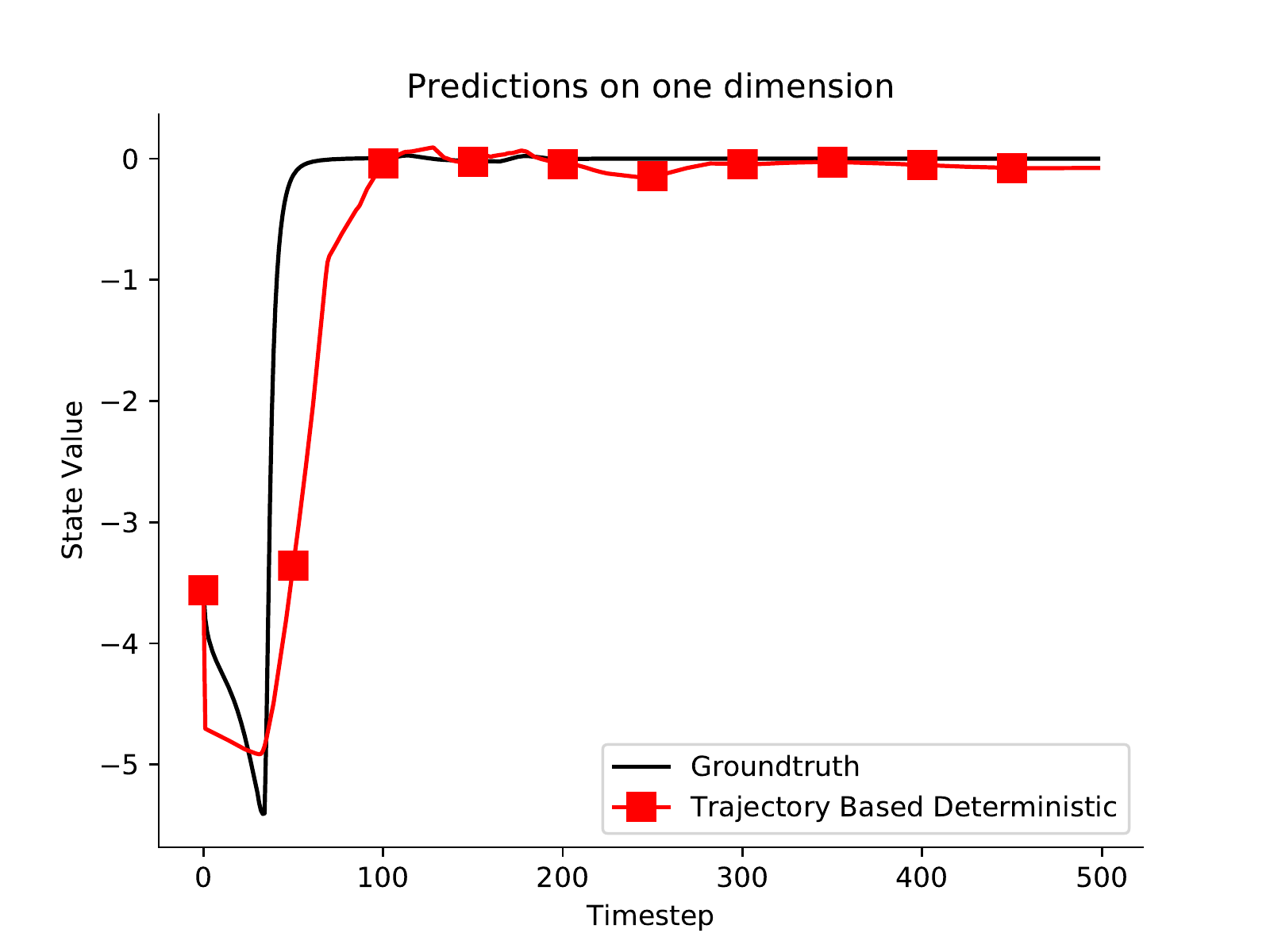}
        \caption{State 13.}    
        \label{fig:state-13}
    \end{subfigure}
    ~
    \begin{subfigure}{0.3\columnwidth}  
        \centering 
        \includegraphics[width=\linewidth]{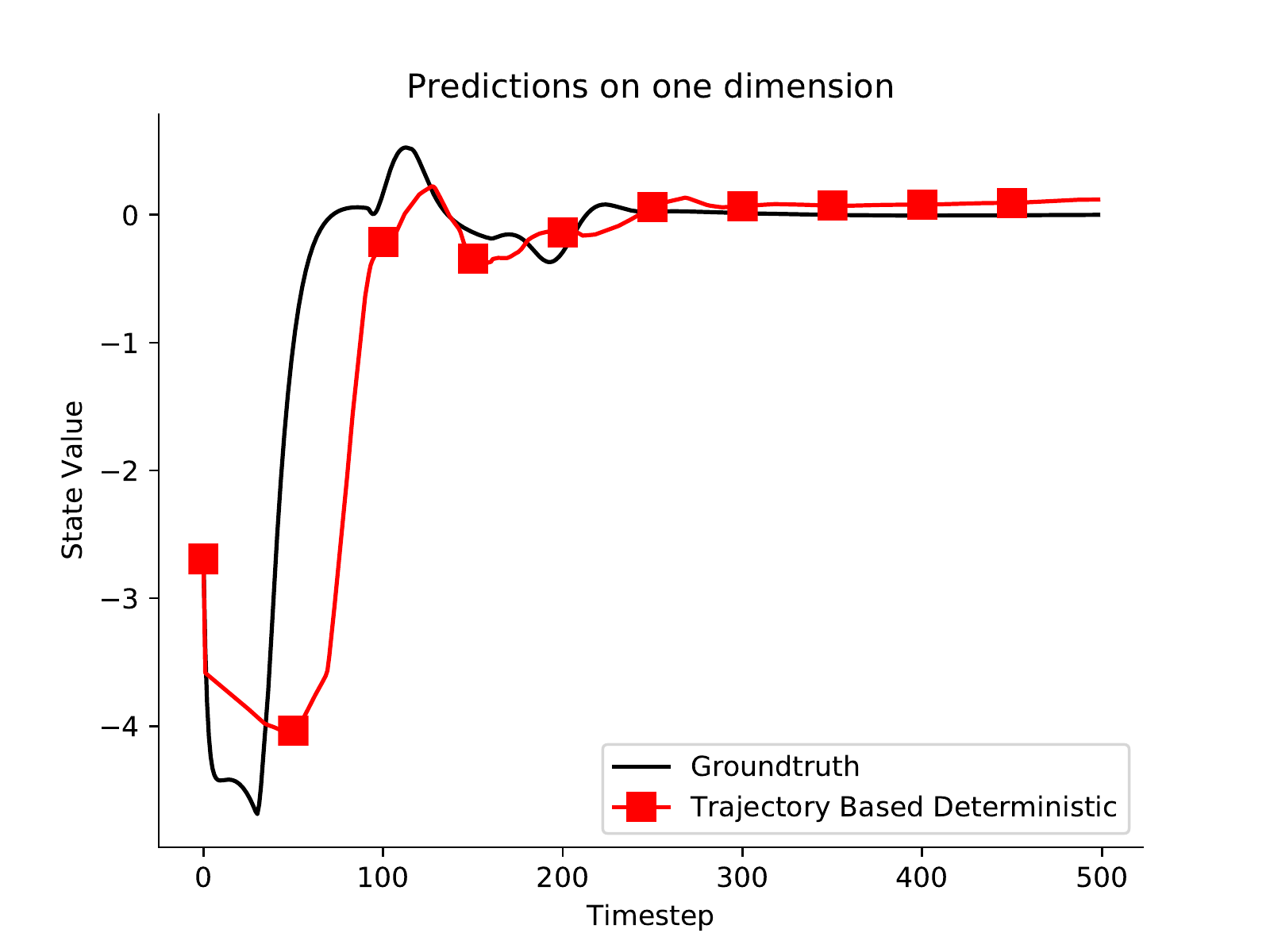}
        \caption{State 14.}    
        \label{fig:state-14}
    \end{subfigure}
    ~
   
    \caption{The T model's predictions on each of the state vector's dimensions for a Reacher episode of length $500$. 
    It achieves impressive short and long term accuracy for all of the states.
    }
    \label{fig:rch-pred}
\end{figure}

\begin{figure}[H]
    \begin{center}
         \small{\cblock{0}{0}{0} Ground Truth
        (\textcolor[rgb]{.0,.0,.0}{\large{$-$}})\
        \cblock{200}{0}{0} T 
        (\textcolor[rgb]{.78,.01,.01}{$\Box$}) \
        \cblock{200}{200}{0} TP 
        (\textcolor[rgb]{.78,.78,.01}{$\pentagon$}) \
        \cblock{0}{0}{200} D 
        (\textcolor[rgb]{.01,.01,.78}{\large{$\circ$}}) \
        \cblock{0}{200}{0} P 
        (\textcolor[rgb]{.01,.78,.01}{$\diamondsuit$}) 
        } 
    \end{center}
    \centering 
    \vspace{-5pt}
    \begin{subfigure}{0.48\columnwidth}  
        \centering 
        \includegraphics[width=\linewidth]{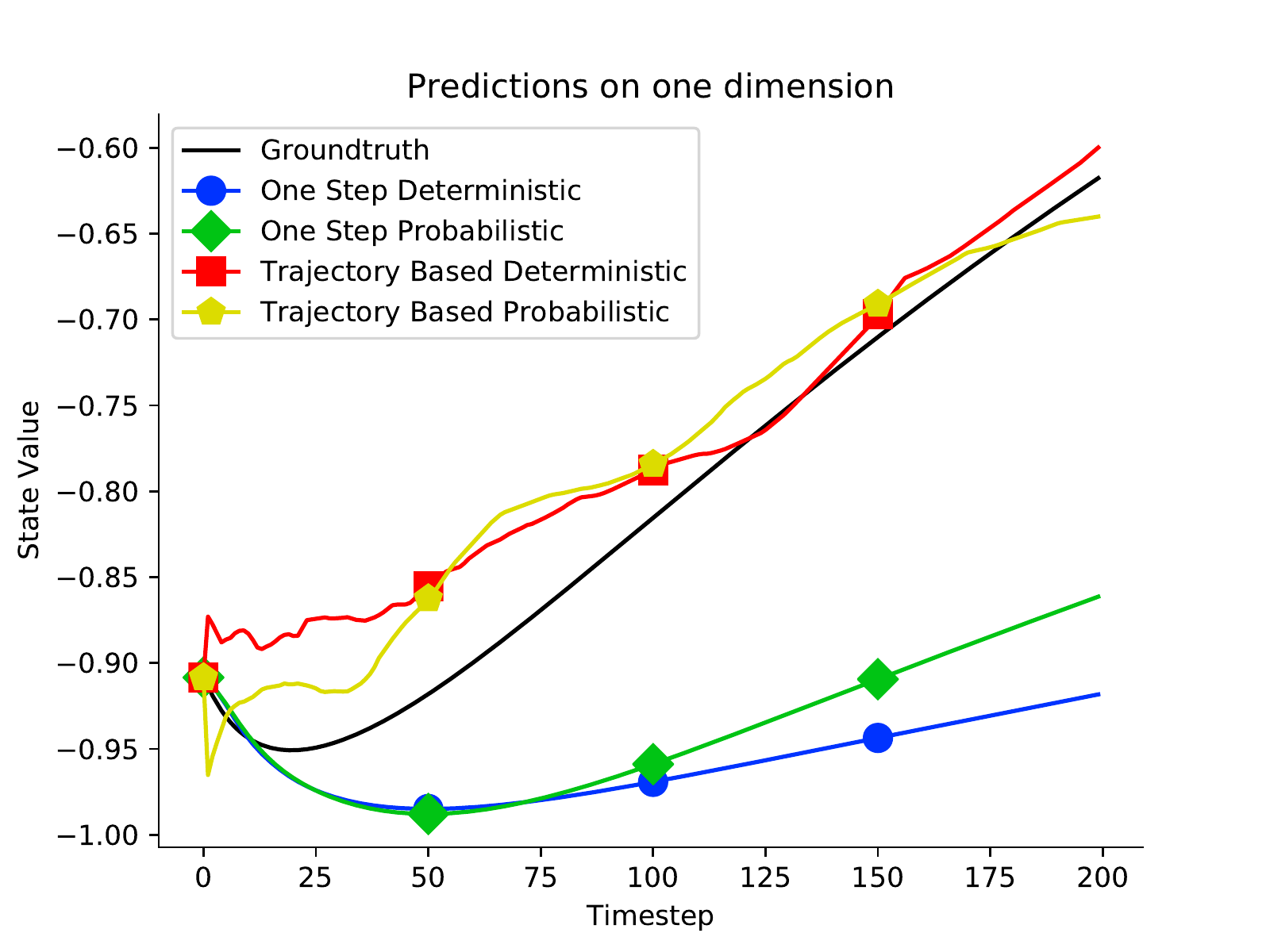}
        \caption{State 0: $x$ position.}    
        \label{fig:state-0}
    \end{subfigure}
    ~
    \begin{subfigure}{0.48\columnwidth}  
        \centering 
        \includegraphics[width=\linewidth]{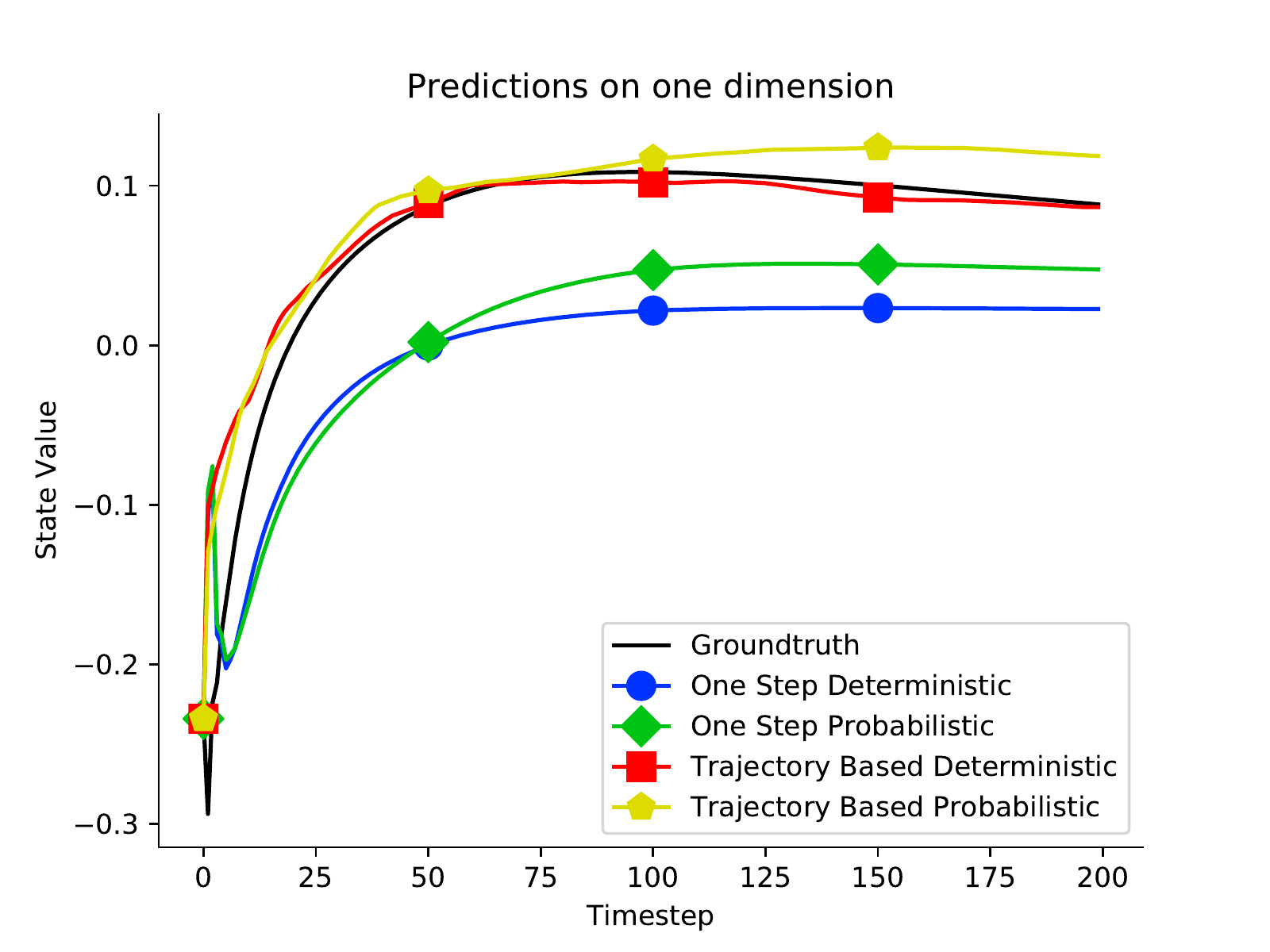}
        \caption{State 1: $x$ velocity.}    
        \label{fig:state-1}
    \end{subfigure}
    ~
    \begin{subfigure}{0.48\columnwidth}  
        \centering 
        \includegraphics[width=\linewidth]{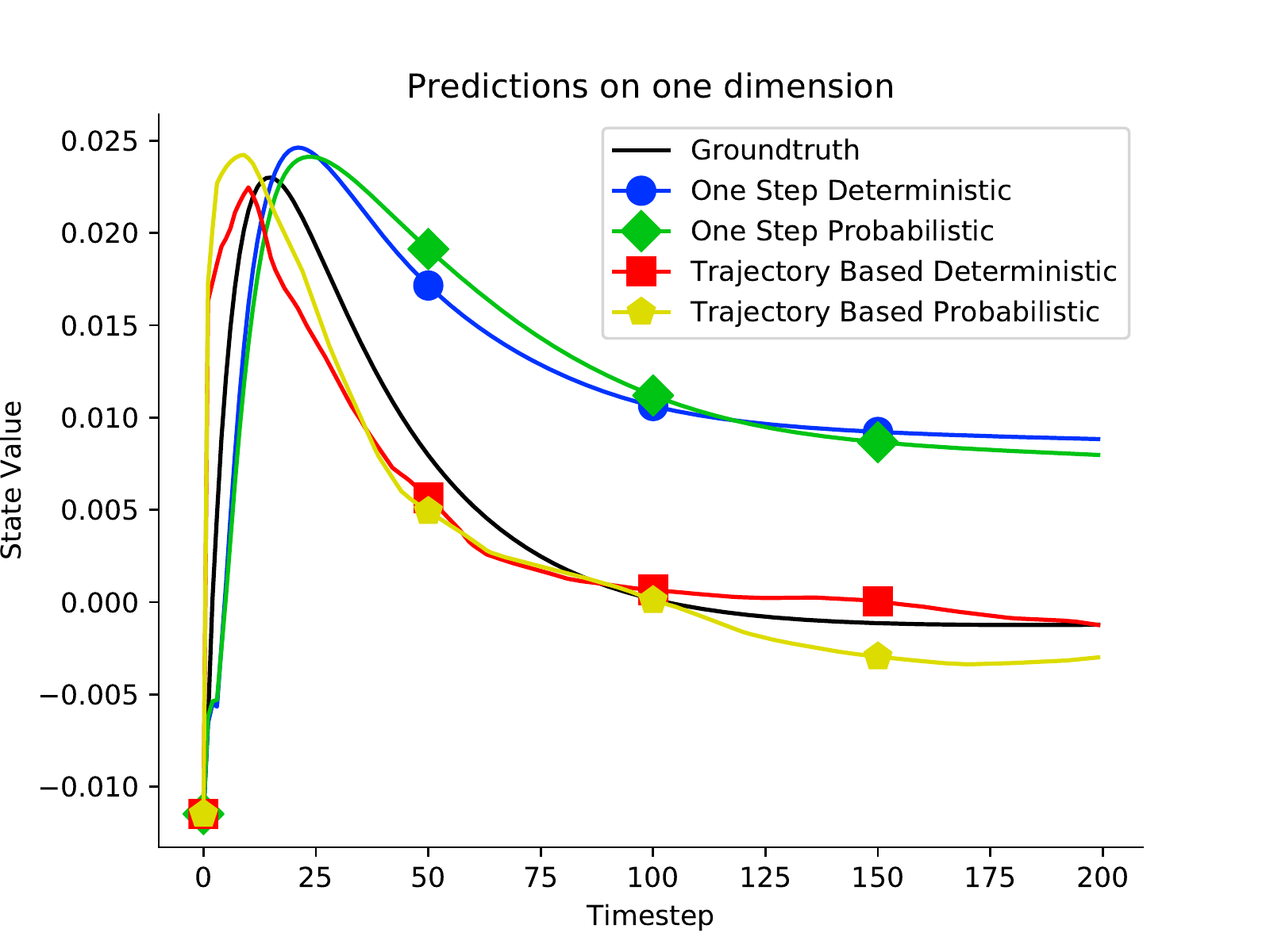}
        \caption{State 2: pole angle.}    
        \label{fig:state-2}
    \end{subfigure}
    ~
    \begin{subfigure}{0.48\columnwidth}  
        \centering 
        \includegraphics[width=\linewidth]{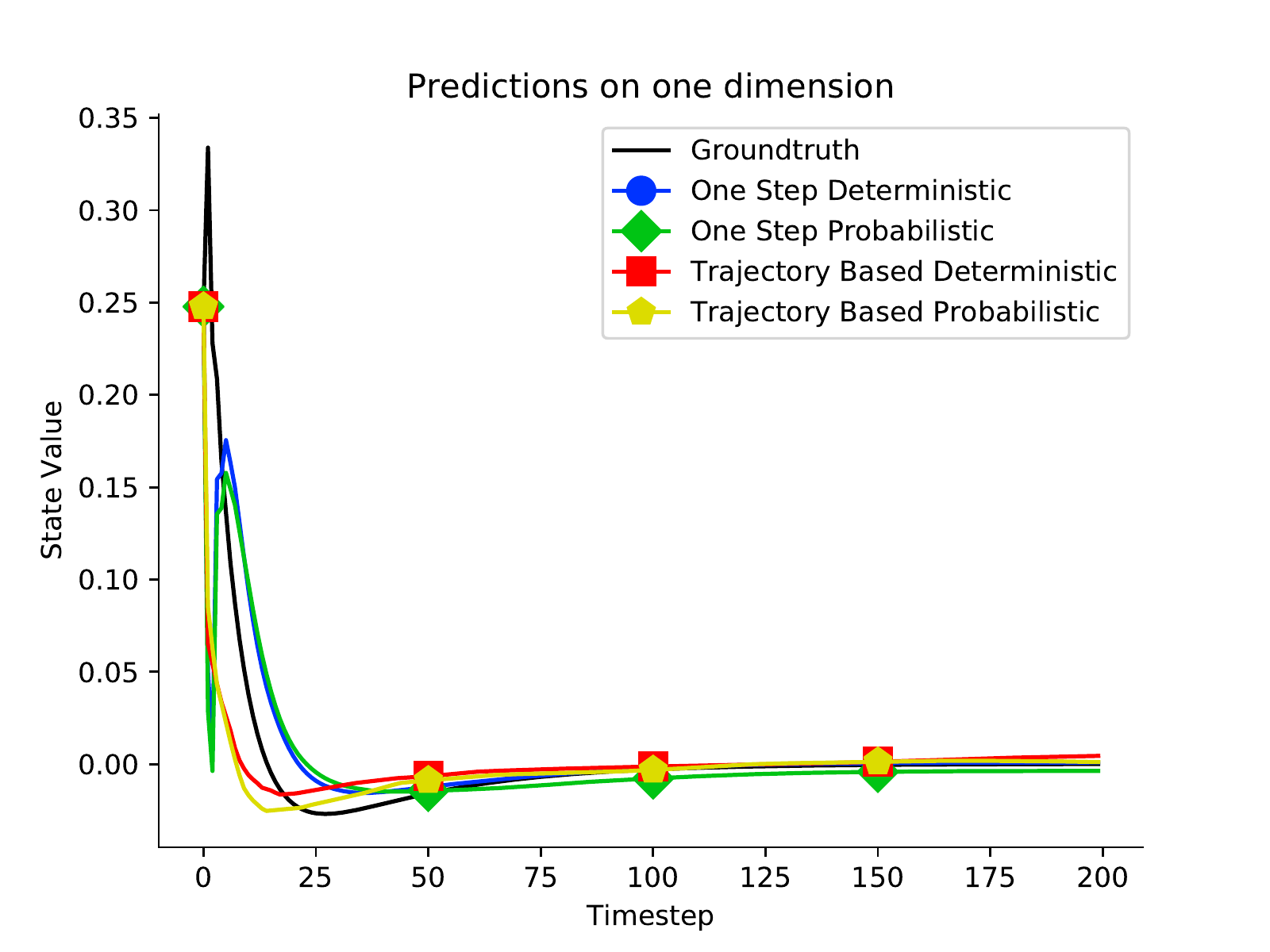}
        \caption{State 3: angular velocity.}    
        \label{fig:state-3}
    \end{subfigure}
    \caption{Cartpole state predictions for the following models: $D,P,T,TP$. 
    This highlights two properties of the cartpole dataset: 
    1) the damped response of versions of LQR control and 
    2) the end behavior of the trajectories varies and does not always converge to 0, where the trajectory-based models are more accurate.}
    \label{fig:cp-pred}
\end{figure}

\begin{figure}[H]
        \begin{center}
        \small{\cblock{0}{0}{200} D (\textcolor[rgb]{.0,.0,.78}{\large{$\circ$}})\ 
        \cblock{20}{128}{20} PE (\textcolor[rgb]{.078,.50,.078}{$\diamondsuit$}) \
        \cblock{200}{0}{0} T
        (\cblock{200}{0}{0})
        \cblock{0}{0}{0} True Trajectory (\textcolor[rgb]{.01,.01,.01}{\large{-}})}
        \end{center}
        \centering 
        \includegraphics[width=\linewidth]{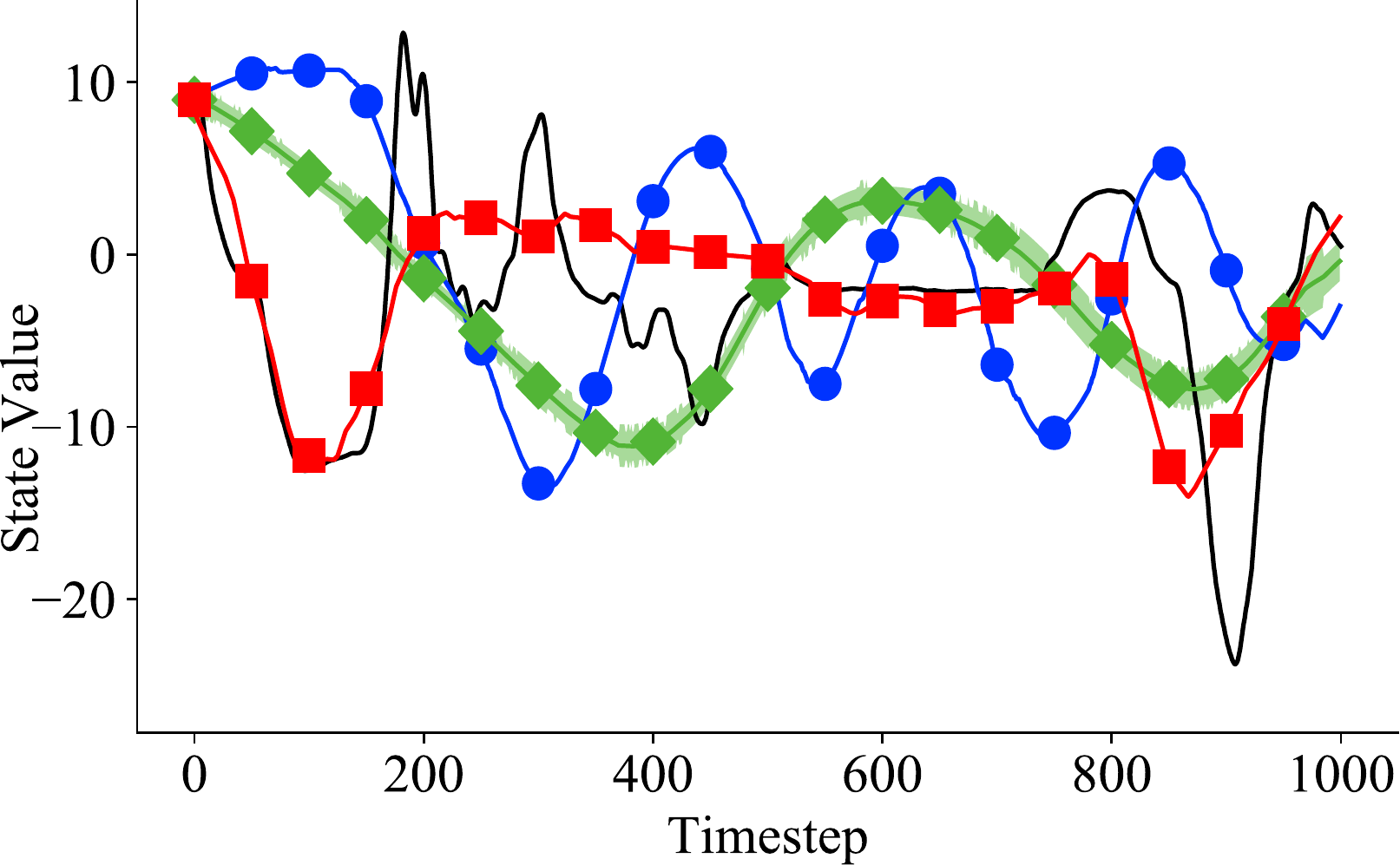}
        \caption{Predicting on real hardware.}    
        \label{fig:ex-real}
\end{figure}

We include visualizations of the state-predictions for the reacher environment, \fig{fig:rch-pred}, the cartpole environment, \fig{fig:cp-pred}, and the real quadrotor, \fig{fig:ex-real}.

\begin{figure*}[t]
    \vspace{6pt}
    \begin{center}
    \small{\cblock{20}{200}{20} pole-angle, $\theta$ (\textcolor[rgb]{.2,.78,.2}{$\upbowtie$})\quad
    \cblock{200}{0}{0} $x$ position
    (\textcolor[rgb]{.78,.01,.01}{\large{$\circ$}}) 
    } 
    \end{center}
    \centering
     \begin{subfigure}{0.32\linewidth}  
        \centering 
        \includegraphics[width=\linewidth]{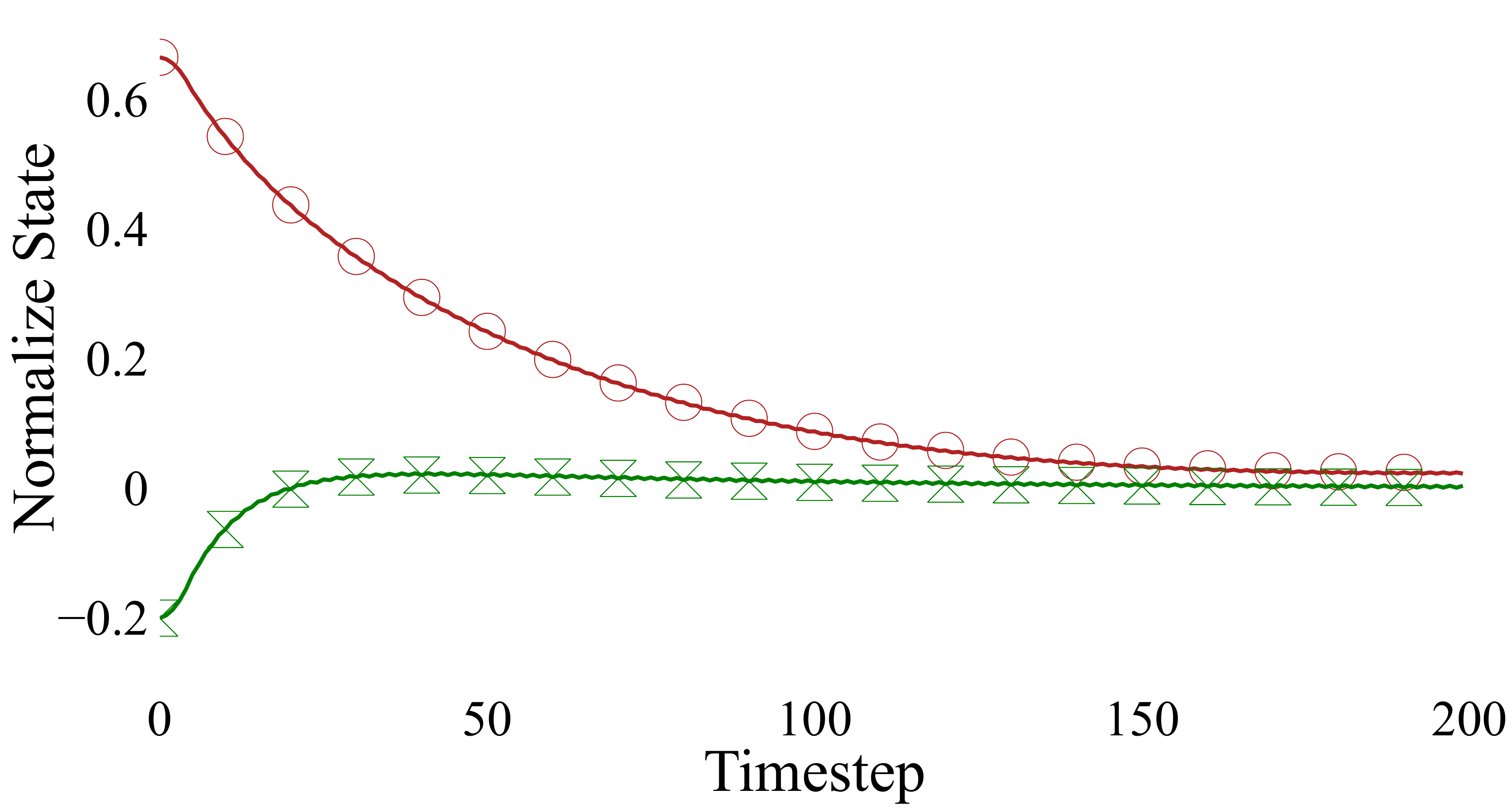}
        \caption{A stable trajectory.}    
        \label{fig:s-s}
    \end{subfigure}
    \hfill
    \begin{subfigure}{0.32\linewidth}  
        \centering 
        \includegraphics[width=\linewidth]{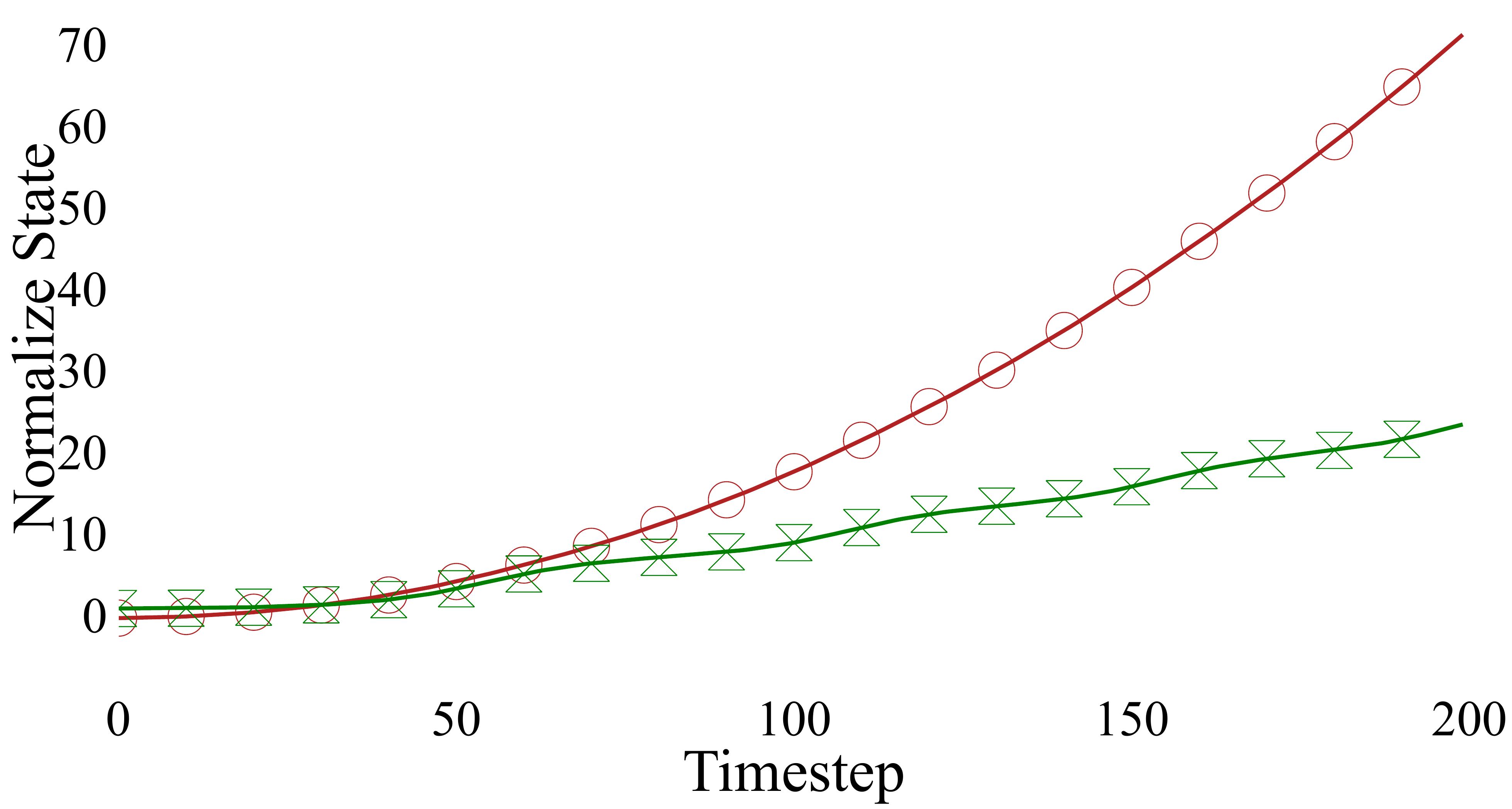}
        \caption{An unstable trajectory.}    
        \label{fig:s-u}
    \end{subfigure}
    \hfill
    \begin{subfigure}{0.32\linewidth}  
        \centering 
        \includegraphics[width=\linewidth]{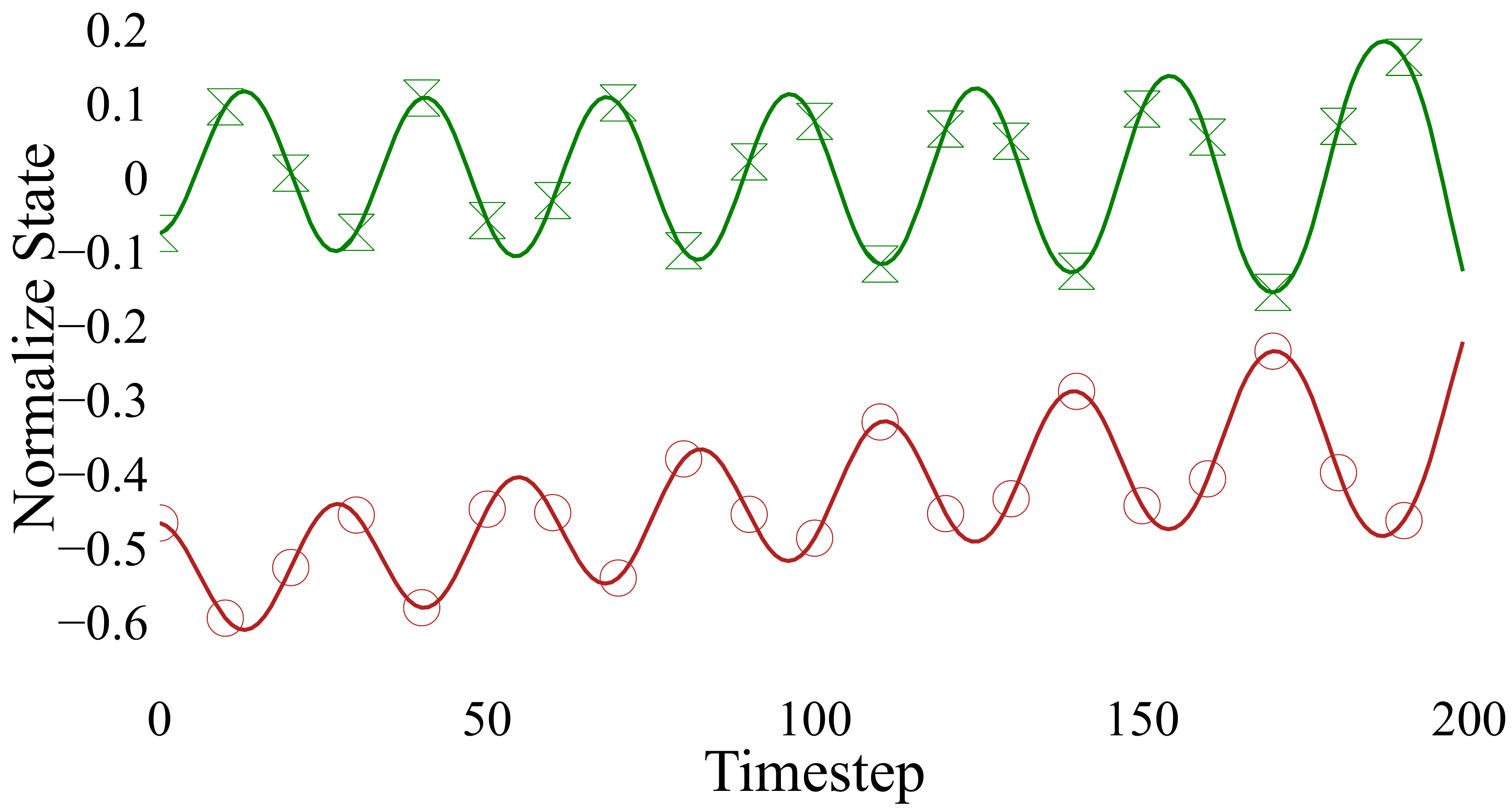}
        \caption{A periodic trajectory.}    
        \label{fig:s-c}
    \end{subfigure}
    \begin{center}
    \small{\cblock{0}{0}{200} Deterministic, one-step: \textit{D} (\textcolor[rgb]{.0,.0,.78}{\large{$\circ$}})\quad
    \cblock{200}{0}{0} Trajectory-based: \textit{T}
    (\textcolor[rgb]{.78,.01,.01}{$\bigplus$}) 
    } 
    \end{center}
    
    \newcommand{\sizefiga}[0]{0.31\linewidth}
    \begin{subfigure}{\sizefiga}  
        \centering 
        \includegraphics[width=\linewidth]{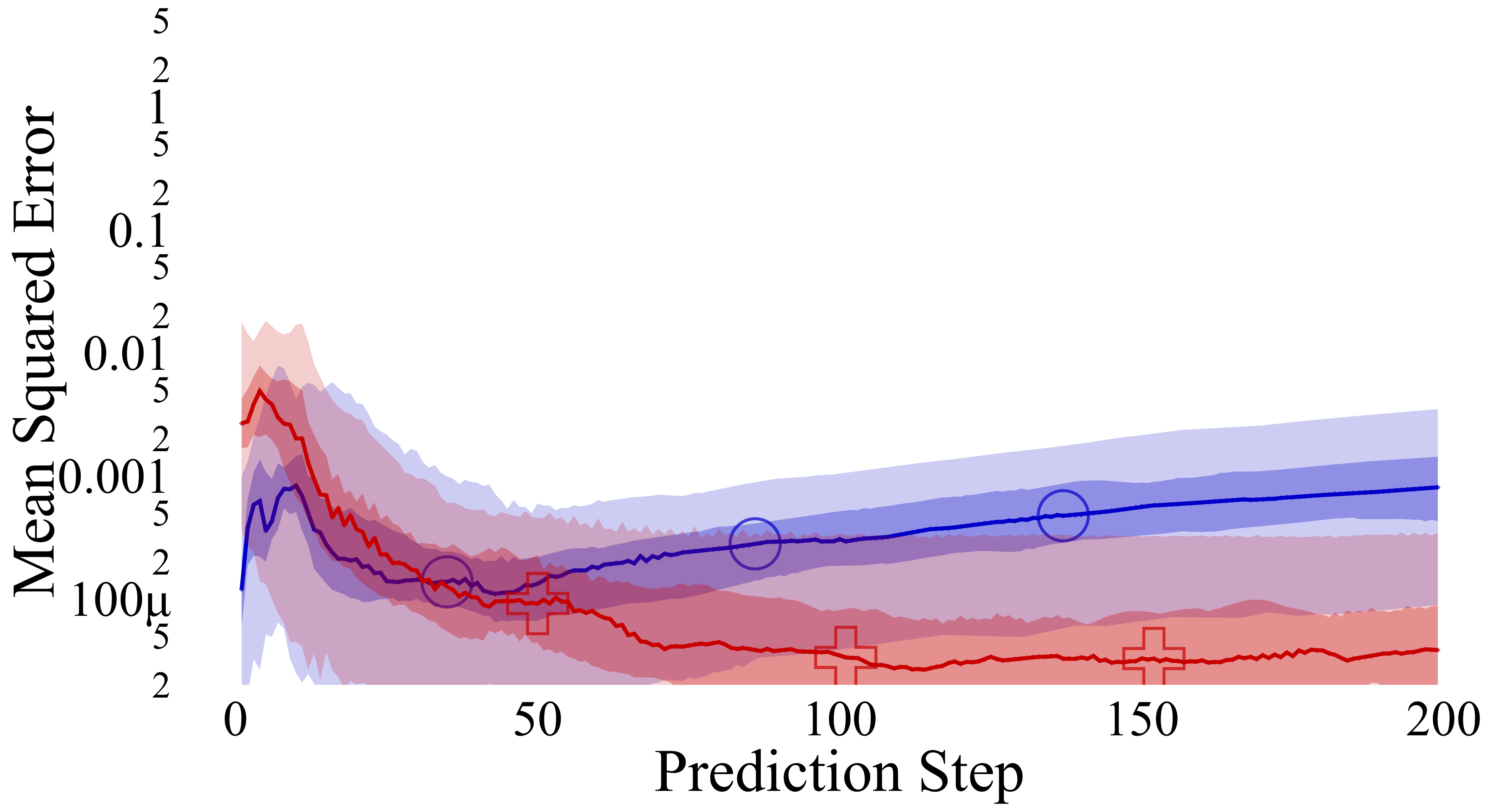}
        \caption{Train: stable, test: stable.}    
        \label{fig:s-s}
    \end{subfigure}
    \hfill
    \begin{subfigure}{\sizefiga}  
        \centering 
        \includegraphics[width=\linewidth]{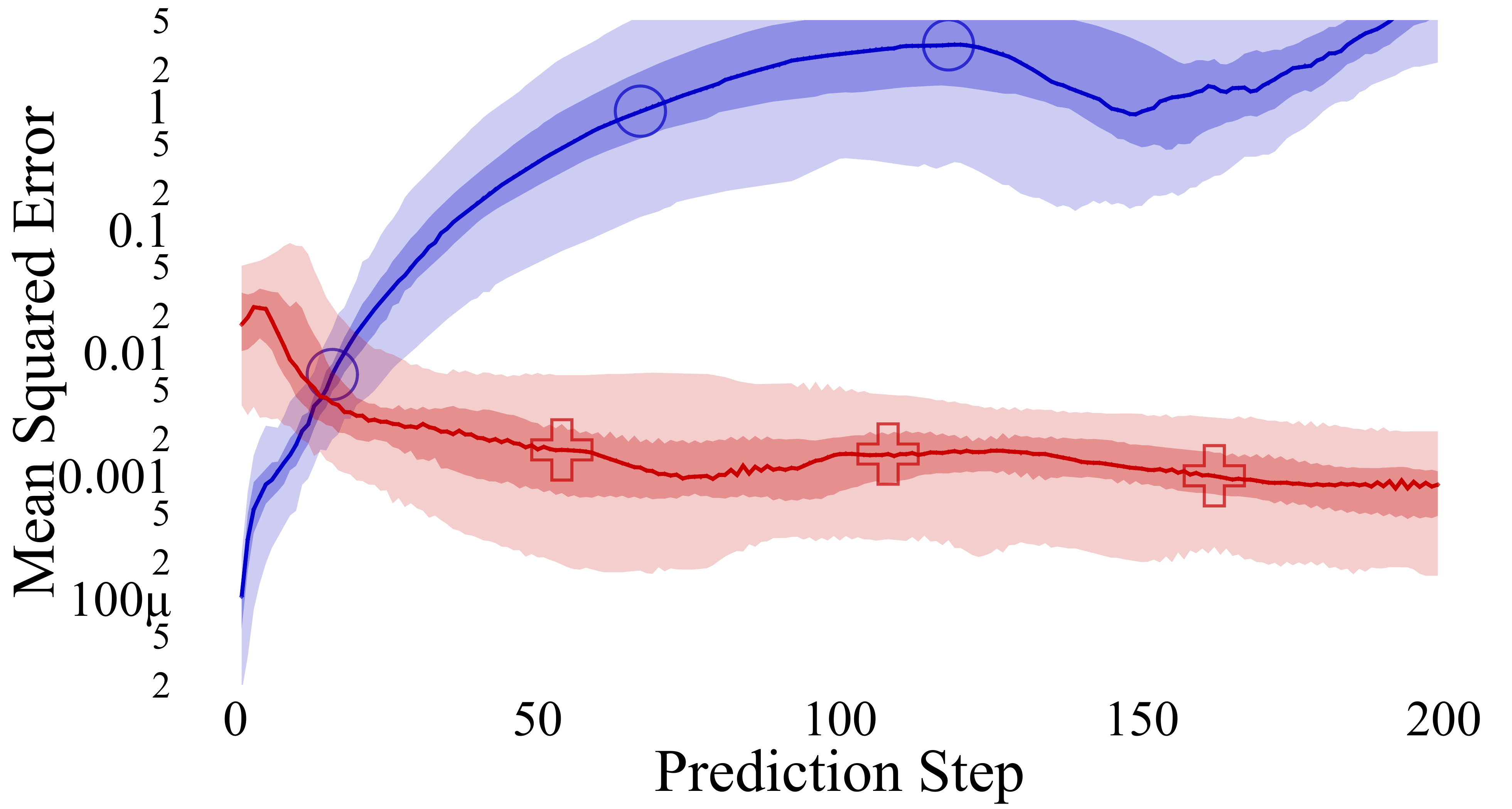}
        \caption{Train: unstable, test: stable.}    
        \label{fig:u-s}
    \end{subfigure}
    \hfill
    \begin{subfigure}{\sizefiga}  
        \centering 
        \includegraphics[width=\linewidth]{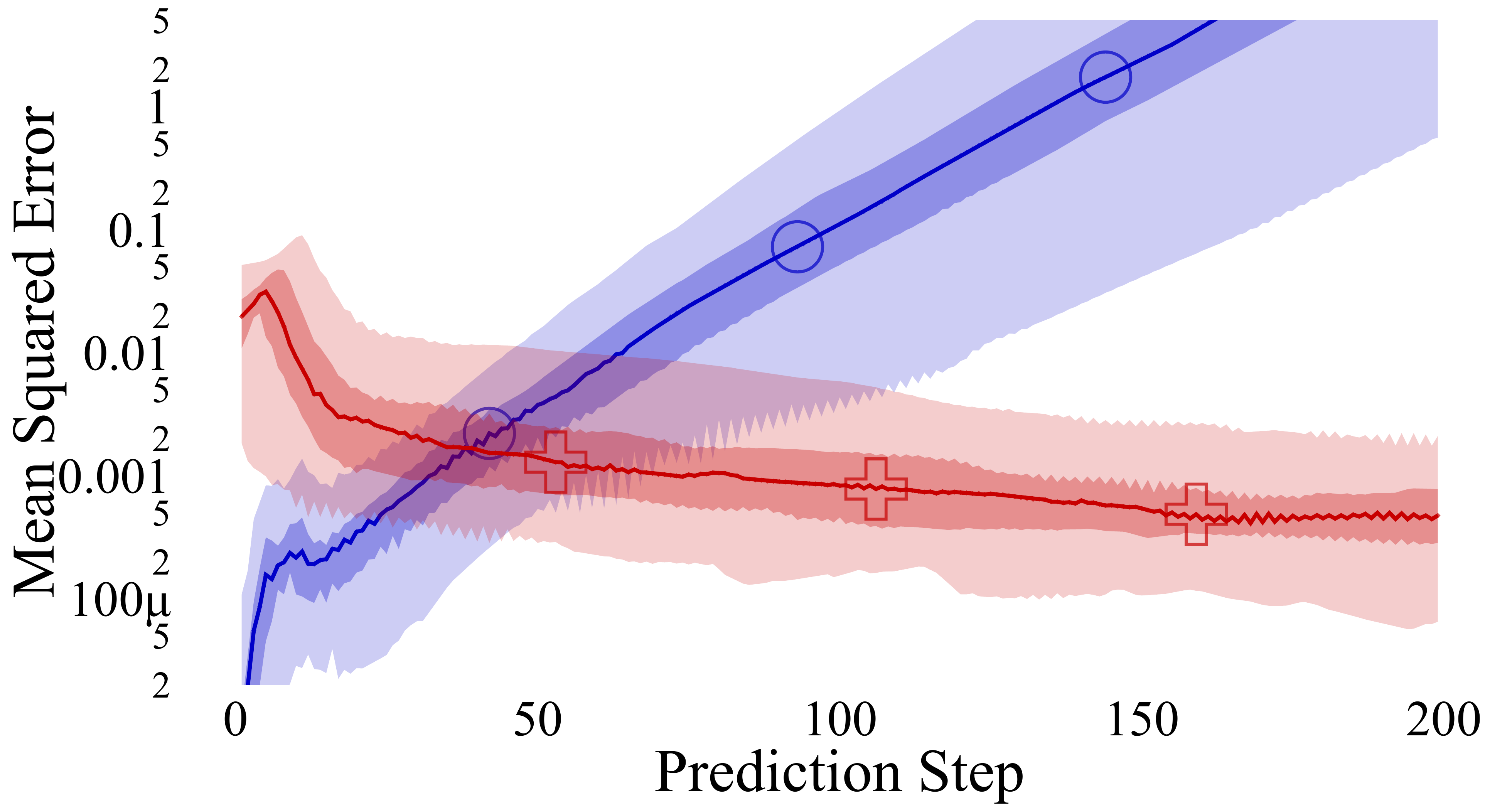}
        \caption{Train: periodic, test: stable.}    
        \label{fig:c-s}
    \end{subfigure}
    \\
    \begin{subfigure}{\sizefiga}  
        \centering 
        \includegraphics[width=\linewidth]{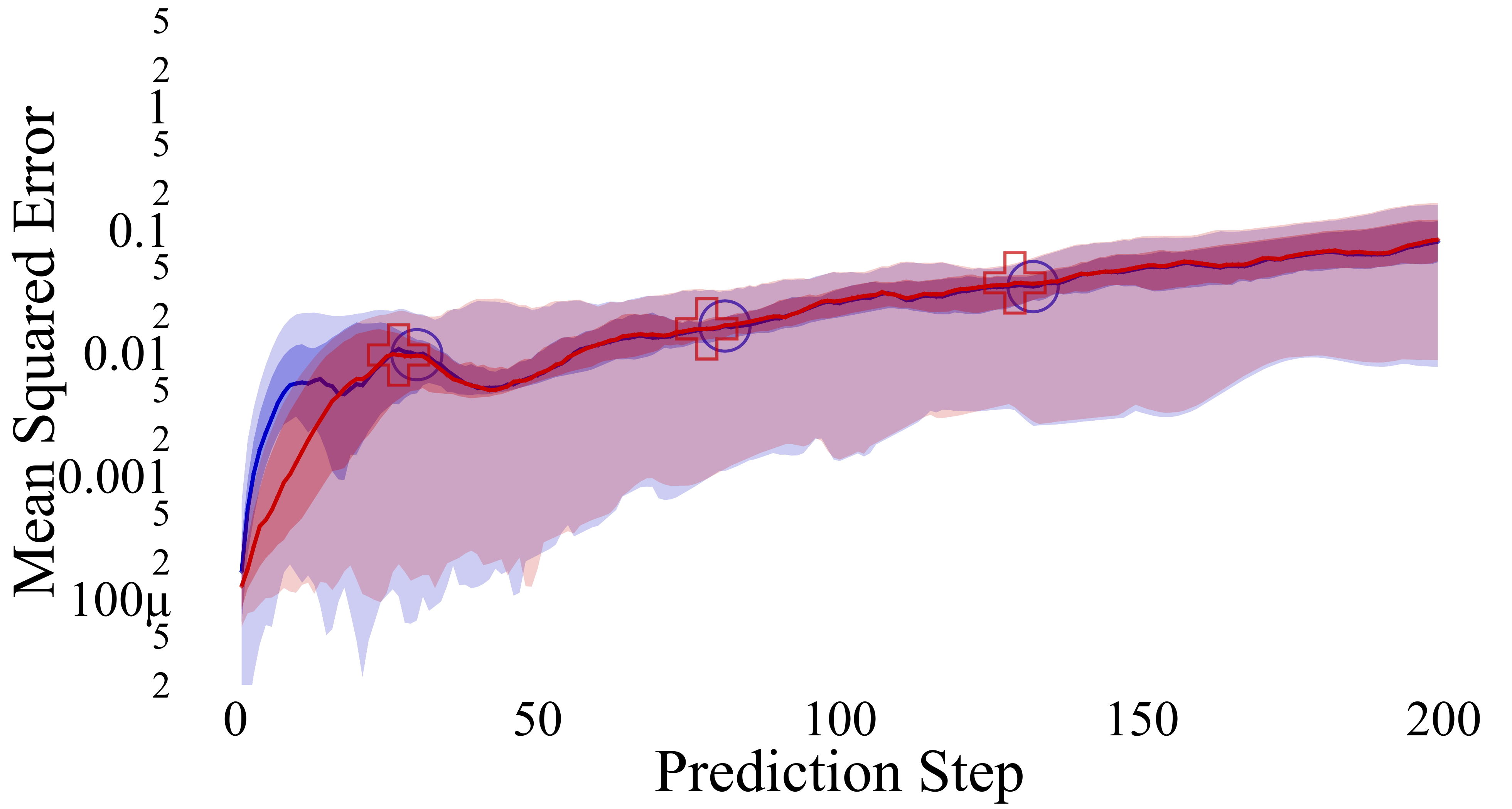}
        \caption{Train: stable, test: unstable.}    
        \label{fig:s-u}
    \end{subfigure}
    \hfill
    \begin{subfigure}{\sizefiga}  
        \centering 
        \includegraphics[width=\linewidth]{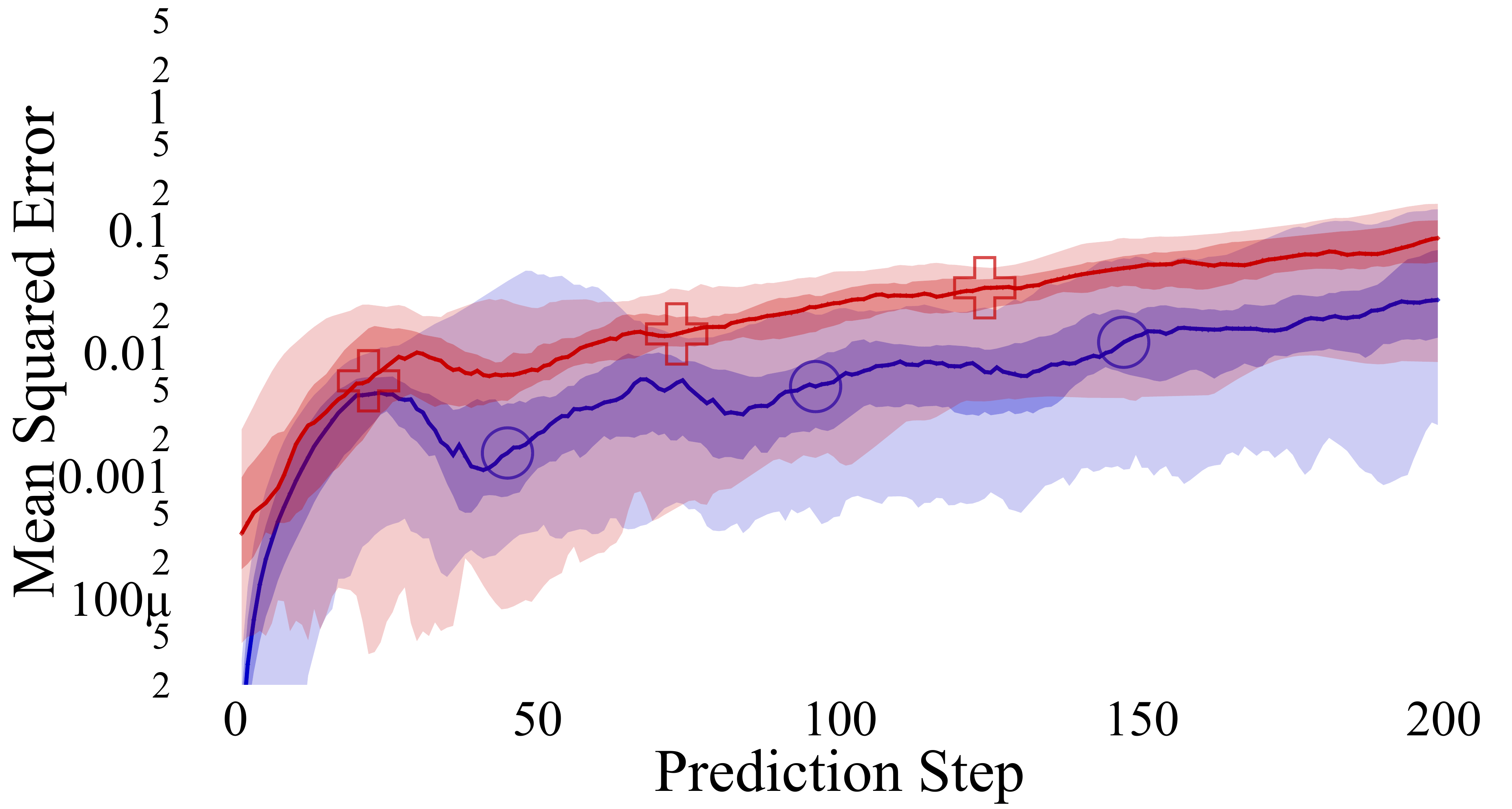}
        \caption{Train: unstable, test: unstable.}    
        \label{fig:u-u}
    \end{subfigure}
    \hfill
    \begin{subfigure}{\sizefiga}  
        \centering 
        \includegraphics[width=\linewidth]{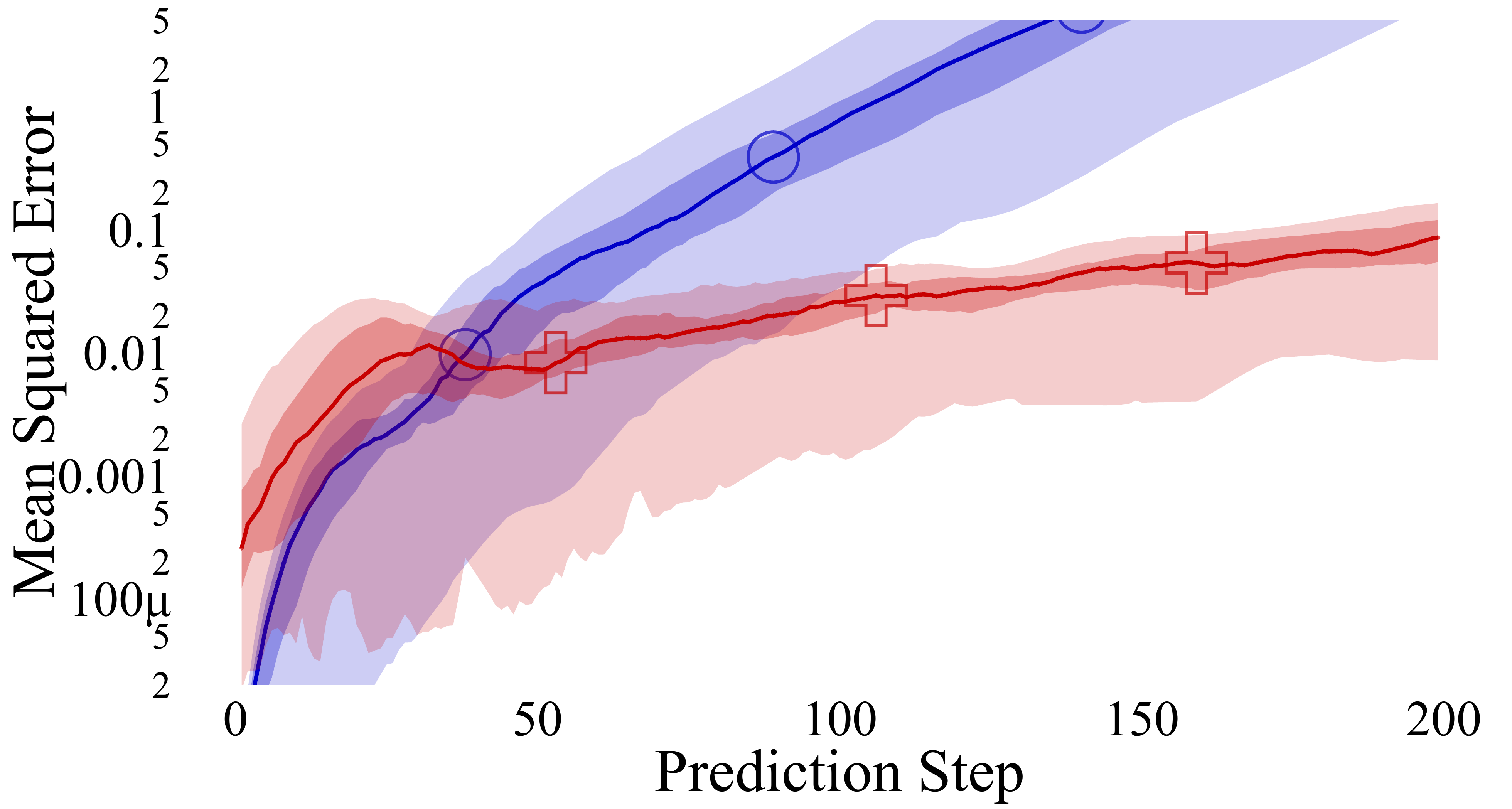}
        \caption{Train: periodic, test: unstable.}    
        \label{fig:c-u}
    \end{subfigure}
    \\
    \begin{subfigure}{\sizefiga}  
        \centering 
        \includegraphics[width=\linewidth]{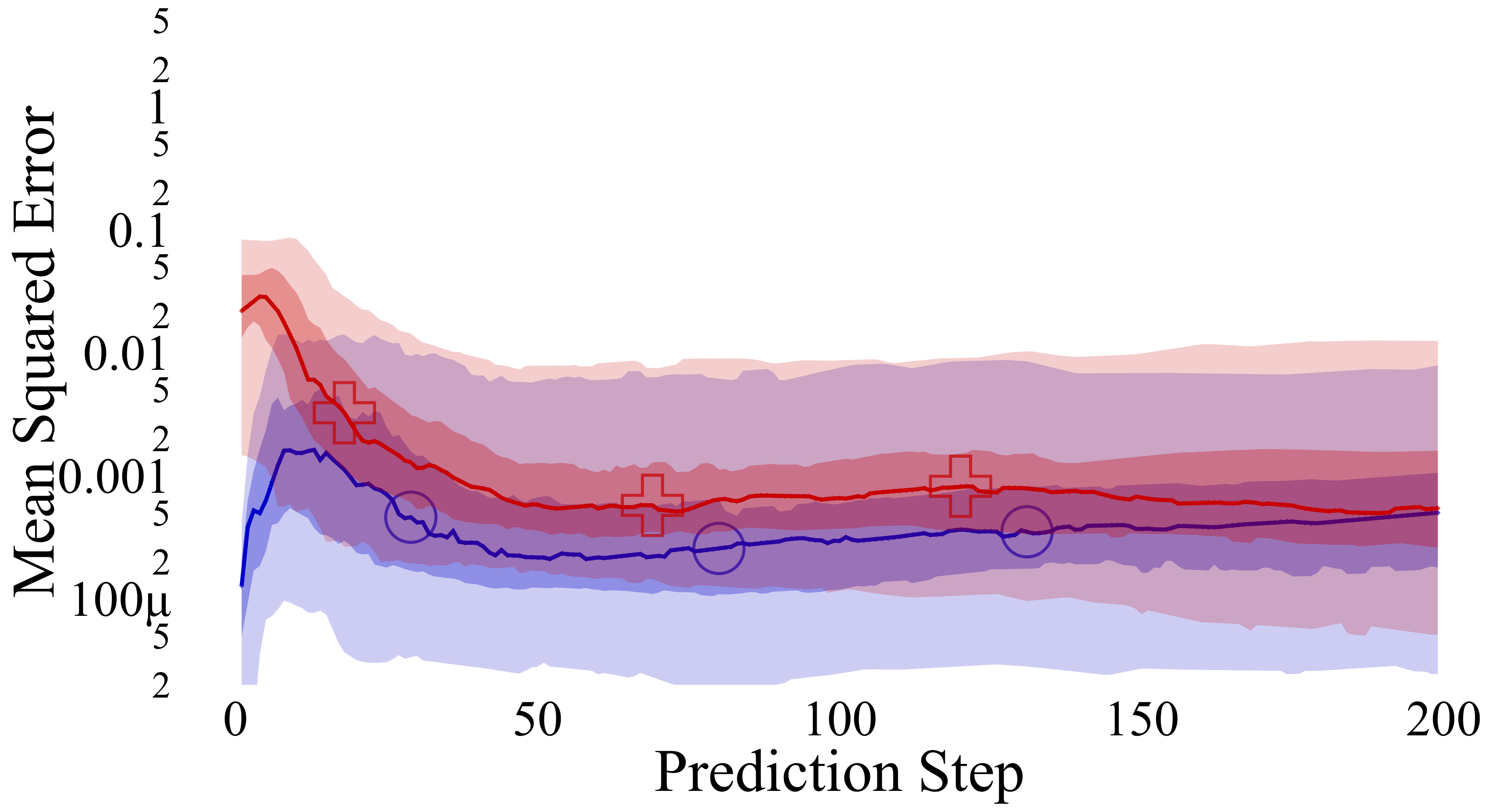}
        \caption{Train: stable, test: periodic.}    
        \label{fig:s-c}
    \end{subfigure}
    \hfill
    \begin{subfigure}{\sizefiga}  
        \centering 
        \includegraphics[width=\linewidth]{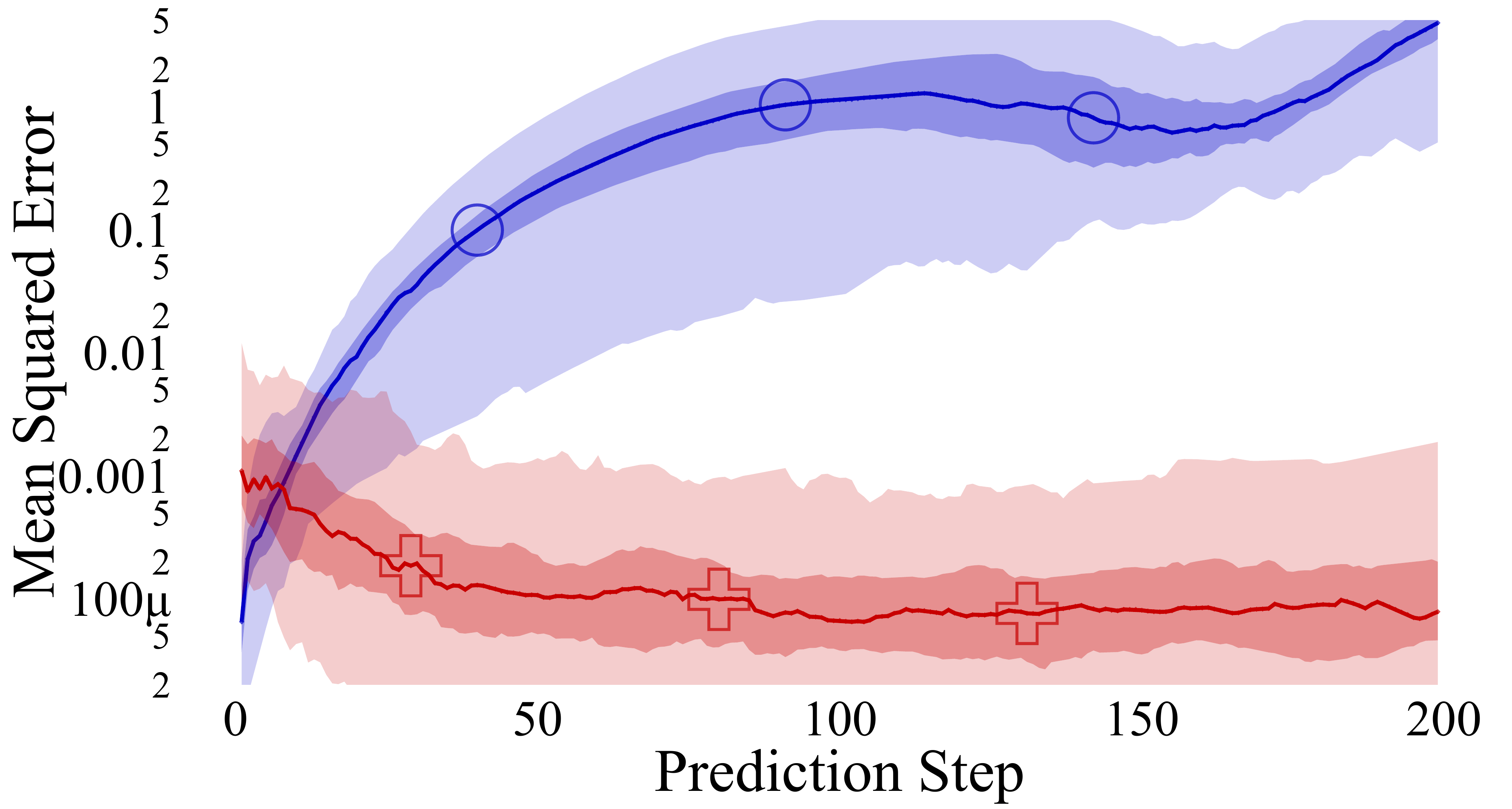}
        \caption{Train: unstable, test: periodic.}    
        \label{fig:u-c}
    \end{subfigure}
    \hfill
    \begin{subfigure}{\sizefiga}  
        \centering 
        \includegraphics[width=\linewidth]{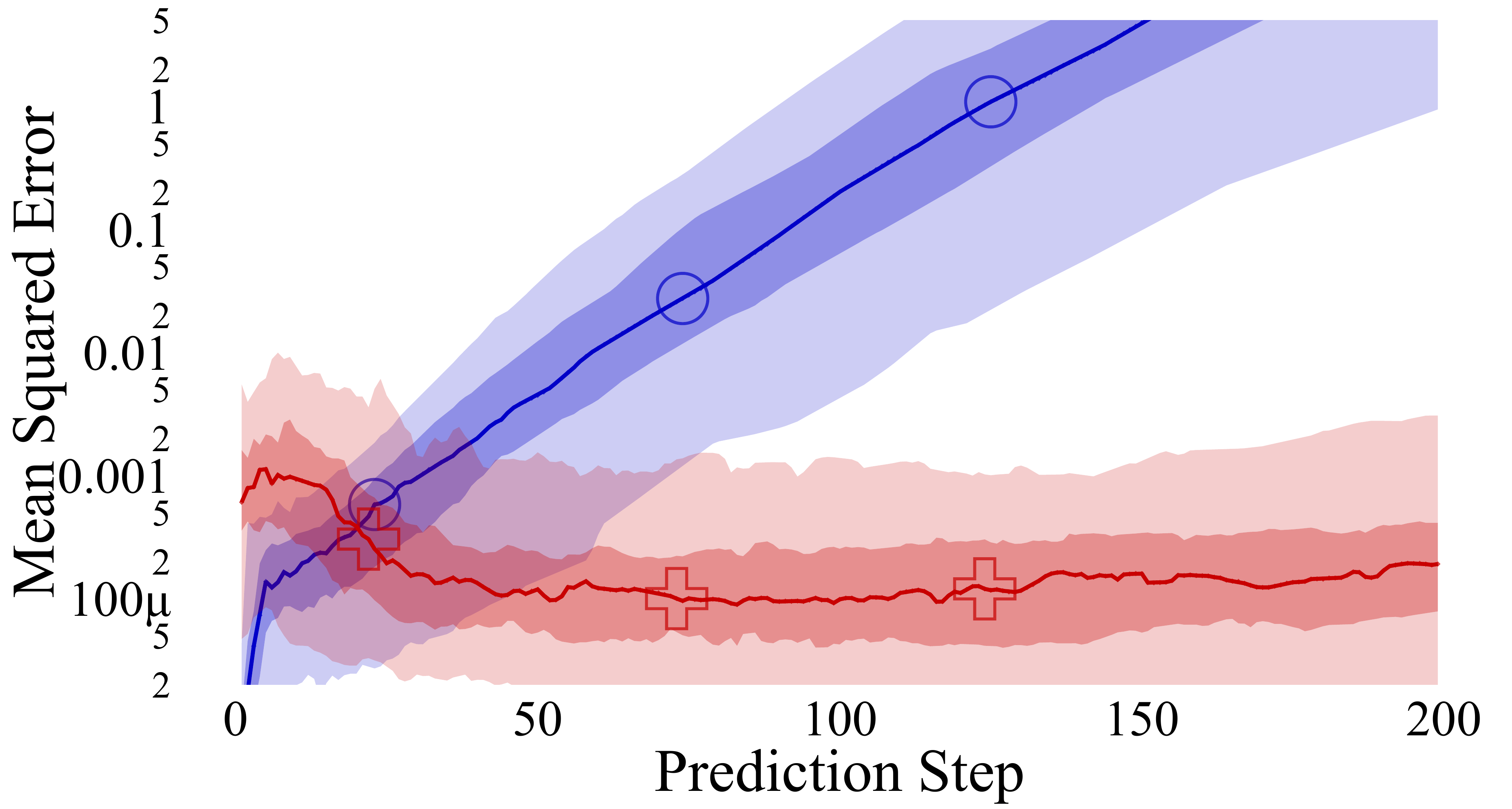}
        \caption{Train: periodic, test: periodic.}    
        \label{fig:c-c}
    \end{subfigure}
    \caption{\textit{Above:} representative trajectories for the stable, unstable, and periodic datasets. 
    \textit{Below:} the evaluation error when using a model trained exclusively on one-variety of data (e.g. unstable) to predict all types of data.
    This figure represents 6 models ($D$ and $T$ trained on $3$ different datasets) and 18 evaluations (the 6 models evaluated on the 3 testing sets).
    }
    \label{fig:data_type}
\vspace{-8pt}
\end{figure*}

\subsection{Predicting Unstable and Periodic Dynamics}
\label{sec:datatype}
Important to the application of dynamics models to robotic tasks is the ability to accurately model dynamics when a) trained on imperfect data (e.g. data with noise and divergent modes) and b) evaluated on different modes of data.
In this section, the evaluation of predicting stable dynamics is extended onto unstable and periodic dynamics.
Representative stable, unstable, and periodic trajectories are shown in \fig{fig:data_type}a,b,c.
Unstable dynamics are designed to be diverging through the trajectory and periodic dynamics have consistent cyclic motion of various frequencies.

To test this, we collect 3 training and testing datasets ($N_{\text{traj}}=100$) for each datatype above in the cartpole environment via different tunings of LQR control.
The stable data is the same used in \sect{sec:pred} where the majority of trajectories converge towards $\vec{s}=0$.
The one-step models maintain similar performance to trajectory-based models when trained on stable data, shown in \fig{fig:data_type}d,g,j.
In the unstable-data trained models (\fig{fig:data_type}e,h,k) and the periodic-data trained models (\fig{fig:data_type}f,i,l), the trajectory-based models demonstrate an impressive performance in modelling dynamics motion across long horizons.
Except when trained and tested on unstable data, the one-step models diverge rapidly -- suggesting two potential modes of training for the one-step models: 
a) they may be memorizing the unstable trajectories and 
b) when training on stable data, the majority of the delta-state predictions are around 0, so when predicting out of distribution data the model may predict no change, slowing divergence.
Conversely, when trained on unstable or periodic data, and testing elsewhere, the one-step models diverge rapidly due to the constant change in state in the training set, shown in \fig{fig:data_type}e,f,i,k,l.
The ability for trajectory-based models to generalize from periodic data to stable dynamics confers a substantial improvement over the one-step model.

\subsection{Optimizer Details for \sect{sec:iterative}}
The sample spaces for optimization are important to convergence (too broad of a space and the algorithms will not converge), and they follow \sect{sec:exp}. 
For the cartpole task, the sample space of the optimization is from $c \in [-0.1,2]$ for $c\cdot \vec{K}$ where $\vec{K}$ is the optimal LQR controller.
For the reacher task, the sample space is the range of PID and target parameters sampled.

\paragraph{Bayesian Optimization}
For Bayesian Optimization, we used the open source platform Ax \url{https://ax.dev/}.
All targets (rewards) were normalized during training to [0,1] to aid in GP fitting.

\paragraph{CMA-ES}
For covariance matrix adaptation evolution strategy (CMA-ES) we use the open source python implementation \url{https://pypi.org/project/cma/}. 
In order to constrain the sample space, any proposed parameter set outside the bounds is given a reward of -10000 (same for all tasks). 

\subsection{Parameter Tuning}
A table of the limited hyperparameter tuning done in this paper is shown in \tab{tab:paramspets}.

\begin{table}[h]
\begin{center}
 \begin{tabular}{ l c c} 
 Parameter & Final Value & Swept values  \\ [1.0ex]

 \hline 
   \multicolumn{3}{c}{Standard Feed-forward Models}  \\ [0.5ex] 
 Optimizer & Adam & Adam, SGD  \\
 \hline
 Hidden width & 250   & 250,300, 500   \\
 \hline
  Hidden depth & 2  &   2,3 \\
 \hline
 Batch Size & 32  & 16,32,64,128, 256  \\ 
 \hline
 Learning Rate & 5E-5 & 1E-4, 1E-5, 5E-5  \\ 
  \hline
 Test Train Split & 0.9 & 0.8, 0.9, 1.0 \\ [1.0ex]

   \multicolumn{3}{c}{Trajectory Feed-forward Models}  \\ [0.5ex] 
 Optimizer & Adam & Adam, SGD  \\
 \hline
 Hidden width & 250   & 250,300, 500   \\
 \hline
  Hidden depth & 2  &   2,3 \\
 \hline
 Batch Size & 64  & 16,32,64,128, 256  \\ 
 \hline
 Learning Rate & 8E-4 & 1E-4, 4E-4, 8E-4  \\ 
  \hline
 Test Train Split & 0.8 & 0.8, 0.9, 1.0 \\
 \hline
 Max training set size & 1E5 & 5E4,1E5,2E5,5E5,1E6  \\ [1.0ex]

   \multicolumn{3}{c}{Recurrent Models}  \\ [0.5ex] 
 Optimizer & Adam & Adam, SGD  \\
 \hline
 Hidden width & 250   & 250,300, 500   \\
 \hline
  Hidden depth & 2  &   2,3 \\
 \hline
 Batch Size & 1  & 1,2,4  \\ 
 \hline
 Learning Rate & 1E-3 & 1E-3, 1E-4, 8E-5  \\ 
  \hline
 Test Train Split & 0.8 & 0.8, 0.9, 0.5 \\
 \hline
 Training trajectory length & L & 100, 200, L  \\ [1.0ex]
\end{tabular}
\caption{Model training hyperparameters}
\label{tab:paramspets}
\end{center}

\end{table}

\end{document}